\documentclass{article}

\usepackage[numbers]{natbib}

    \usepackage[final]{neurips_2022}

\usepackage[utf8]{inputenc} % allow utf-8 input
\usepackage[T1]{fontenc}    % use 8-bit T1 fonts
\usepackage{url}            % simple URL typesetting
\usepackage{booktabs}       % professional-quality tables
\usepackage{amsfonts}       % blackboard math symbols
\usepackage{nicefrac}       % compact symbols for 1/2, etc.
\usepackage{microtype}      % microtypography
\usepackage[dvipsnames]{xcolor}         % colors
\usepackage{graphics}
\usepackage{amsmath}
\usepackage{mathtools}
\usepackage{amsthm}
\usepackage{amssymb}
\usepackage{wrapfig}
\usepackage{lipsum}
\usepackage{array}
\usepackage{floatrow}
\usepackage[footnotesize]{caption}
\usepackage{caption}
\usepackage{algorithm}
\usepackage{algorithmic}
\usepackage[colorlinks=true,citecolor=blue,urlcolor=blue,linkcolor=blue,anchorcolor=blue]{hyperref}

\usepackage{amsmath}
\usepackage{subcaption}
\usepackage{capt-of}
\usepackage{enumitem}
\usepackage{cleveref}
\usepackage{svg}
\usepackage{amsmath,amsfonts,bm}

\def\eqref#1{equation~\ref{#1}}
\def\1{\bm{1}}
\newcommand{\train}{\mathcal{D}}

\DeclareMathAlphabet{\mathsfit}{\encodingdefault}{\sfdefault}{m}{sl}
\SetMathAlphabet{\mathsfit}{bold}{\encodingdefault}{\sfdefault}{bx}{n}

\def\gA{{\mathcal{A}}}

\def\gD{{\mathcal{D}}}

\def\gM{{\mathcal{M}}}
\def\gN{{\mathcal{N}}}
\def\gO{{\mathcal{O}}}
\def\gP{{\mathcal{P}}}

\def\gR{{\mathcal{R}}}
\def\gS{{\mathcal{S}}}
\def\gT{{\mathcal{T}}}
\def\gU{{\mathcal{U}}}

\renewcommand{\train}{\text{train}}
\newcommand{\testt}{\text{test}}
\newcommand{\regret}{\operatorname{Regret}}
\newcommand{\meta}{\text{meta}}

\newcommand{\E}{\mathbb{E}}

\renewcommand{\bar}{\overline}
\DeclareMathOperator*{\argmin}{arg\,min}

\usepackage{float}

\usepackage{thmtools}
\usepackage{thm-restate}

\newtheorem{theorem}{Theorem}[section]

\newtheorem{lemma}[theorem]{Lemma}

\DeclareCaptionLabelFormat{andtable}{#1~#2  \&  \tablename~\thetable}
\title{Distributionally Adaptive Meta Reinforcement Learning}
\makeatletter
\newcommand{\printfnsymbol}[1]{%
  \textsuperscript{\@fnsymbol{#1}}%
}

\author{Anurag Ajay\thanks{\fontsize{8}{8} denotes equal contribution. Authors are also affiliated with Computer Science and Artificial Laboratory (CSAIL). Correspondence to \texttt{aajay@mit.edu} and \texttt{abhgupta@cs.washington.edu}}\;\;\printfnsymbol{2}\printfnsymbol{3}\printfnsymbol{4}, ~Abhishek Gupta \printfnsymbol{1}\printfnsymbol{2}\printfnsymbol{4}, ~Dibya Ghosh\printfnsymbol{5}, ~Sergey Levine\printfnsymbol{5}, ~Pulkit Agrawal\printfnsymbol{2}\printfnsymbol{3}\printfnsymbol{4} \\
Improbable AI Lab\printfnsymbol{2}\\ 
MIT-IBM Watson AI Lab\printfnsymbol{3} \\
Massachusetts Institute Technology\printfnsymbol{4}\\
University of California, Berkeley\printfnsymbol{5}\\
}

\begin{document}

\newcommand{\cmt}[1]{{\footnotesize\textcolor{red}{#1}}}
\newcommand{\todo}[1]{\cmt{(TODO: #1)}}
\newcommand{\kelvin}[1]{{\footnotesize\textcolor{red}{Kelvin: #1}}}
\newcommand{\boldme}[1]{{\footnotesize\textbf{#1}}}
\newcommand{\ourmethod}{\boldme{METHOD\_NAME}}
\newcommand{\rml}{DiAMetR} 

% Math operators and functions
% \newcommand{\E}[2]{\operatorname{\mathbb{E}}_{#1}\left[#2\right]}
\newcommand{\expectation}{\mathbb{E}}
\newcommand{\density}{p}
\newcommand{\proposal}{q}  % proposal distribution
\newcommand{\target}{p}  % target distribution
\newcommand{\kl}[2]{\mathrm{D_{KL}}\left(#1\;\middle\|\;#2\right)}
\newcommand{\fdiv}[2]{D_{f}\left(#1\;\middle\|\;#2\right)}
\newcommand{\entropy}{\mathcal{H}}
\newcommand{\information}{\mathcal{I}}
\newcommand{\ent}{\mathcal{H}}
\newcommand{\sdots}{\,\cdot\,}
\newcommand{\func}{\mathbf{f}}
\newcommand{\methodname}{RLDE }
\newcommand{\piexplore}{$\pi_{\text{explore}}$}
\newcommand{\piexploit}{$\pi_{\text{exploit}}$}
\newcommand{\rhoexploit}{$\rho_{\text{exploit}}(s,a)$ }
\newcommand{\rhoexplore}{$\rho_{\text{explore}}(s,a)$ }

% Density ratio stuff

\newcommand{\ratio}{r(x)}
\newcommand{\classifier}{D(x)}
\newcommand{\dataset}{\mathcal{X}}

%DIAYN
\newcommand{\context}{\mathbf{z}}

\newcommand{\objective}{\mathcal{J}}

% Constant matrices and vectors.
\newcommand{\info}{\mathcal{I}}

% Constant matrices and vectors.
\newcommand{\ones}{\boldsymbol{1}}
\newcommand{\eye}{\boldsymbol{I}}
\newcommand{\zeros}{\boldsymbol{0}}

%meta learning
\newcommand{\task}{\mathcal{T}}
\newcommand{\metatrain}{\mathcal{T}^\text{meta-train}}
\newcommand{\metatest}{\mathcal{T}^\text{meta-test}}

\newcommand{\metatrainset}{\{\mathcal{T}_i~;~ i=1..N\}}
\newcommand{\metatestset}{\{\mathcal{T}_j~;~ j=1..M\} }
%\newcommand{\metatestset}{\{\mathcal{T}_j~;~ j\!\!=\!\!1..M\}^\text{meta-test} }

% parameters
\newcommand{\vphi}{\boldsymbol{\phi}}
% \newcommand{\vtheta}{\boldsymbol{\theta}}
% \newcommand{\vmu}{\boldsymbol{\mu}}

% supervised
% \newcommand{\vx}{\mathbf{x}}
% \newcommand{\vy}{\mathbf{y}}
\newcommand{\discrim}{q_{\omega}}

% MDP
\newcommand{\sspace}{\mathcal{S}}
\newcommand{\aspace}{\mathcal{A}}
\newcommand{\state}{s}
\newcommand{\sz}{{\state_0}}
\newcommand{\stm}{{\state_{t-1}}}
\newcommand{\st}{{\state_t}}
\newcommand{\sti}{{\state_t^{(i)}}}
\newcommand{\stp}{{\state_{t+1}}}
\newcommand{\stpi}{{\state_{t+1}^{(i)}}}
\newcommand{\pdyn}{\density_\state}
\newcommand{\learnedmodel}{\hat{\density}_\state}
\newcommand{\pz}{\density_0}
\newcommand{\horizon}{H_T}
\newcommand{\evalhorizon}{H_E}
\newcommand{\action}{a}
\newcommand{\az}{{\action_0}}
\newcommand{\atm}{{\action_{t-1}}}
\newcommand{\at}{{\action_t}}
\newcommand{\ati}{{\action_t^{(i)}}}
\newcommand{\atj}{{\action_t^{(j)}}}
\newcommand{\attildej}{{\tilde{\action}_t^{(j)}}}
\newcommand{\atij}{{\action_t^{(i,j)}}}
\newcommand{\atp}{{\action_{t+1}}}
\newcommand{\aT}{{\action_T}}
\newcommand{\atk}{{\action_t^{(k)}}}
\newcommand{\aTm}{\action_{\horizon-1}}
\newcommand{\opt}{^*}
% set of MDPs
\newcommand{\mdps}{\mathcal{M}}

% Reset
%\newcommand{\explorepolicy}{\pi^{\text{explore}}_\theta}
%\newcommand{\exploitpolicy}{\pi^{\text{exploit}}_{\phi}}

\newcommand{\explorei}{\pi^{\text{explore}}_{i}}

\newcommand{\explorepolicy}{\pi^{\text{explore}}_{\phi}}
\newcommand{\exploitpolicy}{\pi^{\text{exploit}}_{\theta}}

% Trajectories
\newcommand{\demos}{\mathcal{D}}
\newcommand{\traj}{\tau}
\newcommand{\ptraj}{\density_\traj}
\newcommand{\visits}{\rho}  % Discounted visittation frequency

% Rewards
\newcommand{\reward}{r}
\newcommand{\rti}{\reward^{(i)}_t}
\newcommand{\rmi}{r_\mathrm{min}}
\newcommand{\rmax}{r_\mathrm{max}}

\newcommand{\return}{J}

%

% Loss
\newcommand{\loss}{\mathcal{L}}

% Optimality
\newcommand{\policyopt}{\pi^*}

% Value and Q function
\newcommand{\V}{V}
\newcommand{\Vsoft}{V_\mathrm{soft}}
\newcommand{\Vsoftparams}{V_\mathrm{soft}^\qparams}
\newcommand{\Vhatsoftparams}{\hat V_\mathrm{soft}^\qparams}
\newcommand{\Vhatsoft}{\hat V_\mathrm{soft}}
\newcommand{\Vhard}{V^{\dagger}}
\newcommand{\Q}{Q}
\newcommand{\Qsoft}{Q_\mathrm{soft}}
\newcommand{\Qsoftparams}{Q_\mathrm{soft}^\qparams}
\newcommand{\Qhatsoft}{\hat Q_\mathrm{soft}}
\newcommand{\Qhatsoftparams}{\hat Q_\mathrm{soft}^{\bar\qparams}}
\newcommand{\Qhard}{Q^{\dagger}}
\newcommand{\A}{A}
\newcommand{\Asoft}{A_\mathrm{soft}}
\newcommand{\Asoftparams}{A_\mathrm{soft}^\qparams}
\newcommand{\Ahatsoft}{\hat A_\mathrm{soft}}

% EBM
\newcommand{\energy}{\mathcal{E}}

\newcommand{\tasklosstrain}{\loss_{\task}^{\text{ tr}}}
\newcommand{\tasklosstest}{\loss_{\task}^{\text{ test}}}

\newcommand{\tasklosstraini}{\loss_{\task_i}^{\text{ tr}}}
\newcommand{\tasklosstesti}{\loss_{\task_i}^{\text{ test}}}

\maketitle

\begin{abstract}
% Meta-reinforcement learning algorithms provide a data-driven way to learn adaptive learning algorithms that are able to quickly adapt to new tasks with varying rewards or dynamics functions. However, most meta-reinforcement learning algorithms are effective only within a predefined training task distribution and struggle to adapt in the presence of test-time distribution shift in rewards or transition dynamics. In this work, we show that we can build algorithms that are able to adapt to new tasks even in the presence of distribution shift in the space of tasks. The key insight this work makes is to propose an adaptive distributional robustness framework for meta-reinforcement learning, which allows for a population of adaptive agents to be learned corresponding to varying levels of distribution shift, which can then be used to adapt to arbitrarily shifted test distributions. We empirically show the efficacy of this framework on a variety of different robotics problems in simulation and theoretically show how this framework allows for improved worst case regret on a wide range of realistic distribution shifts.
% DG: Below is my attempt to reword here and there 
% AA: This is really nice. I am going to remove the world "realistic" as our envs are toyish in a way.
Meta-reinforcement learning algorithms provide a data-driven way to acquire policies that quickly adapt to many tasks with varying rewards or dynamics functions. However, learned meta-policies are often effective only on the exact task distribution on which they were trained and struggle in the presence of distribution shift of test-time rewards or transition dynamics. In this work, we develop a framework for meta-RL algorithms that are able to behave appropriately under test-time distribution shifts in the space of tasks. 
% Our framework centers on an adaptive approach to distributional robustness, in which we train a population of meta-policies to be robust to varying levels of distribution shift, so that when evaluated on a potentially shifted test-time distribution of tasks, we can adaptively choose the meta-policy with the most appropriate level of robustness, and use it to perform fast adaptation.
Our framework centers on an adaptive approach to distributional robustness that trains a population of meta-policies to be robust to varying levels of distribution shift. When evaluated on a potentially shifted test-time distribution of tasks, this allows us to choose the meta-policy with the most appropriate level of robustness, and use it to perform fast adaptation.
%This meta-policy is used to perform fast adaptation. 
We formally show how our framework allows for improved regret under distribution shift, and empirically show its efficacy on simulated robotics problems under a wide range of distribution shifts.
\end{abstract}

\section{Introduction}
\label{sec:intro}

% \pulkit{Maybe this is too late for a radical change, but here is one suggestion. There are two kinds of distribution shifts: (i) The support of training goals / rewards is the same, but we sample the same rewards/goals with a different distribution. (ii) We sample new goals and rewards that are beyond the training set.  
% \begin{itemize}
%     \item The idea of having a population of meta-learners can be powerful for (i). 
%     \item The idea of generating new rewards/goals by optimizing for worst case robust attacks (ii). Doing this along with (i) prevents just "robust" optimization from being conservative. 
% \end{itemize}

% The point being there are two distinct ideas for two distinct kinds of "shifts" in test-time distribution -- both them are interesting in their own respect. 

% }

% [Why do we need adaptive agents, motivate meta-RL]
% Reinforcement learning (RL) agents have the potential for continual improvement as they collect data interactively. This becomes
% increasingly important as we deploy reinforcement learning enabled agents in dynamic, real world environments. Ideally, with time, our RL agents should not only become more proficient at a larger variety of tasks, but also become more efficient in learning new tasks. 

\begin{wrapfigure}{r}{0.4\textwidth}
    \vspace{-0.5cm}
    \centering
    \includegraphics[width=\linewidth]{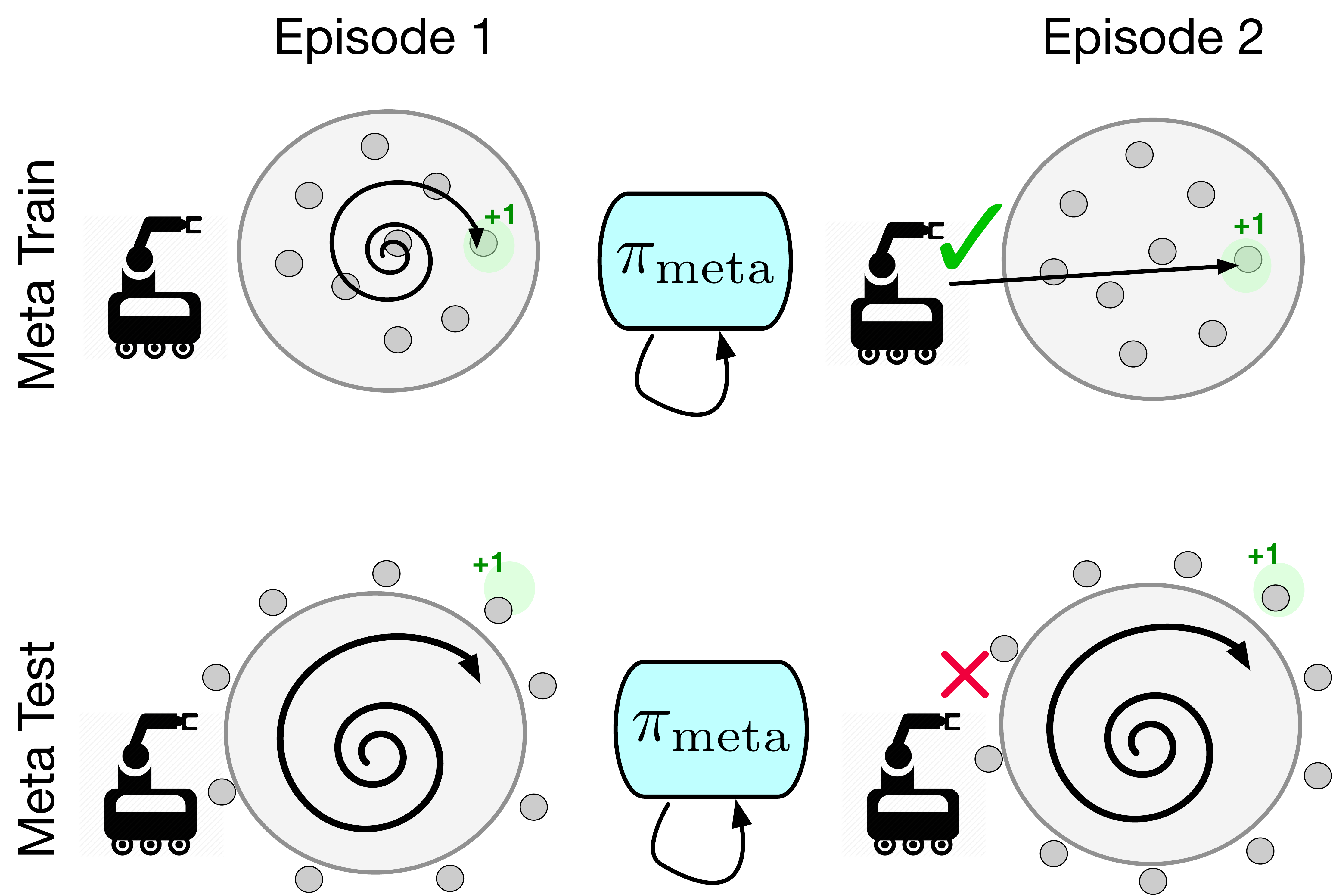}
    \caption{\textbf{Failure of Typical Meta-RL.} On meta-training tasks, $\pi_{\text{meta}}$ explores effectively and quickly learns the optimal behavior (top row).  When test tasks come from a slightly larger task distribution, exploration fails catastrophically, resulting in poor adaptation behavior (bottom row).}
    \label{fig:failure_mtrl}
\end{wrapfigure}

% DG try
The diversity and dynamism of the real world require reinforcement learning (RL) agents that can quickly adapt and learn new behaviors when placed in novel situations. Meta reinforcement learning provides a framework for conferring this ability to RL agents, by learning a ``meta-policy'' trained to adapt as quickly as possible to tasks from a provided training distribution \citep{thrun98metalearning, finn2017model,  rakelly2019efficient, zintgraf2019varibad}. 
%Meta-RL algorithms train on a distribution of training tasks to obtain ``meta-policies" that can quickly adapt to new tasks drawn from the same task distribution. 
Unfortunately, meta-RL agents assume tasks to be always drawn from the training task distribution and often behave erratically when asked to adapt to tasks beyond the training distribution \citep{deleu2018effects, fallah2021generalization}. As an example of this negative transfer, consider using meta-learning to teach a robot to navigate to goals quickly (illustrated in  Figure~\ref{fig:failure_mtrl}). The resulting meta-policy learns to quickly adapt and walk to any target location specified in the training distribution, but explores poorly and fails to adapt to any location not in that distribution. This is particularly problematic for the meta-learning setting, since the scenarios where we need the ability to learn quickly are usually exactly those where the agent experiences distribution shift. This type of meta-distribution shift afflicts a number of real-world problems including autonomous vehicle driving~\citep{filos2020can}, in-hand manipulation~\citep{ke2021grasping, pmlr-v164-chen22a}, and quadruped locomotion~\citep{miki2022learning, margolis2022rapid, kumar2021rma}, where training task distribution may not encompass all real-world scenarios.
In this work, we study meta-RL algorithms that learn meta-policies resilient to task distribution shift at test time. We assume the test-time distribution shift to be unknown but fixed. One approach to enable this resiliency is to leverage the framework of distributional robustness~\citep{sinha2017certifying}, training meta-policies that prepare for distribution shifts by optimizing the \emph{worst-case} empirical risk against a set of task distributions which lie within a bounded distance from the original training task distribution (often referred to as an \emph{uncertainty set})). This allows meta-policies to deal with potential test-time task distribution shift, bounding their worst-case test-time regret for distributional shifts within the chosen uncertainty set. However, choosing an appropriate uncertainty set can be quite challenging without further information about the test environment, significantly impacting the test-time performance of algorithms under distribution shift. Large uncertainty sets allow resiliency to a wider range of distribution shifts, but the resulting meta-policy adapts very slowly at test time; smaller uncertainty sets enable faster test-time adaptation, but leave the meta-policy brittle to task distribution shifts. Can we get the best of both worlds?

\begin{figure}[!t]
    \centering
    \includegraphics[width=\linewidth]{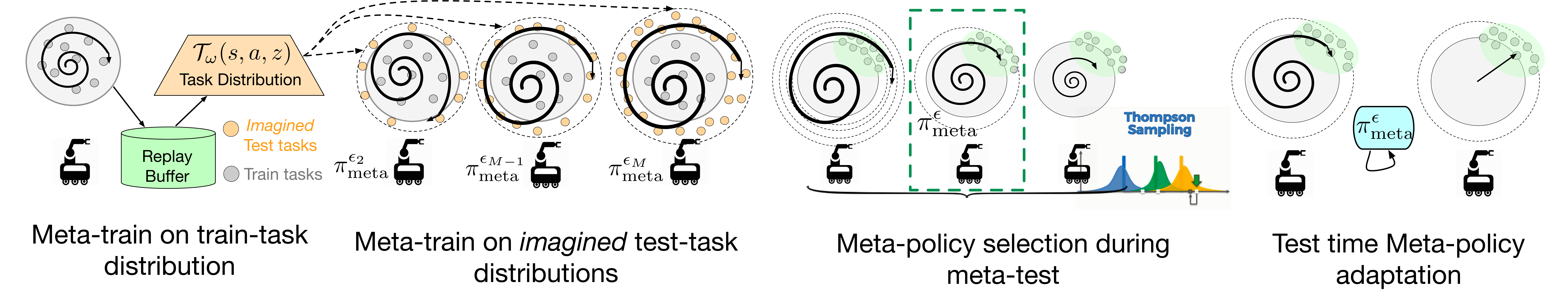}
    \caption{During meta-train, \rml~learns a meta-policy $\pi_{\text{meta}}^{\epsilon_1}$ and task distribution model $\gT_\omega(s,a,z)$ on train task distribution. Then, it uses the task distribution model to imagine different shifted test task distributions on which it learns different meta-policies $\{\pi_{\text{meta}}^{\epsilon_i}\}_{i=2}^M$, each corresponding to a different level of robustness. During meta-test, it chooses an appropriate meta-policy based on inferred test task distribution shift with Thompson's sampling and then quickly adapts the selected meta-policy to individual tasks.}
    \label{fig:robust_meta_rl_train}
\end{figure}

% % [Discuss how we introduce distributional robustness with test time adaptation, and discusss why this does something sensible ]
% However, choosing an appropriate uncertainty set can be quite challenging without further information about the test environment, and can significantly impact the performance of these algorithms under distribution shift. If the uncertainty set is too large, the resulting behavior is overly conservative and leads to very slow adaptation at test time, whereas if the uncertainty set is too limited, the algorithm is very susceptible to distribution shift at test time. \pulkit{This paragraph is great :) -- to the point}
% A choice of meta-RL algorithm that is robust to a variety of distribution shift requires a more nuanced choice. 
% \begin{wrapfigure}{r}{0.4\linewidth}
%     \centering
%     \includegraphics[width=\linewidth]{figs/placeholder.jpg}
%     \caption{Illustrative Example of Problem Setting}
%     \label{fig:illustrative_example}
% \end{wrapfigure}
Our key insight is that we can prepare for a variety of potential test-time distribution shifts by constructing and training against different uncertainty sets at training time. By preparing for adaptation against each of these uncertainty sets, an agent is able to adapt to a variety of potential test-time distribution shifts by adaptively choosing the most appropriate level of distributional robustness for the test distribution at hand. We introduce a conceptual framework called distributionally adaptive meta reinforcement learning, formalizing this idea. At train time, the agent learns robust meta-policies with widening uncertainty sets, preemptively accounting for different levels of test-time distribution shift that may be encountered. At test time, the agent infers the level of distribution shift it is faced with, and then uses the corresponding meta-policy to adapt to the new task (Figure~\ref{fig:robust_meta_rl_train}). In doing so, the agent can adaptively choose the best level of robustness for the test-time task distribution, preserving the fast adaptation benefits of meta RL, while also ensuring good asymptotic performance under distribution shift. We instantiate a practical algorithm in this framework (\rml), using learned generative models to imagine new task distributions close to the provided training tasks that can be used to train robust meta-policies.

The contribution of this paper is to propose a framework for making meta-reinforcement learning resilient to a variety of task distribution shifts, and \rml, a practical algorithm instantiating the framework. \rml~trains a population of meta-policies to be robust to different degrees of distribution shifts and then adaptively chooses a meta-policy to deploy based on the inferred test-time distribution shift. Our experiments verify the utility of adaptive distributional robustness under test-time task distribution shift in a number of simulated robotics domains.  

\section{Related Work}
\label{sec:related-work}
\vspace{-10pt}
% Our work builds on a long line of work on distributional robustness and  meta-reinforcement learning, but proposes a novel perspective on building meta-reinforcement learning algorithms that are robust to distribution shift. 
% [meta-RL algorithms]: MAML, Rl2, SNAIL, ProMP, pearl, MAESN, VariBad, Borel, louis kirsch algorithms, RMA. 
Meta-reinforcement learning algorithms aim to leverage a distribution of training tasks to ``learn a reinforcement learning algorithm", that is able to learn as quickly on new tasks drawn from the same distribution. A variety of algorithms have been proposed for meta-RL, including memory-based ~\cite{duan2016rl, mishra2017simple}, gradient-based~\cite{finn2017model, rothfuss2018promp, gupta2018meta} and latent-variable based~\cite{rakelly2019efficient, zintgraf2019varibad, zhao2020meld, fu2021towards} schemes. These algorithms show the ability to generalize to new tasks drawn from the same distribution, and have been applied to problems ranging from robotics~\cite{nagabandi2018learning, zhao2020meld, kumar2021rma} to computer science education~\cite{wu2021prototransformer}. This line of work has been extended to operate in scenarios without requiring any pre-specified task distribution ~\cite{gupta2018unsupervised, jabri2019unsupervised}, in offline settings ~\cite{dorfman2020offline, nair2020awac, mitchell2021offline} or in hard (meta-)exploration settings~\cite{zintgraf2021exploration, zhang2021metacure}, making them more broadly applicable to a wider class of problems. However, most meta-RL algorithms assume source and target tasks are drawn from the same distribution, an assumption rarely met in practice. Our work shows how the machinery of meta-RL can be made compatible with distribution shift at test time, using  ideas from distributional robustness. Some recent work shows that model based meta-reinforcement learning can be made to be robust to a particular level distribution shift ~\cite{mendonca2020mier, lin2021robustmeta} by learning a shared dynamics model against adversarially chosen task distributions. We show that we can build model-free meta-reinforcement learning algorithms, which are not just robust to a particular level of distribution shift, but can adapt to various levels of shift.

% [distributional robustness]: duchi,  namkoong, liang, wilds, 
Distributional robustness methods have been studied in the context of building supervised learning systems that are robust to the test distribution being different than the training one. The key idea is to train a model to not just minimize empirical risk, but instead learn a model that has the lowest worst-case empirical risk among an ``uncertainty-set" of distributions that are boundedly close to the empirical training distribution~\citep{sinha2017certifying, madry2017towards, cohen2019certified, hong2021federated}. If the uncertainty set and optimization are chosen carefully, these methods have been shown to obtain models that are robust to small amounts of distribution shift at test time~\citep{sinha2017certifying, madry2017towards, cohen2019certified, hong2021federated}, finding applications in problems like federated learning~\citep{hong2021federated} and image classification~\citep{madry2017towards}. This has been extended to the min-max robustness setting for specific algorithms like model-agnostic meta-learning ~\cite{collins2020trmaml}, but are critically dependent on correct specification of the appropriate uncertainty set and applicable primarily in supervised learning settings. Alternatively, several RL techniques aim to directly tackle the robustness problem, aiming to learn policies robust to adversarial perturbations~\cite{vinitsky2020robust, zhang2021robust, pinto2017rarl, oikarinen2021robbust}. \cite{xie2022robust} conditions the policy on uncertainty sets to make it robust to different perturbation sets. While these methods are able to learn conservative, robust policies, they are unable to adapt to new tasks as \rml~does in the meta-reinforcement learning setting. 
% A closely related class of methods do adversarial perturbation at test time to find adversarial examples for the model~\citep{madry2017towards, cohen2019certified, Xie_2019_CVPR}, incorporating them into the training process to induce robustness. 
% However, in both these classes of methods, the parameterization of the uncertainty set needs to be chosen carefully, and the level of uncertainty to be robust over has to chosen exactly~\citep{madry2017towards, cohen2019certified}. 
In our work, rather than choosing a single uncertainty set, we learn many meta-policies for widening uncertainty sets, thereby accounting for different levels of test-time distribution shift. 
% During test-time, we infer the level of distribution shiftand select the appropriate meta-learned policy. 
% [robust-control]: 
% TODO
% [generalization in RL]:  
% Our work is also closely related to the problem of \textbf{generalization in RL}, across rewards and dynamics. Operating in the multi-task setting involves learning a single policy that can quickly (or directly adapt) to a new scenario with limited experience~\citep{gupta2019relay, lynch2020learning, choi2021variational}. These methods would typically assume that the task is specified by a known context or try to infer this from a small amount of experience in the environment. Given this context, they aim to learn policies that generalize at test time, but really only to tasks drawn from the same distribution as training time~\citep{gupta2019relay, lynch2020learning, choi2021variational}. In our work, we show how this assumption can be relaxed and tasks drawn from different distributions can be learned quickly as well. This makes these methods much more applicable to problems of real world significance, where there is always likely to be  distribution shift~\citep{miki2022learning, filos2020can}.

% epistemic POMDP, successor features, domain randomization, etc.

% Explain how we are robust, meta and generalize to distribution shift not just in distribution.

\section{Preliminaries}
\label{sec:prelim}

\textbf{Meta-Reinforcement Learning} aims to learn a fast reinforcement learning algorithm or a ``meta-policy" that can quickly maximize performance on tasks $\gT$ from some distribution $p(\mathcal{T})$.  Formally, each task $\gT$ is a Markov decision process (MDP) $\mathcal{M} = (\mathcal{S}, \mathcal{A}, \mathcal{P}, \mathcal{R}, \gamma, \mu_0)$; the goal is to exploit regularities in the structure of rewards and environment dynamics across tasks in $p(\gT)$ to acquire effective exploration and adaptation mechanisms that enable learning on new tasks much faster than learning the task naively from scratch. A meta-policy (or fast learning algorithm) $\pi_\meta$ maps a history of environment experience $h \in (\gS \times \gA \times \gR)^*$ in a new task to an action $a$, and is trained to acquire optimal behaviors on tasks from $p(\gT)$ within $k$ episodes: 
% over MDPs to %a \emph{fast} reinforcement learning algorithm $\mathrm{Adapt}:\Pi,\gM \rightarrow \Pi$, that can quickly learn optimal policies in new MDPs drawn from a testing task distribution $p_{\text{test}}(\mathcal{T})$, which is typically assumed to be the same as $p_{\text{train}}(\mathcal{T})$. 
% \pulkit{Is this actually true? This would be the case if we were doing RL2 or RMA -- but our optimization is quite different. This can confuse a reader.} 
% The goal is to exploit regularities in the training tasks to acquire learning strategies for quick task adaptation, rather than having to learn a new task from scratch. Most meta-reinforcement learning algorithms can be structured as some version of a bi-level optimization problem, where the goal is to use an outer loop ``slow" optimization algorithm (such as policy gradient) to learn parameterized inner loop update rules that can perform very quick and performant updates. The objective function optimized during \textit{meta-training} is:
\begin{align}
     &\min_{\pi_{\text{meta}}} \E_{\gT \sim p(\gT)}\left[\regret(\pi_{\text{meta}}, \gT)\right]\nonumber,\\
     \regret(\pi_\meta, \gT) &= J(\pi_\gT^*) - \E_{a_t^{(i)}\sim\pi_\meta(\cdot|h_t^{(i)}), \gT}\left[\frac{1}{k}\sum_{i=1}^k \sum_{t=1}^T r_t^{(i)}\right], \;\; J(\pi_\gT^*) = \max_\pi \E_{\pi,\gT}[\sum_t r_t] \nonumber\\ 
     \;\text{where}\; &h_t^{(i)} = (s_{1:t}^{(i)}, r_{1:t}^{(i)}, a_{1:t-1}^{(i)}) \cup (s_{1:T}^{(j)}, r_{1:T}^{(j)}, a_{1:T}^{(j)} )_{j=1}^{i-1}.
     % \vspace{-10pt}
\label{eq:meta_train}
\end{align}
Intuitively, the meta-policy has two components: an exploration mechanism that ensures that appropriate reward signal is found for all tasks in the training distribution, and an adaptation mechanism that uses the collected exploratory data to generate optimal actions for the current task. In practice, the meta-policy may be represented explicitly as an exploration policy conjoined with a policy update\citep{finn2017model,  rakelly2019efficient}, or implicitly as a black-box RNN \citep{duan2016rl, zintgraf2019varibad}. We use the terminology ``meta-policies" interchangeably with that of ``fast-adaptation" algorithms, since our practical implementation builds on ~\cite{ni2022recurrent} (which represents the adaptation mechanism using a black-box RNN). 
% The same framework can apply to any meta-learning algorithm, including those based on gradient based updates ~\cite{finn2017model, gupta2018meta} or latent variable inference ~\cite{rakelly2019efficient, zintgraf2019varibad}. 
Our work focuses on the setting where there is potential drift between $p_\train(\gT$), the task distribution we have access to during training, and $p_\testt(\gT)$, the task distribution of interest during evaluation.
\textbf{Distributional robustness} ~\citep{sinha2017certifying} learns models that do not minimize empirical risk against the training distribution, but instead prepare for distribution shift by optimizing the \emph{worst-case} empirical risk against a set of data distributions close to the training distribution (called an \emph{uncertainty set}):
% \pulkit{This is good -- directly come to the point}. 
% This results in the following optimization: 
% \pulkit{parameterization of what? What is the distribution being defined over? Its a bit vague} 
% TODO: Describe the min-max robustness setting 
\begin{align}
    \min_\theta &~\max_\phi \E_{x \sim q_\phi(x)}[l(x; \theta)]
    ~~~~~~\text{s.t.}~~~D(p_{\text{train}}(x) || q_\phi(x)) \leq \epsilon
\end{align}
% % TODO: Describe. how the chocie of uncertainty set etc plays a role. 
This optimization finds the model parameters $\theta$ that minimizes worst case risk $l$ over distributions $q_\phi(x)$ in an $\epsilon$-ball (measured by an $f$-divergence) from the training distribution $p_{\text{train}}(x)$. 
% We build on this framework to construct distributionally adaptive meta-learners as we discuss next.
% However, it is unclear how to parameterize $q_\phi$ over MDPs. In the following section, we propose a framework for distributionally adaptive meta-RL and then instantiate it with a practical algorithm \rml~while showing how to appropriately parameterize uncertainty sets over MDPs.
% Note that this problem is still underdefined, the choice of parameterization for $q_\phi$ is still unclear, especially when lifted to problems like Markov decision processes, where the appropriate measure is not immediately clear\pulkit{What is lifted to means? Maybe put it simply that how  to define distributions over MDPs is unclear?}. Moreover, as currently stated, the distributional robustness framework crucially depends on the choice of uncertainty set parameterization $q_\phi$ and the choice of $\epsilon$ as well. These choices become important when building distributionally robust meta-RL algorithms. In the following section, we instantiate an algorithm for robust meta-reinforcement learning that is robust to varying levels of meta test-time distribution shift, and show how to parameterize uncertainty sets appropriately for reinforcement learning. 

\section{Distributionally Adaptive Meta-Reinforcement Learning}
\label{sec:approach}
\begin{figure}[!t]
    \centering
    \includegraphics[width=0.95\linewidth]{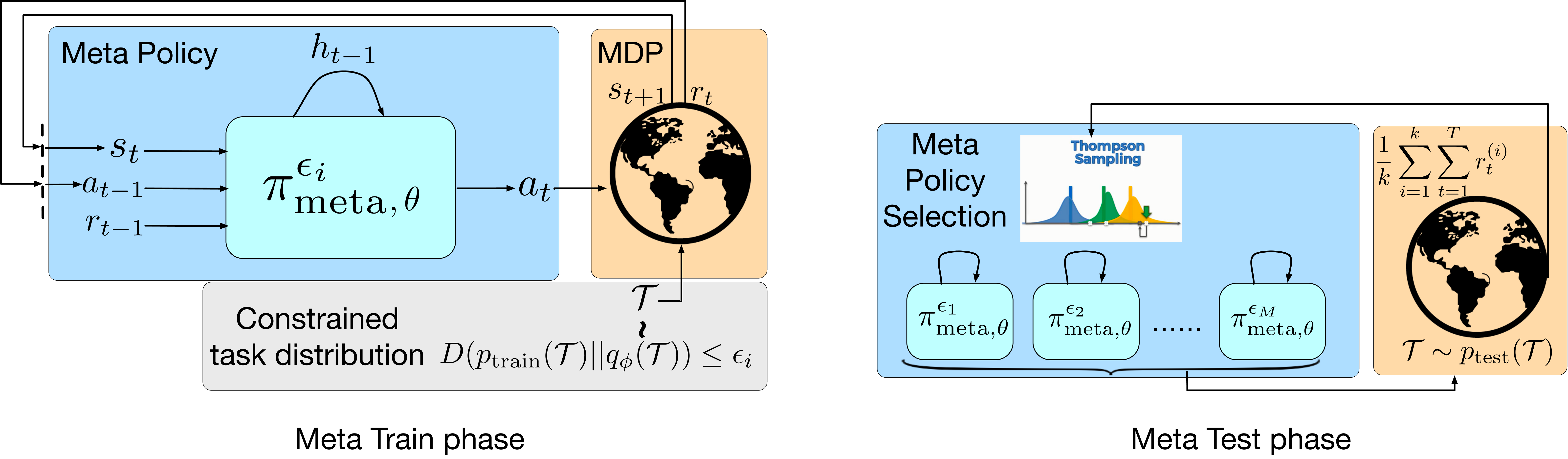}
    \caption{During meta-train phase, \rml~learns a family of meta-policies robust to varying levels of distribution shift (as characterized by $\epsilon_i$). During meta-test phase, given a potentially shifted test-time distribution of tasks, \rml~chooses the meta-policy with the most appropriate level of robustness and use it to perform fast adaptation for new tasks sampled from the same shifted test task distribution.}
    \label{fig:diametr_diagram}
\end{figure}

 In this section, we develop a framework for learning meta-policies, that given access to a training distribution of tasks $p_{\text{train}}(\mathcal{T})$, is still able to adapt to tasks from a test-time distribution $p_{\testt}(\mathcal{T})$ that is similar but not identical to the training distribution. We introduce a framework for distributionally adaptive meta-RL below and instantiate it as a practical method in Section~\ref{sec:practical}.

\subsection{Known Level of Test-Time Distribution Shift}

We begin by studying a simplified problem where we can exactly quantify the degree to which the test distribution deviates from the training distribution. Suppose we know that $p_\testt$ satisfies $D(p_\testt(\gT) || p_\train(\gT)) < \epsilon$ for some $\epsilon > 0$, where $D(\cdot \| \cdot)$ is a probability divergence on the set of task distributions (e.g. an $f$-divergence~\citep{renyi1961measures} or a Wasserstein distance~\citep{vaserstein1969markov}). A natural learning objective to learn a meta-policy under this assumption is to minimize the \textit{worst-case} test-time regret across any test task distribution $q(\gT)$ that is within some $\epsilon$ divergence of the train distribution:
\begin{align}
     \min_{\pi_{\text{meta}}} ~&\gR(\pi_{\text{meta}}, p_{\train}(\gT), \epsilon)\nonumber,\\
     \gR(\pi_{\text{meta}}, p_{\train}(\gT), \epsilon) &= \max_{q(\gT)} \E_{\gT \sim q(\gT)}\left[\regret(\pi_{\text{meta}}, \gT)\right]~~~~~~\text{s.t.}~~D(p_{\train}(\gT) \| q(\gT)) \leq \epsilon
\label{eq:robust_meta_train}
\end{align}
% \begin{align}
%      \min_{\pi_{\text{meta}}} &\max_{q(\gT)} \E_{\gT \sim q(\gT)}\left[\regret(\pi_{\text{meta}}, \gT)\right]~~~~~~\text{s.t.}~~D(q(\gT) || p_{\train}(\gT)) \leq \epsilon
% \label{eq:robust_meta_train}
% \end{align}

Solving this optimization problem results in a meta-policy that has been trained to adapt to tasks from a \textit{wider} task distribution than the original training distribution. It is worthwhile distinguishing this robust meta-objective, which incentivizes a \textit{robust adaptation mechanism} to a wider set of tasks, from robust objectives in standard RL, which produce base policies robust to a wider set of dynamics conditions. The objective in Eq \ref{eq:robust_meta_train} incentivizes an agent to explore and adapt more broadly, not act more conservatively as standard robust RL methods ~\citep{pinto2017rarl} would encourage. Naturally, the quality of the robust meta-policy depends on the size of the uncertainty set. If $\epsilon$ is large, or the geometry of the divergence poorly reflect natural task variations, then the robust policy will have to adapt to an overly large set of tasks, potentially degrading the speed of adaptation. 

 %Since the adversarial choice of task distribution focuses on tasks near the training distribution that are difficult for the current policy, the learned meta-policy must perform well across all such nearby tasks, which is the desired objective from our robust objective.

% The optimization proposed in~\ref{eq:robust_meta_train} is critically dependent on the choice of parameterization of uncertainty set $q$ and the divergence metric $D$, which are non-trivial to define. We next describe how to propose uncertainty sets for this distributionally robust optimization via generative modeling.
% This goes to last section
% In this work, we use off-policy $\text{RL}^2$ to train the meta-RL agent which parameterizes $\mathrm{Adapt}$ with a recurrent neural network and use soft-actor critic (SAC)~\citep{haarnoja2018soft} for outer-loop optimization \pulkit{the outerloop is the distribution proposer?}. For brevity, we refer to off-policy $\text{RL}^2$ as just $\text{RL}^2$. We use $1$-step proximal policy optimization (PPO)~\citep{schulman2017ppo} to train the distribution proposer \pulkit{ Its unclear why it needs to be off-policy? Also there seems to be three algorithms -- PPO, SAC and RL2 .. where each one is used is unclear. For each one of them detail things ... atleast the inputs and outputs. Why can't I just sample the worst case task using something like CMAES and then just optimize RL2 on that task. Why is SAC needed?}.
 
\subsection{Handling Arbitrary Levels of Distribution Shift}
\label{subsec:train_multi_robust}
% Discuss why this may be insufficient, instead we can learn over a distribution over various epsilons. Intuition, preparing for variable distribution shift at test time. 
% Describe what this looks like (how are epsilons sampled), are we doing discrete/continuous learning of meta-learned policies. 
% The framework described in Section~\ref{subsec:dmrl} shows how to build meta-RL algorithms robust to a particular level of distribution shift characterized by a particular choice of $\epsilon$. If $\epsilon$ is chosen to be too large, the resulting behavior is overly conservative, making learning very slow and if it is chosen to be too small, the resulting behavior can fail to explore sufficiently, making test time regret infinite. 
In practice, it is not known how the test distribution $p_\testt$ deviates from the training distribution, and consequently it is challenging to determine what $\epsilon$ to use in the meta-robustness objective. We propose to overcome this via an adaptive strategy: to train meta-policies for \textit{varying} degrees of distribution shift, and at test-time, inferring which distribution shift is most appropriate through experience.

We train a population of meta-policies $\{\pi_\meta^{(i)}\}_{i=1}^M$, each solving the distributionally robust meta-RL objective (eq~\ref{eq:robust_meta_train}) for a different level of robustness $\epsilon_i$:
\begin{equation}
    \left\{\pi_{\meta}^{\epsilon_i} \coloneqq \arg\min_{\pi_\meta} \gR(\pi_\meta, p_\train(\gT), \epsilon_i)\right\}_{i=1}^M ~~~\text{ where } \epsilon_M > \epsilon_{M-1} > \ldots > \epsilon_1 = 0
\end{equation}
% $\mathrm{Adapt}(\pi_{\theta_1, \epsilon_1}, \sbullet; k)$ is the \textit{base} fast learning algorithm that just minimizes test-time regret over train task distribution $p_{\text{train}}(\mathcal{T})$. 
In choosing a spectrum of $\epsilon_i$, we learn a set of meta-policies that have been trained on increasingly large set of tasks: at one end ($i=1$), the meta-policy is trained only on the original training distribution, and at the other ($i=M$), the meta-policy trained to adapt to any possible task within the parametric family of tasks. These policies span a tradeoff between being robust to a wider set of task distributions with larger $\epsilon$ (allowing for larger distribution shifts), and being able to adapt quickly to any given task with smaller $\epsilon$ (allowing for better per-task regret minimization).

With a set of meta-policies in hand, we must now decide how to leverage test-time experience to discover the right one to use for the actual test distribution $p_\testt$. We recognize that the problem of policy selection can be treated as a stochastic multi-armed bandit problem (precise formulation in Appendix~\ref{sec:test_time_select}), where pulling arm $i$ corresponds to running the meta-policy $\pi_\meta^{\epsilon_i}$ for an entire meta-episode ($k$ task episodes). If a zero-regret bandit algorithm (eg: Thompson's sampling~\citep{585893}) is used , then after a certain number of test-time meta episodes, we can guarantee that the meta-policy selection mechanism will converge to the meta-policy that best balances the tradeoff between adapting quickly while still being able to adapt to all the tasks from $p_\testt(\gT)$.

To summarize our framework for distributionally adaptive meta-RL, we train a population of meta-policies at varying levels of robustness on a distributionally robust objective that forces the learned adaptation mechanism to also be robust to tasks not in the training task distribution. At test-time, we use a bandit algorithm to select the meta-policy whose adaptation mechanism has the best tradeoff between robustness and speed of adaptation specifically on the test task distribution. Combining distributional robustness with test-time adaptation allows the adaptation mechanism to work even if distribution shift is present, while obviating the decreased performance that usually accompanies overly conservative, distributionally robust solutions. 
% \begin{figure}[!t]
%     \centering
%     \includegraphics[width=0.65\linewidth]{figs/robust_meta_rl_test.pdf}
%     % \includegraphics[width=0.5\linewidth]{figs/robust_meta_rl_test.pdf}
%     \caption{\rml~chooses appropriate meta-policy based on inferred distribution shift with Thompson's sampling and then quickly adapts the selected meta-policy to individual tasks during meta-test.
%     % \pulkit{Maybe color the plots in thompson sampling to corredpond with the color of dots in the three circles on the left. Also lets explain whats happening on the right -- I am confused by the arrow that goes from the sprite's hand to the dot -- what is it supposed to indicate? Its unclear to me what different cocnentric circles means. I understand that the concentric arrow is the exploration path.}
%     }
%     % \pulkit{Text is too small in the figure .. its unclear what is happening in the right most sub-figure. Is there an intended difference between the rightmost two circles?}}
%     \label{fig:robust_meta_rl_test}
% \end{figure}
\subsection{Analysis}
To provide some intuition on the properties of this algorithm, we formally analyze adaptive distributional robustness in a simplified meta RL problem involving tasks $\gT_g$ corresponding to reaching some unknown goal $g$ in a deterministic MDP $\gM$, exactly at the final timestep of an episode. We assume that all goals are reachable, and use the family of meta-policies that use a stochastic exploratory policy $\pi$ until the goal is discovered and return to the discovered goal in all future episodes. The performance of a meta-policy on a task $\gT_g$ under this model can be expressed in terms of the state distribution of the exploratory policy: $\regret(\pi_\meta, \gT_g) = \frac{1}{d_\pi^T(g)}$. This particular framework has been studied in ~\cite{gupta2018unsupervised, lee2019smm}, and is a simple, interpretable framework for analysis. 

We seek to understand performance under distribution shift when the original training task distribution is relatively concentrated on a subset of possible tasks. We choose the training distribution $p_\train(\gT_g) = (1-\beta) \text{Uniform}(\gS_0) + \beta \text{Uniform}(\gS \backslash \gS_0)$, so that $p_\train$ is concentrated on tasks involving a subset of the state space $\gS_0 \subset \gS$, with $\beta$ a parameter dictating the level of concentration, and consider test distributions that perturb under the TV metric. Our main result compares the performance of a meta-policy trained to an $\epsilon_2$-level of robustness when the true test distribution deviates by $\epsilon_1$.\\
\begin{restatable}{prop}{tvrobustness}
\vspace{-0.3cm}
Let $\bar{\epsilon_i} = \min\{\epsilon_i + \beta, 1 - \frac{|\gS_0|}{|\gS|}\}$. There exists $q(\gT)$ satisfying $D_{TV}(p_\train, q) \leq \epsilon_1$ where an $\epsilon_2$-robust meta policy incurs excess regret over the optimal $\epsilon_1$-robust meta-policy: %$\pi_{\meta}^\epsilon = \argmin \gR(\pi_\meta, p_\train, \epsilon)$ and $\pi_{\meta}^{\epsilon^*} = \argmin \gR(\pi_\meta, p_\train, \epsilon^*)$. 
\begin{align}
    \E_{q(\gT)}[\operatorname{Regret}(\pi_{\meta}^{\epsilon_2}, \gT) - \operatorname{Regret}(\pi_{\meta}^{\epsilon_1}, \gT)] \geq  ~\left(c(\epsilon_1, \epsilon_2) + \frac{1}{c(\epsilon_1, \epsilon_2)} - 2\right)\\
    ~\sqrt{\bar{\epsilon_1}(1 - \bar{\epsilon_1})|\gS_0|(|\gS| - \gS_0|)}
\end{align}
The scale of regret depends on $c(\epsilon_1, \epsilon_2) = \sqrt{\tfrac{\bar{\epsilon_2}^{-1} - 1}{\bar{\epsilon_1}^{-1} - 1}}$, a measure of the mismatch between $\epsilon_1$ and $\epsilon_2$.
% \left(\sqrt{1 + \frac{\epsilon}{\beta}} + \frac{1}{\sqrt{1 + \frac{\epsilon}{\beta}}} - 2\right)
\end{restatable}
% The constant $c(\epsilon_1, \epsilon_2)$ quantifies the scale of excess regret in this bound, leading to no regret if $c = 1$ (e.g when $\epsilon_1 = \epsilon_2$) and higher regret as $c$ deviates in either direction. 

We first compare robust and non-robust solutions by analyzing the bound when $\epsilon_2 = 0$. In the regime of $\beta \ll 1$, excess regret scales as $\gO(\epsilon_1\sqrt{\tfrac{1}{\beta}})$, meaning that the robust solution is most necessary when the training distribution is highly concentrated in a subset of the task space. At one extreme, if the training distribution contains no examples of tasks outside $\gS_0$ ($\beta = 0$), the non-robust solution incurs \emph{infinite excess regret}; at the other extreme, if the training distribution is uniform on the set of all possible tasks ($\beta = 1 - \tfrac{|\gS_0|}{|\gS|}$), \emph{robustness provides no benefit}.  
% \begin{takeaway}
% The excess regret of a non-robust policy under distribution shift ranges from $\infty$ ($p_\train$ concentrated on a subset of tasks) to zero ($p_\train$ uniformly covers all possible tasks)
% \end{takeaway}

We next quantify the effect of mis-specifying the level of robustness in the meta-robustness objective, and what benefits \textit{adaptive} distributional robustness can confer. For small $\beta$ and fixed $\epsilon_1$, the excess regret of an $\epsilon_2$-robust policy scales as $\gO(\sqrt{\max\{\tfrac{\epsilon_2}{\epsilon_1}, \tfrac{\epsilon_1}{\epsilon_2}\}})$, meaning that excess regret gets incurred if the meta-policy is trained either to be too robust $(\epsilon_2 \gg \epsilon_1)$ or not robust enough $\epsilon_1 \gg \epsilon_2$. Compared to a fixed robustness level, our strategy of training meta-policies for a sequence of robustness levels $\{\epsilon_i\}_{i=1}^M$ ensures that this misspecification constant is at most the relative spacing between robustness levels: $\max_{i} \frac{\epsilon_i}{\epsilon_{i-1}}$. This enables the distributionally adaptive approach to \emph{control} the amount of excess regret by making the sequence more fine-grained, while a fixed choice of robustness incurs larger regret (as we verify empirically in our experiments as well). 

\begin{algorithm}[t]
    \small
    \caption{\textbf{\rml}:Meta-training phase}
    \label{alg:meta_train_code}
    \begin{algorithmic}[1]
    \STATE Given: $p_{\text{train}}(\gT)$, Return: $\{\pi_{\text{meta}, \theta}^{\epsilon_i}\}_{i=1}^M$
    \STATE $\pi_{\text{meta}, \theta}^{\epsilon_1}$, $\gD_{\text{Replay-Buffer}} \leftarrow$ Solve equation~\ref{eq:meta_train} with off-policy $\text{RL}^2$
    \STATE Prior $p_\train(\gT) \leftarrow$ Solve eq~\ref{eq:prac_vae_main} using $\gD_{\text{Replay-Buffer}}$ 
    \FOR{$\epsilon$ in $\{\epsilon_2,\ldots,\epsilon_M\}$}
        \STATE Initialize $q_\phi$, $\pi_{\text{meta}, \theta}^\epsilon$ and $\lambda \geq 0$.
      	\FOR{iteration $n=1, 2, ...$}
      	    \STATE \textbf{Meta-policy:} Update $\pi_{\text{meta},\theta}^\epsilon$ using off-policy $\text{RL}^2$~\cite{ni2022recurrent}
                \begin{equation}
                    \theta \coloneqq \theta + \alpha\nabla_\theta \E_{\gT \sim q_\phi(\gT)}[\E_{\pi_{\text{meta},\theta}^\epsilon,\gP_\gT}[\frac{1}{k}\sum_{i=1}^k\sum_{t=1}^T r_\gT(s_t^{(i)}, a_t^{(i)})]]\nonumber
                \end{equation}
            \STATE \textbf{Adversarial task distribution:} Update $q_\phi$ using Reinforce~\cite{sutton1999policy}
            \begin{equation}
            \phi \coloneqq \phi - \alpha\nabla_\phi (\E_{\gT \sim q_\phi(\gT)}[\E_{\pi_{\text{meta},\theta}^\epsilon,\gP_\gT}[\frac{1}{k}\sum_{i=1}^k\sum_{t=1}^T r_\gT(s_t^{(i)}, a_t^{(i)})]] + \lambda D_{\text{KL}}(p_\train(\gT)\|q_\phi(\gT)))\nonumber
            \end{equation}
      	\STATE \textbf{Lagrange constraint multiplier:} Update $\lambda$ to enforce $D_{\text{KL}}(p_\train \|q_\phi) < \epsilon$,
      	    \begin{equation}
                \lambda \coloneqq_{\lambda \geq 0} \lambda + \alpha (D_{\text{KL}}(p_\train(\gT)\|q_\phi(\gT)) - \epsilon)\nonumber
            \end{equation}
            \vspace{-0.5cm}
      	\ENDFOR
    %   	\STATE Add $\pi_{\text{meta}, \theta}^\epsilon$ to $\Pi$
    \ENDFOR
    \end{algorithmic}
    % \vspace{-0.5cm}
\end{algorithm}

\section{\rml: A Practical Algorithm for Meta-Distribution Shift}
\label{sec:practical}
In order to instantiate our distributionally adaptive framework into a practical algorithm, we must address how task distributions should be parameterized and optimized over . We must also address how the robust meta-RL problem can be solved with stochastic gradient methods. We first introduce the individual components of task parameterization and robust optimization, describe the overall algorithm in Algorithm~\ref{alg:meta_train_code} and ~\ref{alg:meta_test_code}, and visualize components of \rml~in Fig~\ref{fig:diametr_diagram}. 

\subsection{Parameterizing Task Distributions}
\textbf{\textcolor{blue}{Handling in-support distribution shifts:}} For handling in-support task distribution shifts, we propose to represent new task distributions as re-weighted training task distribution $q(\gT) \propto w_\gT p_\train(\gT)$ where $w_\gT > 0$ is a parameter. Since we have a finite set of training tasks, say $\{\gT_i\}_{i=1}^{n_\text{tr}}$, new task distributions become $q(\gT_i) = \frac{w_{\gT_i}}{\sum_{i=1}^{n_\text{tr}} w_{\gT_i}}$. With a slight abuse of notation, we can write empirical training task distribution as $p_\train(\gT_i) = \frac{1}{n_\text{tr}}$. We can use KL divergence to measure the divergence between training task distribution and test task distribution $D(p_\train(\gT) \| q_\phi(\gT)) = \sum_{i=1}^{n_\text{tr}} \frac{1}{n_\text{tr}}\log\frac{\sum_{i=1}^{n_\text{tr}} w_{\gT_i}}{n_\text{tr} w_{\gT_i}}$. We collectively represent the parameters $\phi = \{w_{\gT_i}\}_{i=1}^{n_\text{tr}}$. Using this parameterization, the training objective (equation~\ref{eq:robust_meta_train}) becomes
\begin{align}
    \max_\theta &\min_\phi \E_{\gT\sim p_{\train}(\gT)}\left[\E_{\pi_{\text{meta},\theta}^\epsilon,\gP}\left[\frac{nw_\gT}{\sum_{i=1}^{n_{\text{tr}}} w_{\gT_i}}\frac{1}{k}\sum_{i=1}^k\sum_{t=1}^T r_t^{(i)}\right]\right] \nonumber\\
    &~D_{\text{KL}}(p_{\text{train}}(\gT) || q_\phi(\gT)) \leq \epsilon
\label{eq:weight_robust_meta_train}
\end{align}

\textbf{Handling out-of-support distribution shift:} 
For handling out-of-support task distribution shifts, we propose to learn a probabilistic model of the training task distribution, and use the learned latent representation as a space on which to parameterize uncertainty sets over new task distributions. Specifically, we jointly train a task encoder $q_{\psi}(z | h)$ that encodes an environment history into the latent space, and a decoder $\gT_\omega(s, a, z)$ mapping a latent vector $z$ to a property of the task using a dataset of trajectories collected from the training tasks. To describe the exact form of $\gT_\omega$, we consider how tasks can differ and list two scenarios: (1) \textcolor{purple}{Tasks differ in reward functions:} $\gT_\omega$ takes form of reward functions $r_\omega(s, a, z)$ that maps a latent vector $z$ to a reward function and (2) \textcolor{brown}{Tasks differ in dynamics:} $\gT_\omega$ takes form of dynamics $p_\omega(s, a, z)$ that maps a latent vector $z$ to a dynamics function. This generative model can be trained as a standard latent variable model by maximizing a standard evidence lower bound (ELBO), trading off reward prediction and matching a prior $p_\train(z)$ (chosen to be the unit gaussian).
\vspace{-0.25cm}
\begin{align}
    \min_{\omega, \psi} \E_{h\sim\gD}\left[\E_{z \sim q_\psi(z|h)}\left[\sum_{t=1}^T l(\gT_\omega(s_t, a_t, z),h,t)\right] + D_{\text{KL}}(q_\psi(z|h) || \gN(0, I))\right]\nonumber\\
    l(\gT_\omega(s_t, a_t, z),h,t) = \textcolor{purple}{(r_\omega(s_t, a_t, z) - r_t)^2} \;\;\; \text{\textcolor{purple}{when rewards differ}}\nonumber\\
    l(\gT_\omega(s_t, a_t, z),h,t) = \textcolor{brown}{\|p_\omega(s_t, a_t, z) - s_{t+1}\|^2} \;\;\; \text{\textcolor{brown}{when dynamics differ}}
    \label{eq:prac_vae_main}
\end{align}
Having learned a latent space, we can parameterize new task distributions $q(\gT)$ as distributions $q_\phi(z)$ (the original training distribution corresponds to $p_\train(z) = \gN(0, I)$, and measure the divergence between task distributions as well using the KL divergence in this latent space $D(p_\train(z) \| q_\phi(z))$. Using this parameterization, the training objective (equation~\ref{eq:robust_meta_train}) becomes
\begin{align}
     \max_\theta &\min_\phi \E_{z \sim q_\phi(z)}\left[\E_{\pi_{\text{meta},\theta}^\epsilon,\gP}\left[\frac{1}{k}\sum_{i=1}^k\sum_{t=1}^T \textcolor{purple}{r_\omega(s_t^{(i)}, a_t^{(i)},z)}\right]\right] \;\;\; \textcolor{purple}{\text{when rewards differ}}\nonumber\\
    \max_\theta &\min_\phi \E_{z \sim q_\phi(z)}\left[\E_{\pi_{\text{meta},\theta}^\epsilon,\textcolor{brown}{p_\omega(\cdot,\cdot,z)}}\left[\frac{1}{k}\sum_{i=1}^k\sum_{t=1}^T r_t^{(i)}\right]\right] \;\;\; \textcolor{brown}{\text{when dynamics differ}}\nonumber\\
    &~D_{\text{KL}}(p_{\text{train}}(z) || q_\phi(z)) \leq \epsilon
\label{eq:gen_robust_meta_train}
\end{align}

\subsection{Training and test-time selection of meta-policies}
\textbf{Learning Robust Meta-Policies:} Given this task parameterization, the next question becomes how to actually perform the robust optimization laid out in Eq:\ref{eq:robust_meta_train}. The distributional meta-robustness objective can be modelled as an adversarial game between a meta-policy $\pi_{\meta}^\epsilon$ and a task proposal distribution $q(\gT)$. As described above, this task proposal distribution is parameterized as a distribution over latent space $q_\phi(z)$, while $\pi_{\meta}^\epsilon$ is parameterized a typical recurrent neural network policy as in ~\citep{ni2022recurrent}. We parameterize $\{\pi_{\meta}^{\epsilon_i}\}_{i=1}^M$ as a discrete set of meta-policies, with one for each chosen value of $\epsilon$.

This leads to a simple alternating optimization scheme (see Algorithm~\ref{alg:meta_train_code}), where the meta-policy is trained using a standard meta-RL algorithm (we use off-policy \text{RL}$^2$~\citep{ni2022recurrent} as a base learner), and the task proposal distribution with an constrained optimization method (we use dual gradient descent \citep{nesterov2009primal}). Each iteration $n$, three updates are performed: 1) the meta-policy $\pi_{\meta,\theta}$ updated to improve performance on the current task distribution, 2) the task distribution $q_\phi(\gT)$ updated to increase weight on tasks where the current meta-policy adapts poorly and decreases weight on tasks that the current meta-policy can learn, while staying close to the original training distribution, and 3) a penalty coefficient $\lambda$ is updated to ensure that $q_\phi(\gT)$ satisfies the divergence constraint. 

% \begin{wrapfigure}{r}{0.4\textwidth}
% \vspace{-0.5cm}
\begin{algorithm}[H]
    \small
    \caption{\textbf{\rml}: Meta-test phase}
      	\label{alg:test_method}
      	\begin{algorithmic}[1]
      	\STATE Given: $p_{\text{test}}(\gT)$, $\Pi = \{\pi_{\text{meta}, \theta}^{\epsilon_i}\}_{i=1}^M$
      	\STATE Initialize \texttt{TS = Thompson-Sampler()}
      	\FOR{meta-episode $n=1, 2,...$}
      	    \STATE Choose meta-policy $i$ = \texttt{TS.sample()}
      	    \STATE Run $\pi_{\text{meta}, \theta}^{\epsilon_i}$ for meta-episode %($k$ environment episode) 
      	    \STATE \texttt{TS.update(\\
      	   ~~~~arm=i,\\~~~~reward=meta-episode return)}
      	\ENDFOR
    %   	\STATE Return $\pi_{\text{meta}, \theta}^\epsilon \leftarrow$ \texttt{Thompson-Sampler.sample()}
      	\end{algorithmic}
\label{alg:meta_test_code}
\end{algorithm}
% \vspace{-0.5cm}
% \end{wrapfigure}

\textbf{Test-time meta-policy selection:} Since test-time meta-policy selection can be framed as a multi-armed bandit problem, we use a generic Thompson's sampling~\citep{585893} algorithm (see Algorithm~\ref{alg:meta_test_code}). Each meta-episode $n$, we sample a meta-policy $\pi_{\meta}^\epsilon$ with probability proportional to its estimated average episodic reward, run the sampled meta-policy $\pi_{\meta}^\epsilon$ for an meta-episode ($k$ environment episodes) and then update the estimate of the average episodic reward. Since Thompson's sampling is a zero-regret bandit algorithm, it will converge to the meta-policy that achieves the highest average episodic reward and lowest regret on the test task distribution.
\section{Experimental Evaluation}
\label{sec:experiments}
\begin{figure}[t]
    \centering
    \begin{subfigure}[b]{0.24\textwidth}
        \includegraphics[width=\linewidth]{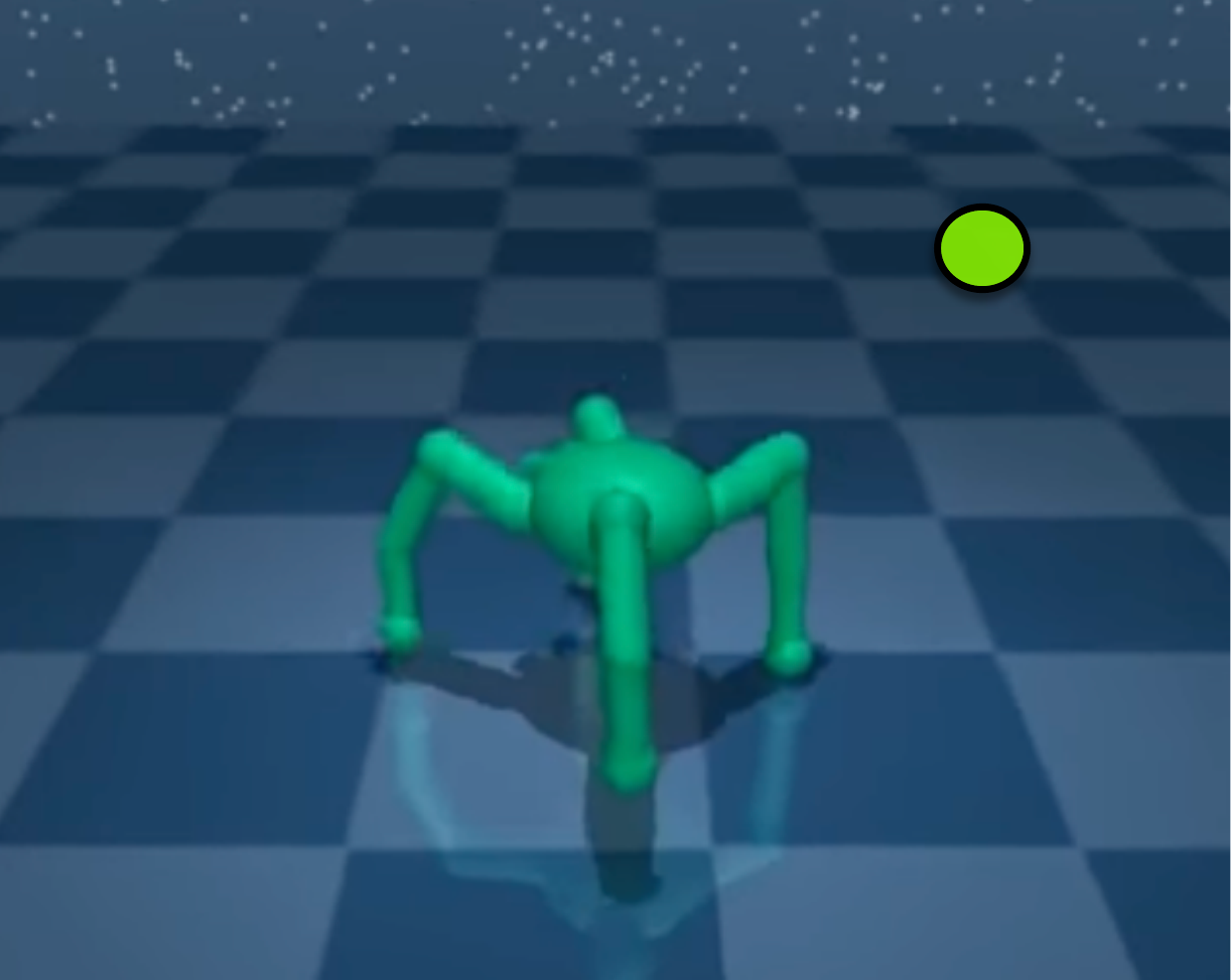}
        \caption{Ant navigation}\label{fig:ant_env}
    \end{subfigure}
    \begin{subfigure}[b]{0.24\textwidth}
        \includegraphics[width=\linewidth]{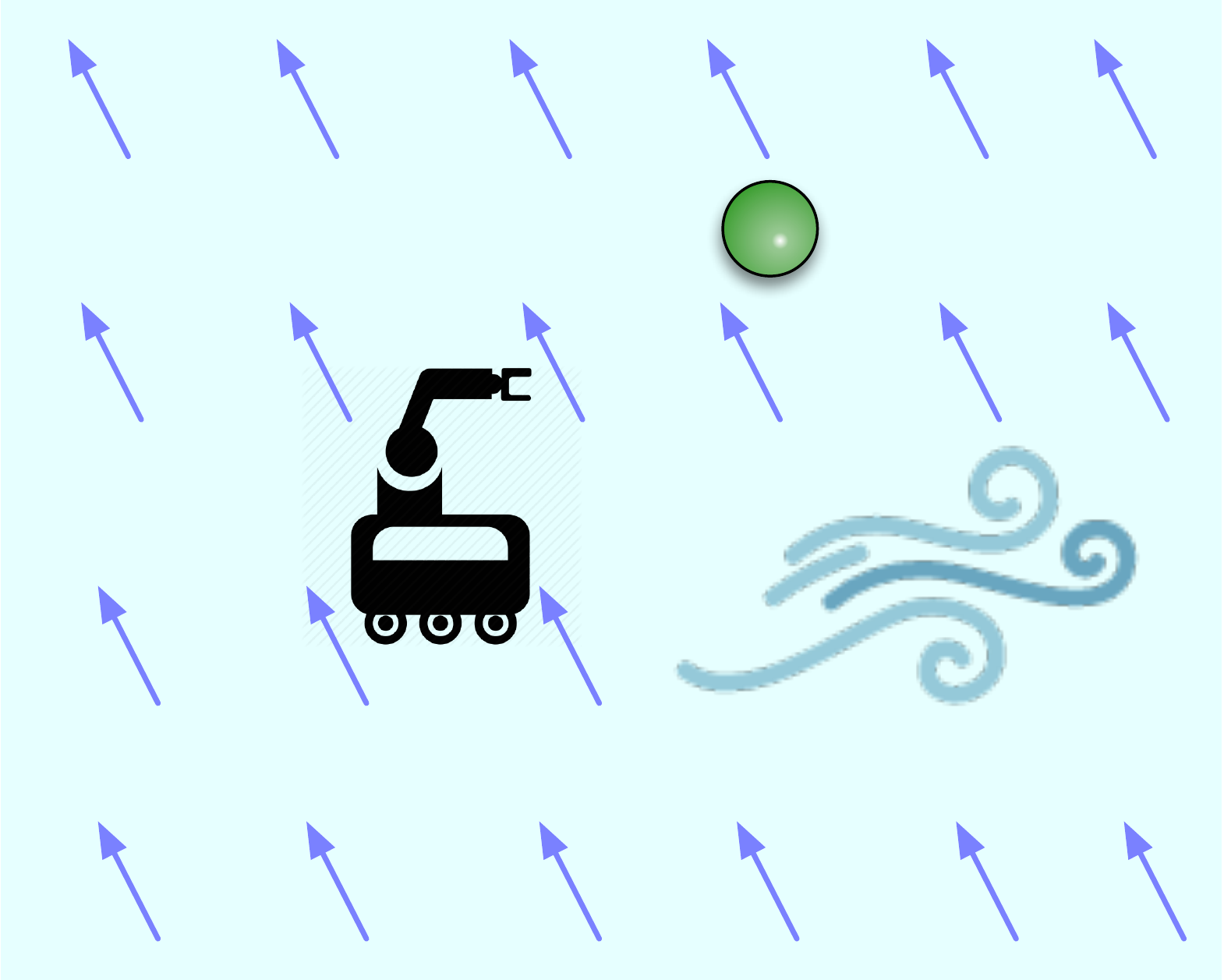}
        \caption{Wind navigation}\label{fig:wind_env}
    \end{subfigure}
    \begin{subfigure}[b]{0.195\textwidth}
        \includegraphics[width=\linewidth]{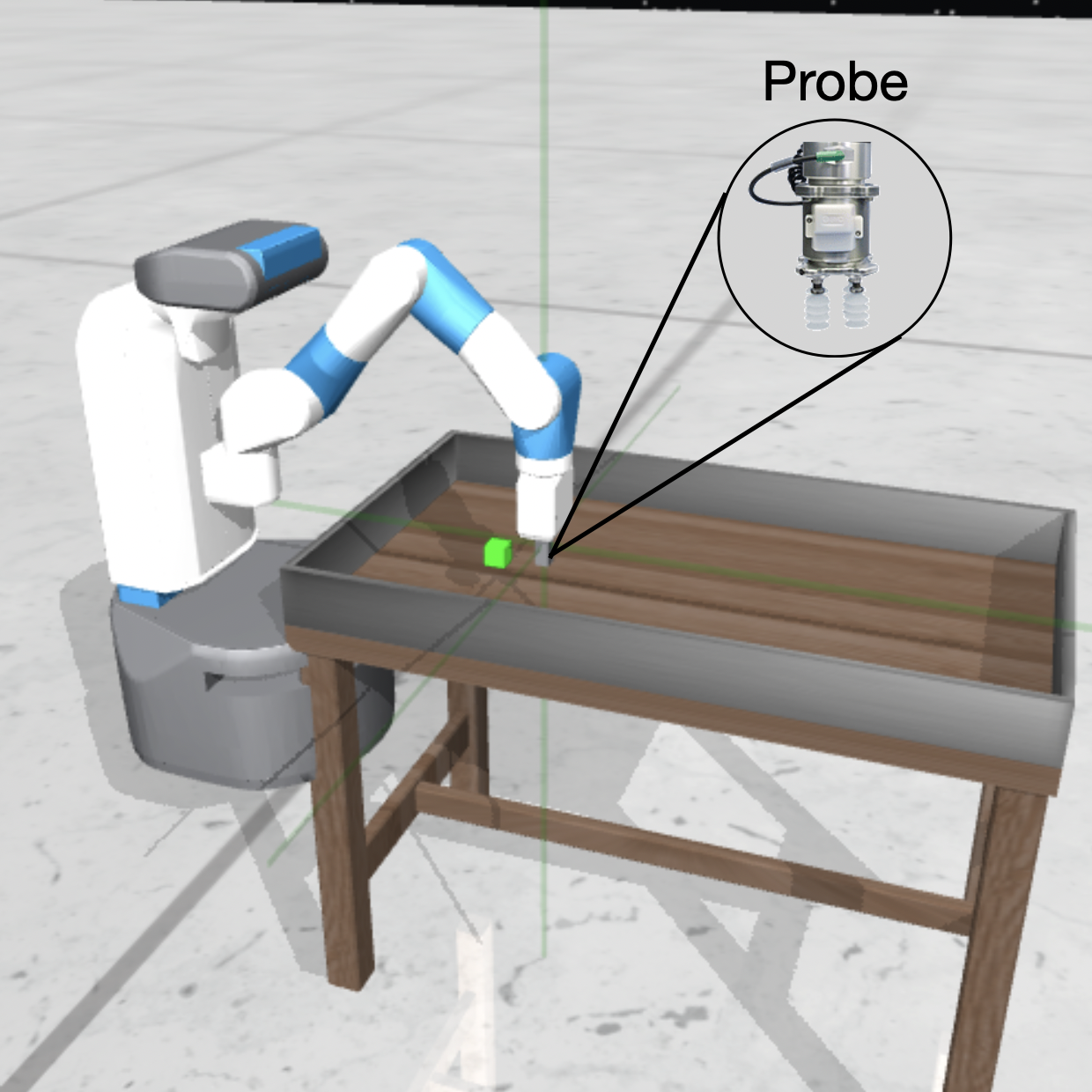}
        \caption{Object localization}\label{fig:fetch_reach}
    \end{subfigure}
    \begin{subfigure}[b]{0.24\textwidth}
        \includegraphics[width=\linewidth]{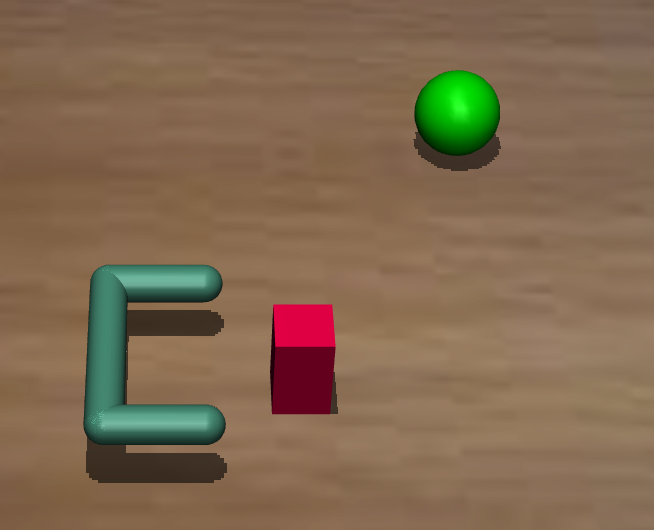}
        \caption{Block push}\label{fig:block_push}
    \end{subfigure}
    \caption{The agent needs to either navigate in absence of winds, navigate in presence of winds, use its gripper to localize an object at an unobserved target location or push the block to an unobserved target location, indicated by green sphere (or cube), by exploring its environment and experiencing reward. While tasks vary in reward functions for \texttt{Ant navigation}, \texttt{Object localization} and \texttt{Block push}, they vary in dynamics function for \texttt{Wind navigation}.}
    \label{fig:my_label}
    \vspace{-0.3cm}
\end{figure}
% \end{wrapfigure}
% \vspace{-0.4cm}
We aim to comprehensively evaluate \rml~and answer the following questions: \textbf{(1)} Do meta-policies learned via \rml~allow for quick adaptation under different distribution shifts in the test-time task distribution? \textbf{(2)} Does learning for multiple levels of robustness actually help the algorithm adapt more effectively than a particular chosen uncertainty level? \textbf{(3)} Does proposing uncertainty sets via generative modeling provide useful distributions of tasks for robustness?
 
\textbf{Setup.} We use \rml~on four continuous control environments: \texttt{Ant navigation} (controlling a four-legged robotic quadruped), \texttt{Wind navigation}~\cite{dorfman2020offline} (controlling a linear system robot in presence of wind), \texttt{Object localization} (controlling a Fetch robot to localize an object through its gripper) and \texttt{Block push} (controlling a robot arm to push an object)~\cite{gupta2018meta} (\Cref{fig:ant_env,fig:wind_env,fig:fetch_reach,fig:block_push}) (see Appendix~\ref{sec:env_desc} for details about reward function and dynamics). We design various meta RL tasks from these environments. Each meta RL task has a train task distribution $\mathcal{T}_i \sim p_{\text{train}}(\mathcal{T})$ such that each task $\mathcal{T}_i$ either parameterizes a reward function $r_i(s,a) \coloneqq r(s,a,\mathcal{T}_i)$ or a dynamics function $p_i(s,a) \coloneqq p(s,a,\mathcal{T}_i)$. $\gT_i$ itself remains unobserved, the agent simply has access to reward values and executing actions in the environment. The learned meta-policies are evaluated on different distributionally shifted test task distributions $\{p^i_{\text{test}}(\mathcal{T})\}_{i=1}^K$ which are either in-support or out-of-support of training task distribution. In all meta RL tasks, the train and test task distribution is determined by the distribution of an underlying task parameter (i.e. wind velocity $w_\gT$ for \texttt{Wind navigation} and target location $s_\gT$ for other environments), which either determines the reward function or the dynamics function. While tasks vary in reward functions for \texttt{Ant navigation}, \texttt{Object localization} and \texttt{Block push}, they vary in dynamics function for \texttt{Wind navigation} (exact task distributions in Table~\ref{tbl:detail_task_dist}). We use $4$ random seeds for all our experiments and include the standard error bars in our plots.

\subsection{Adaptation to Varying Levels of Distribution Shift}
\begin{figure}[!t]
    \centering
    \includegraphics[width=0.24\linewidth]{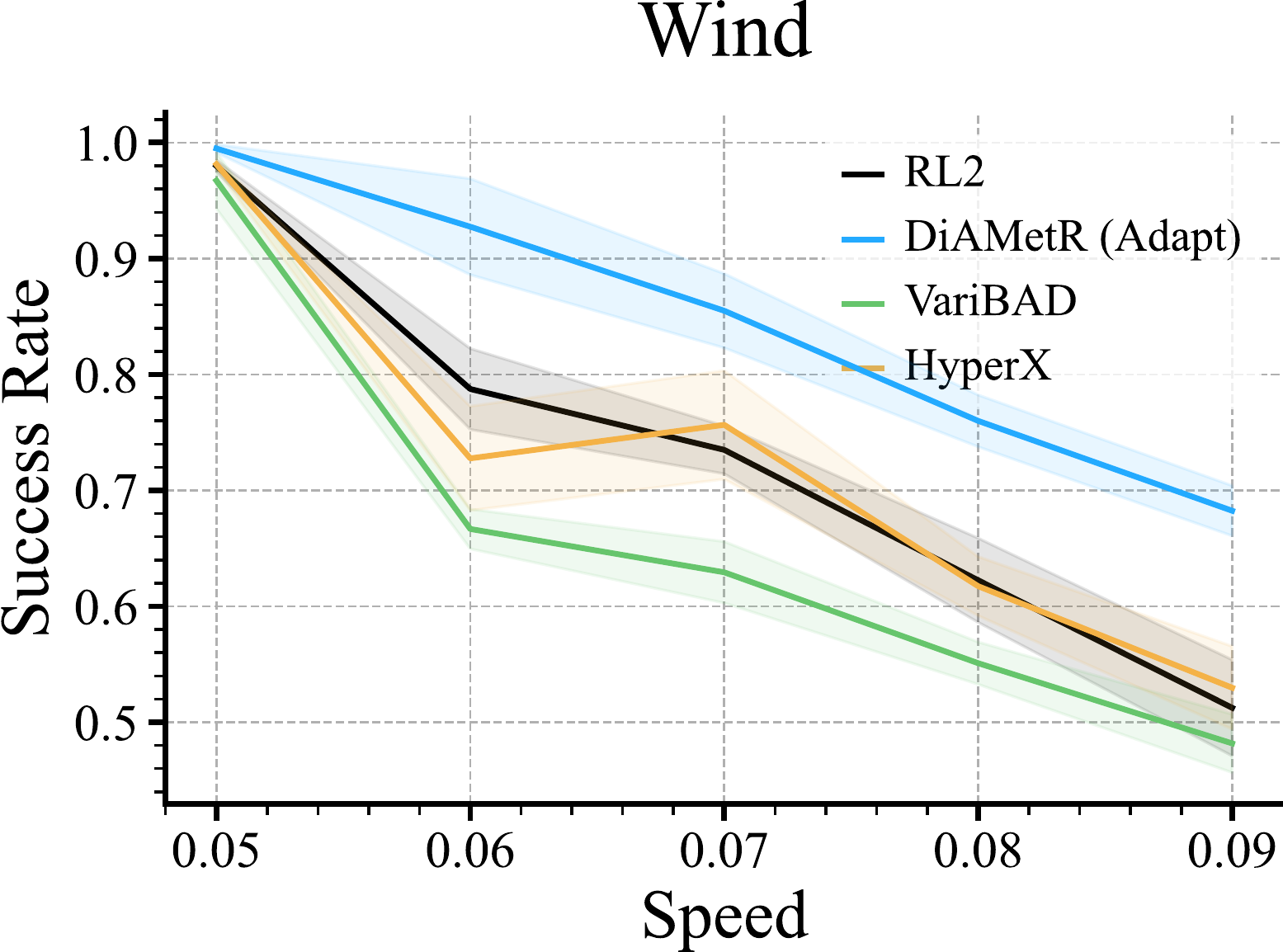}
    \includegraphics[width=0.24\linewidth]{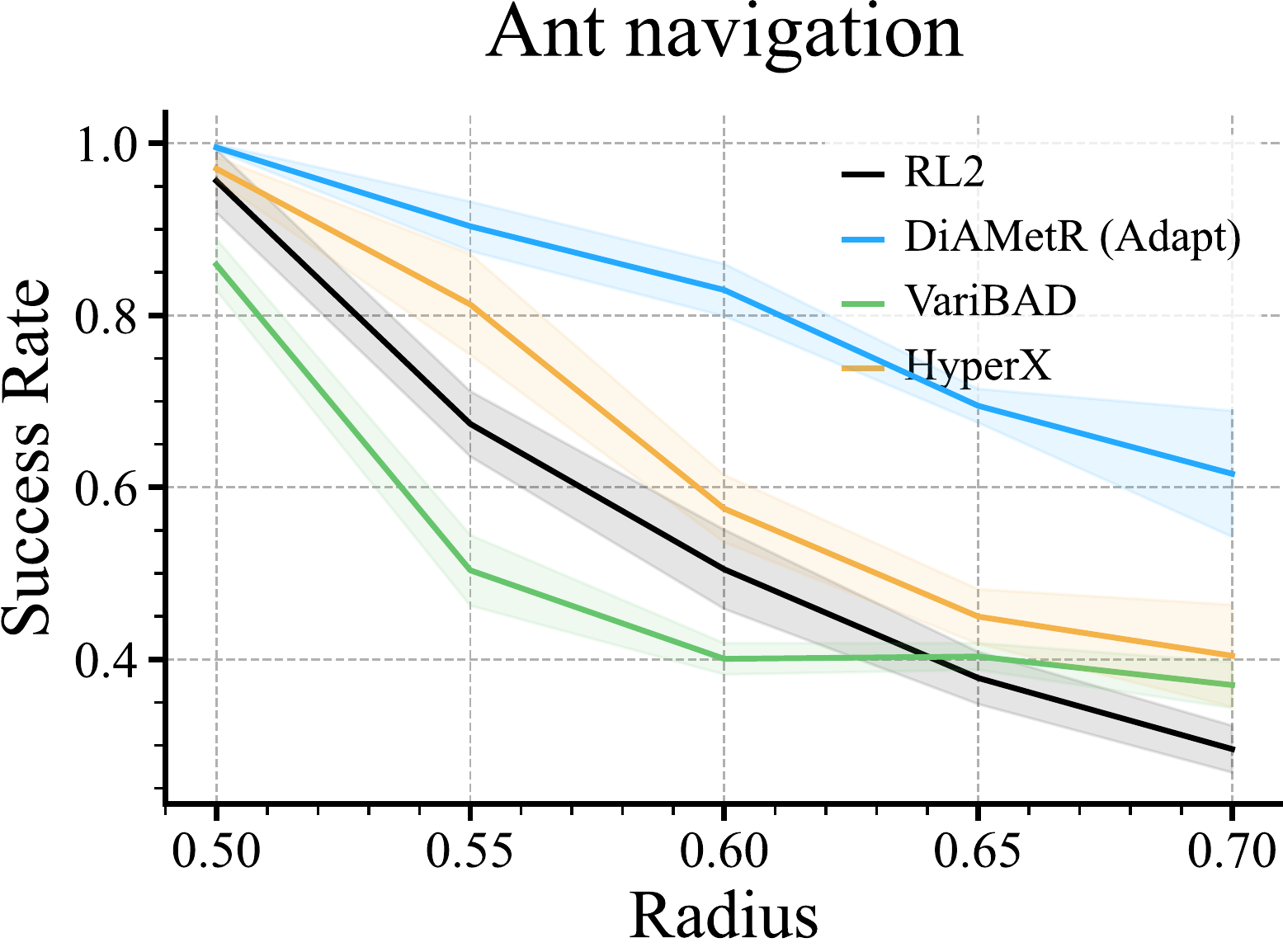}
    \includegraphics[width=0.24\linewidth]{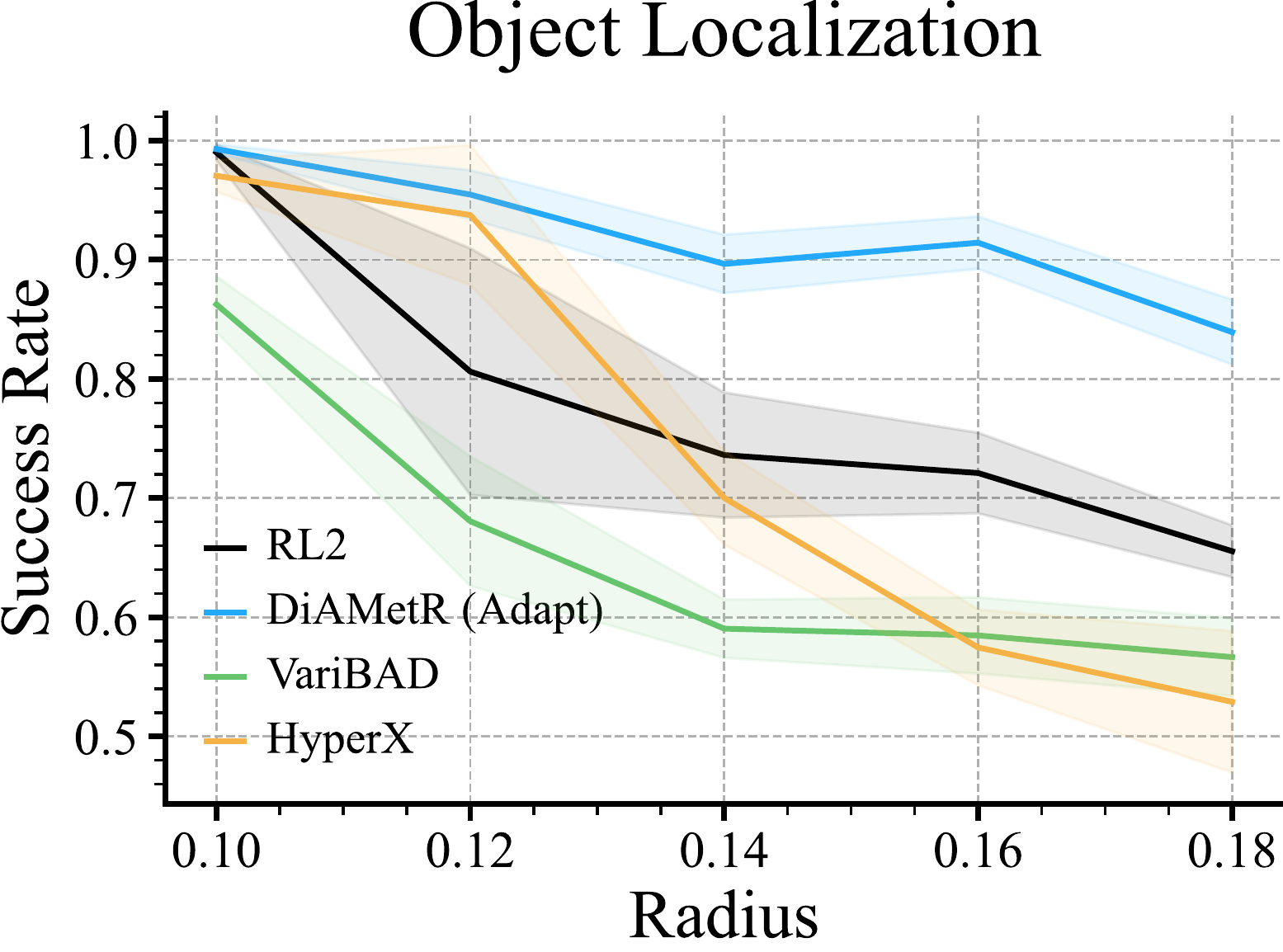}
    \includegraphics[width=0.24\linewidth]{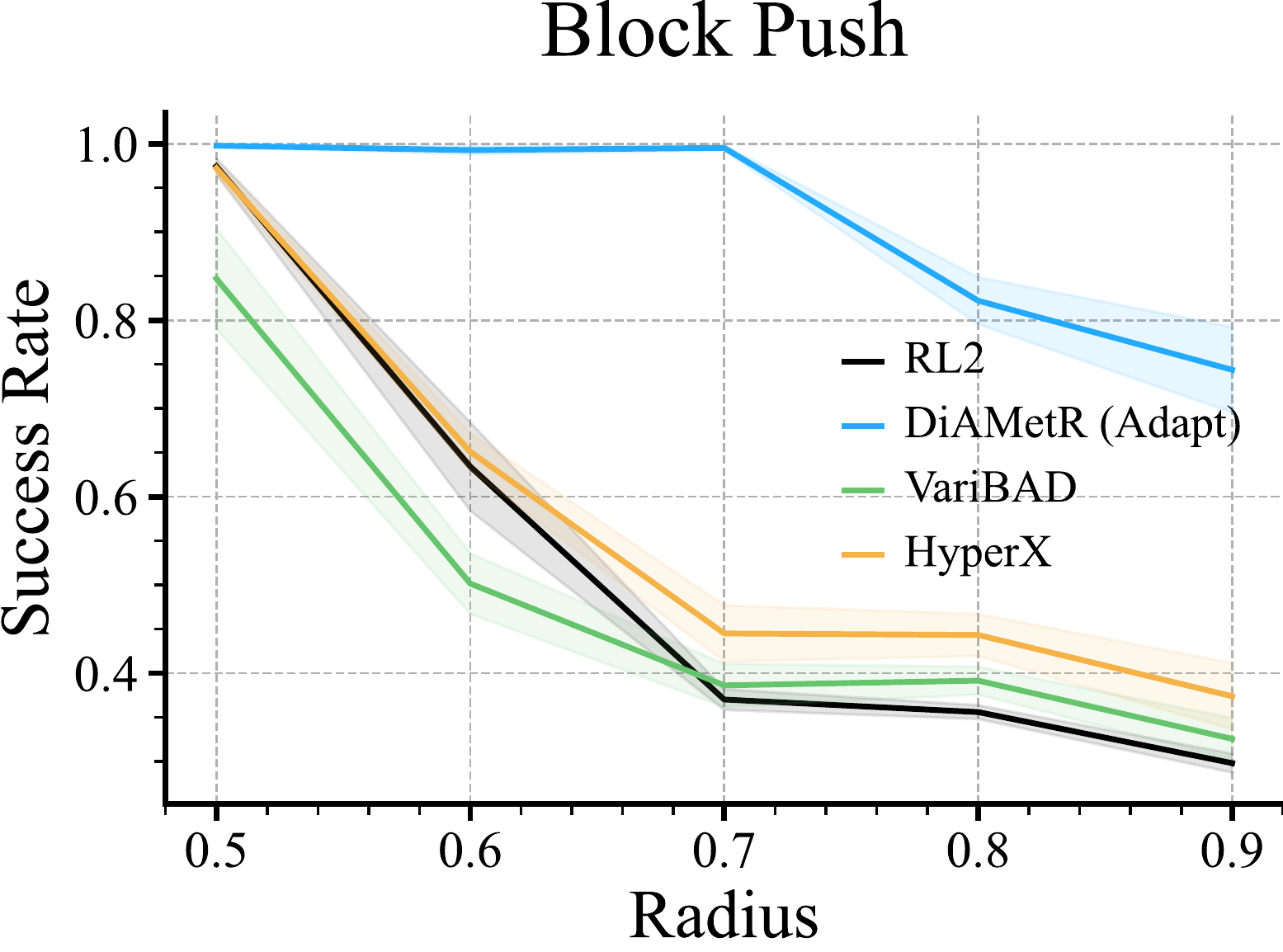}\\
    \includegraphics[width=0.24\linewidth]{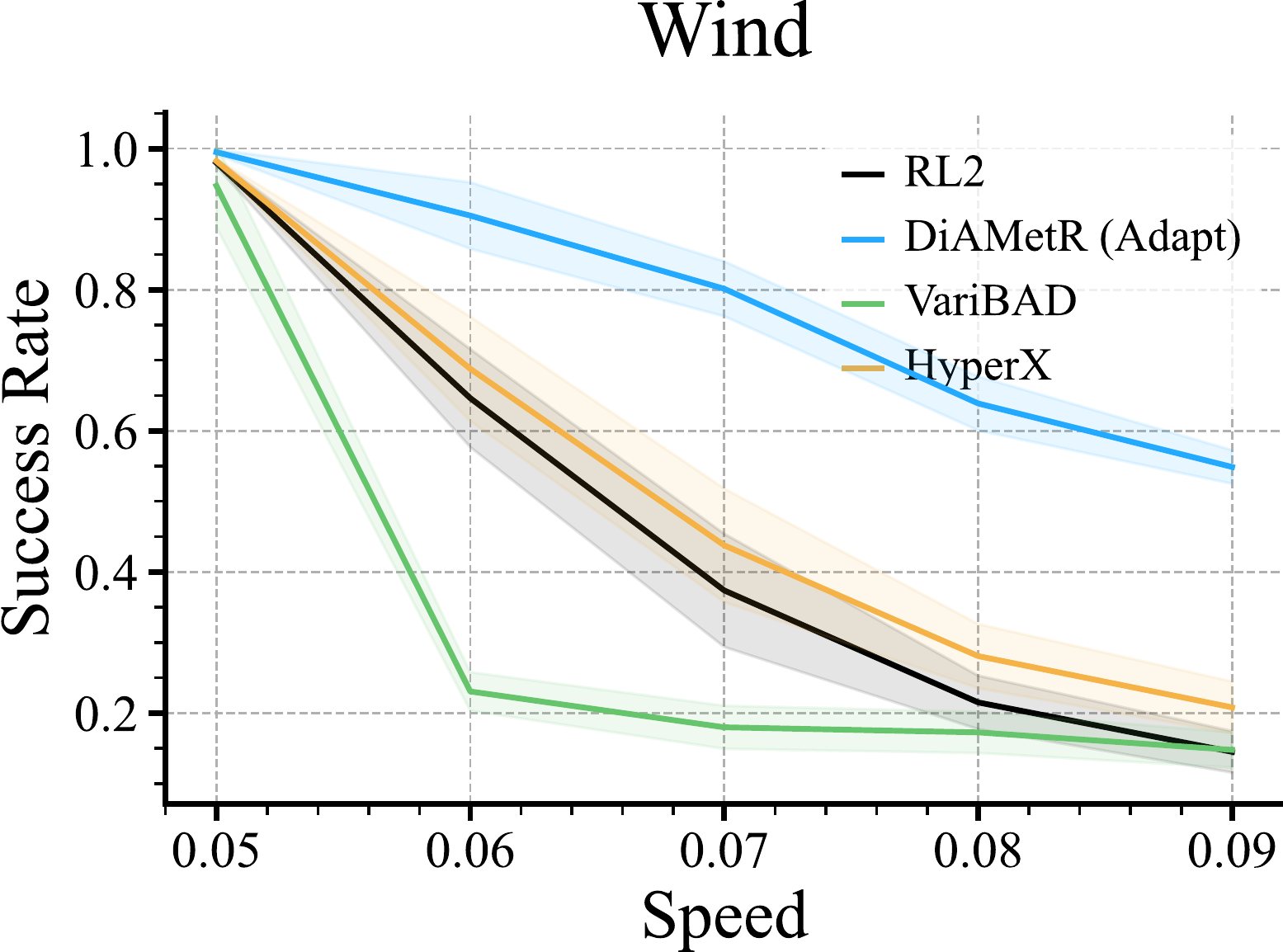}
    \includegraphics[width=0.24\linewidth]{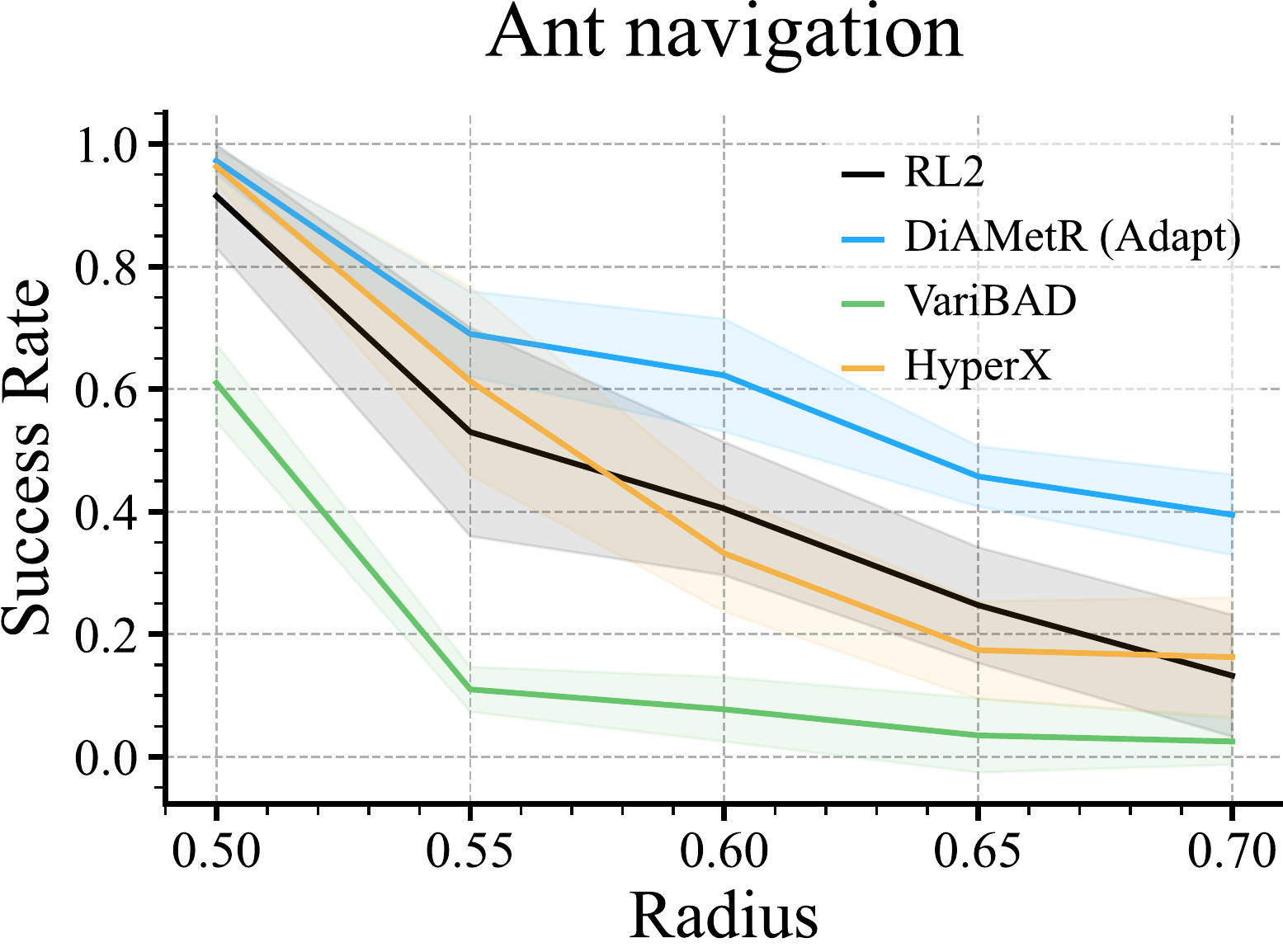}
    \includegraphics[width=0.24\linewidth]{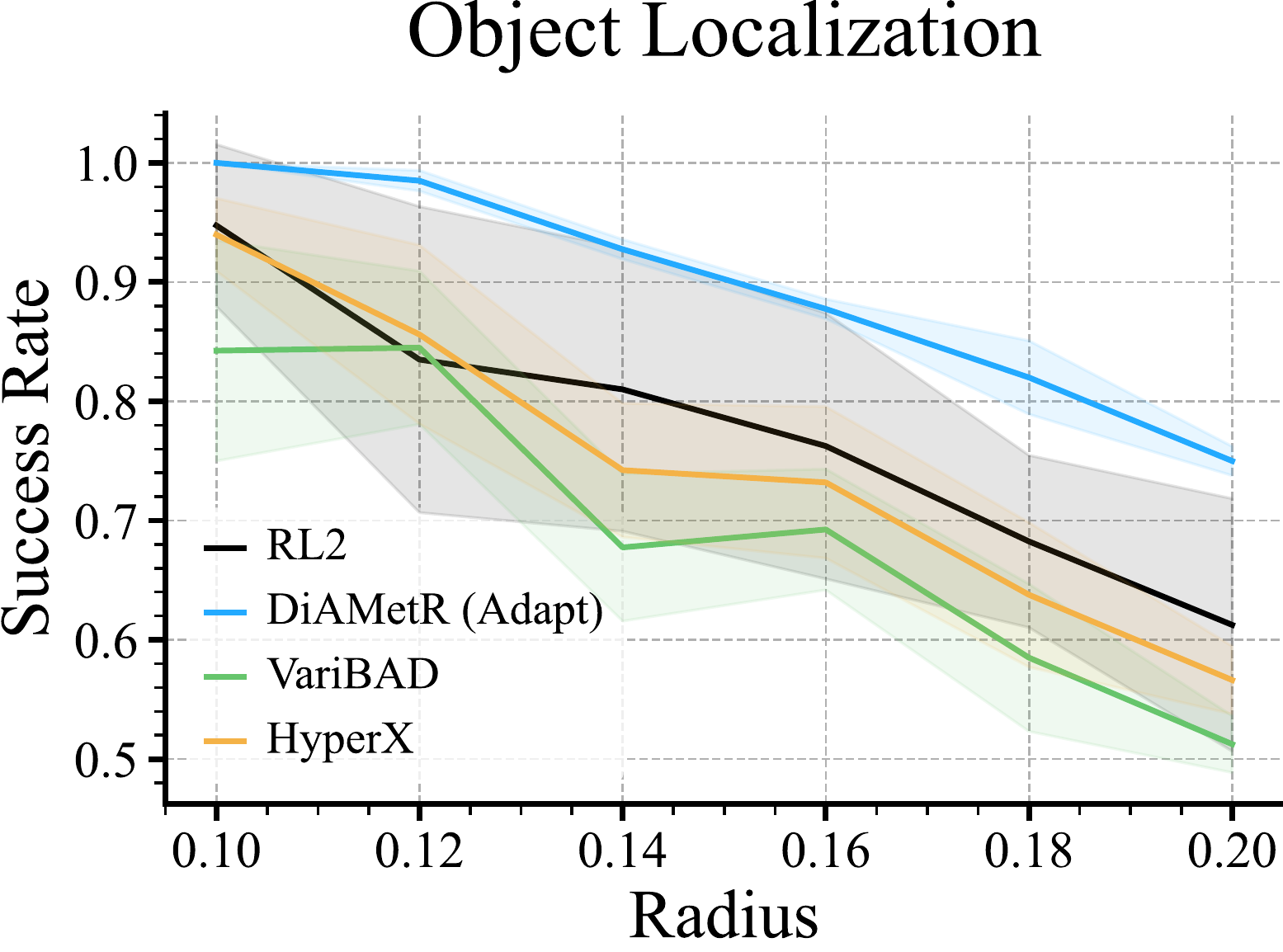}
    \includegraphics[width=0.24\linewidth]{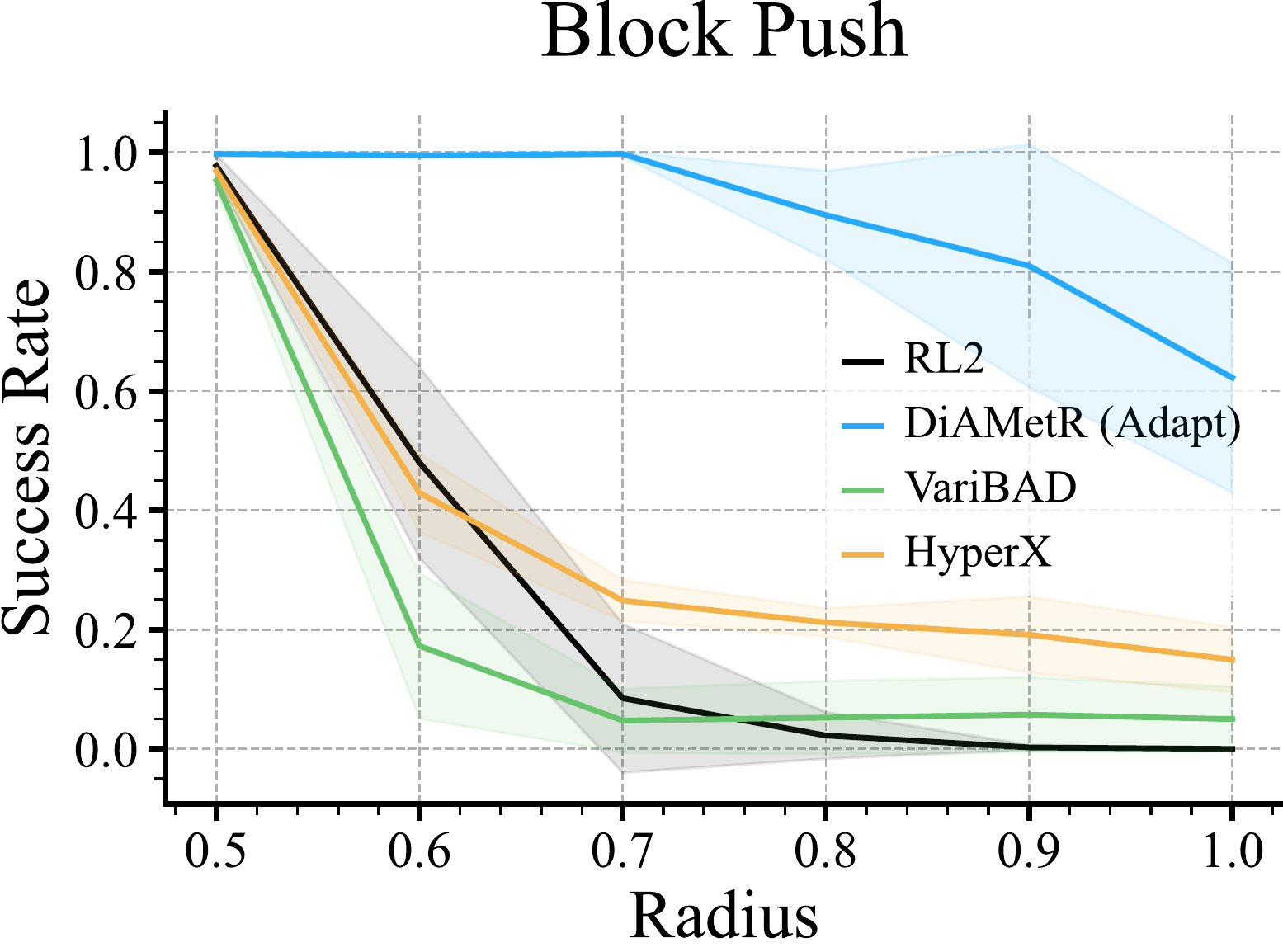}
    \caption{We evaluate \rml~and meta RL algorithms ($\text{RL}^2$, VariBAD and HyperX) on different \textit{in-support} (top row) and \textit{out-of-support} (bottom row) shifted test task distributions. \rml~either matches or outperforms $\text{RL}^2$, VariBAD and HyperX on these task distributions. The first point $p_1$ on the horizontal axis indicates the task parameter ($\Delta$) distribution $\mathcal{U}(0, p_1)$ and the subsequent points $p_i$ indicate task parameter ($\Delta$) distribution $\mathcal{U}(p_{i-1}, p_i)$. While task parameter for \texttt{wind navigation} is wind velocity $w_\gT$, it is target location $s_\gT$ for other environments. Table~\ref{tbl:detail_task_dist} details the task distributions used in this evaluation.} 
    \label{fig:adapt_diff_shifts}
    % \vspace{-0.2cm}
\end{figure}

During meta test, given a test task distribution $p_\testt(\gT)$, \rml~uses Thompson sampling to select the appropriate meta-policy $\pi_{\text{meta},\theta}^\epsilon$ within $N=250$ meta episodes. $\pi_{\text{meta},\theta}^\epsilon$ can then solve any new task $\gT \sim p_\testt(\gT)$ within $1$ meta episode ($k$ environment episode). To test \rml's ability to adapt to varying levels of distribution shift, we evaluate it on different test task distributions, as detailed in Table~\ref{tbl:detail_task_dist}. We compare \rml~with meta RL algorithms such as (off-policy) $\text{RL}^2$~\citep{ni2022recurrent}, VariBAD~\citep{zintgraf2019varibad} and HyperX~\citep{zintgraf2021exploration}. Since \rml~uses $250$ meta-episodes to adaptively choose a meta-policy during test time, we finetune $\text{RL}^2$, VariBAD and HyperX with $250$ meta-episodes of test task distribution to make the comparisons fair (see Appendix~\ref{sec:meta_test_adapt} for the finetuning curves). Figure~\ref{fig:adapt_diff_shifts} show that \rml~outperforms $\text{RL}^2$, VariBAD and HyperX on \textit{out-of-support} and \textit{in-support} shifted test task distributions. Furthermore, the performance gap between \rml~and other baselines increase as distribution shift between test task distribution and train task distribution increases. Naturally, the performance of \rml~also deteriorates as the distribution shift is increased, but as shown in Fig~\ref{fig:adapt_diff_shifts}, it does so much more slowly than other algorithms. We also evaluate \rml~on train task distribution to see if it incurs any performance loss. Figure~\ref{fig:adapt_diff_shifts} shows that \rml~either matches or outperforms $\text{RL}^2$, VariBAD, and HyperX on the train task distribution. We refer readers to Appendix~\ref{sec:ablate} for ablation studies and further experimental evaluations. 

\subsection{Analysis of Tasks Proposed by Latent Conditional Uncertainty Sets}
\begin{figure}
    \centering
    \begin{subfigure}[b]{0.12\textwidth}
        \includegraphics[width=\linewidth]{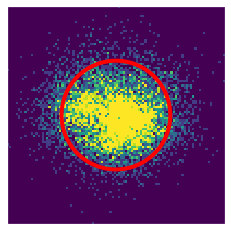}
        \caption{$\epsilon = 0.1$}
    \end{subfigure}
    \begin{subfigure}[b]{0.12\textwidth}
        \includegraphics[width=\linewidth]{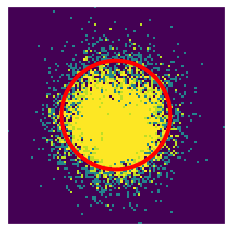}
        \caption{$\epsilon = 0.2$}
    \end{subfigure}
    \begin{subfigure}[b]{0.12\textwidth}
        \includegraphics[width=\linewidth]{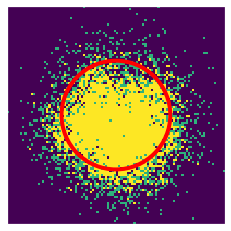}
        \caption{$\epsilon = 0.4$}
    \end{subfigure}
    \begin{subfigure}[b]{0.12\textwidth}
        \includegraphics[width=\linewidth]{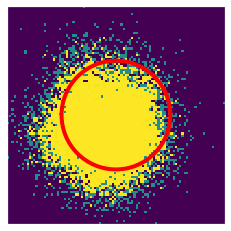}
        \caption{$\epsilon = 0.8$}
    \end{subfigure}
    \caption{Imagined test reward distributions in \texttt{Ant-navigation} environment in increasing order of distribution shifts. Train reward distribution is uniform within the red circle.}
    \label{fig:test_task_proposal}
\end{figure}

We visualize the imagined test reward distribution through their heatmaps for various distribution shifts. To generate an imagined reward function, we sample $z \sim q_\phi(z)$ and pass the $z$ into $r_\omega(s,a,z) = r_\omega(s,z)$ (given the learned reward is only dependent on state as mentioned in Appendix~\ref{sec:structure_vae}). We then take the agent (for instance the ant) and reset its $(x,y)$ location to different points in the (discretized) grid $[-1,1]^2$ and calculate $r_\omega(s,z)$ at all those points. This gives us a reward map for a single imagined reward function. We sample $10000$ of these reward functions and plot them together. Figure~\ref{fig:test_task_proposal} visualizes the imagined test reward distribution in \texttt{Ant-navigation} environment in increasing order of distribution shifts with respect to train reward distribution (with distribution shift parameter $\epsilon$ increasing from left to right). The train distribution of rewards has uniformly distributed target locations within the red circle. As clearly seen in Figure~\ref{fig:test_task_proposal}, the learned reward distribution model imagines more target locations outside the red circle as we increase the distribution shifts.

% \begin{wrapfigure}{r}{0.5\linewidth}

\subsection{Analysis of Importance of Multiple Uncertainty Sets}
\begin{figure}[!t]
    \centering
    \includegraphics[width=0.24\linewidth]{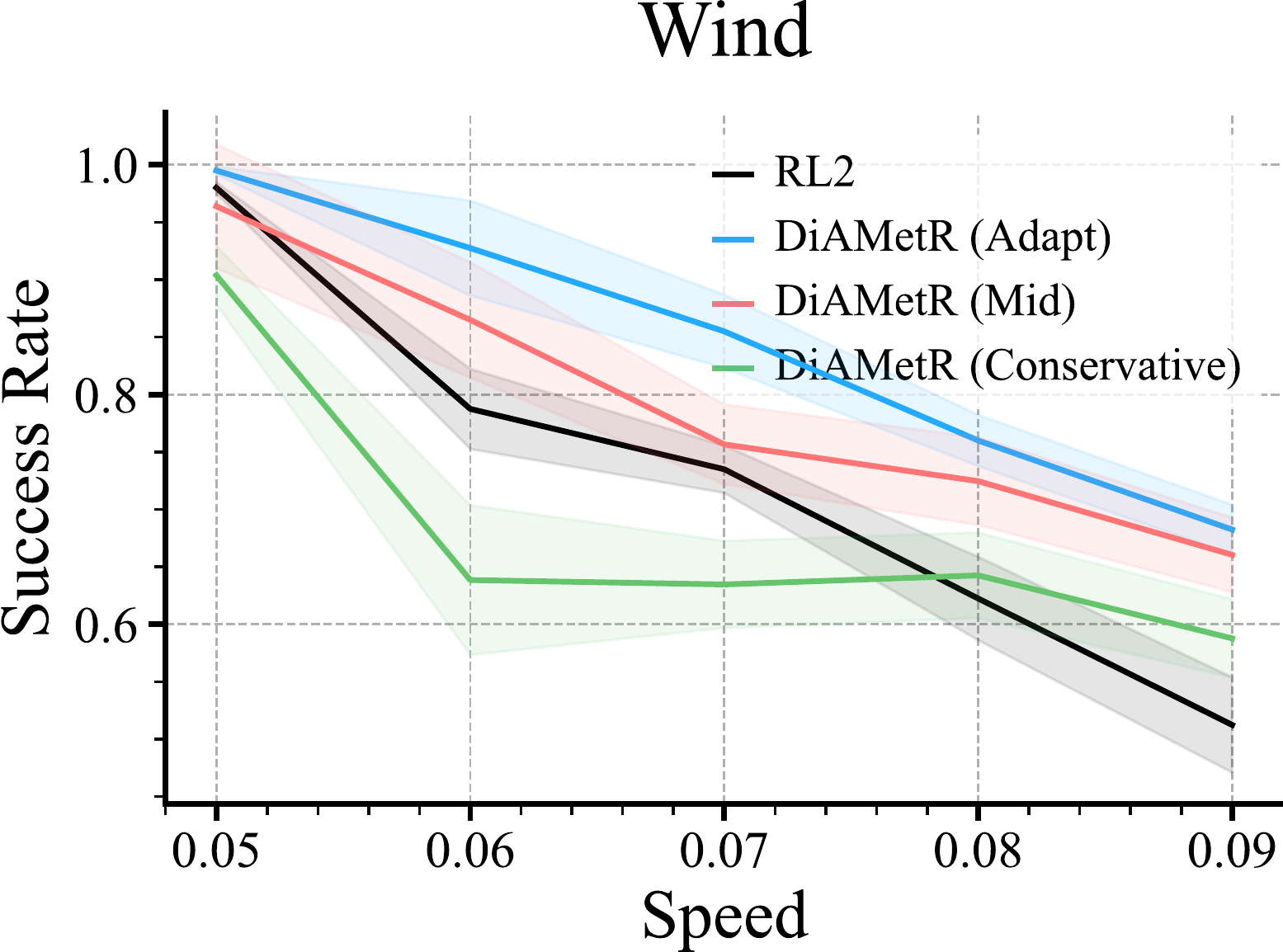}
    \includegraphics[width=0.24\linewidth]{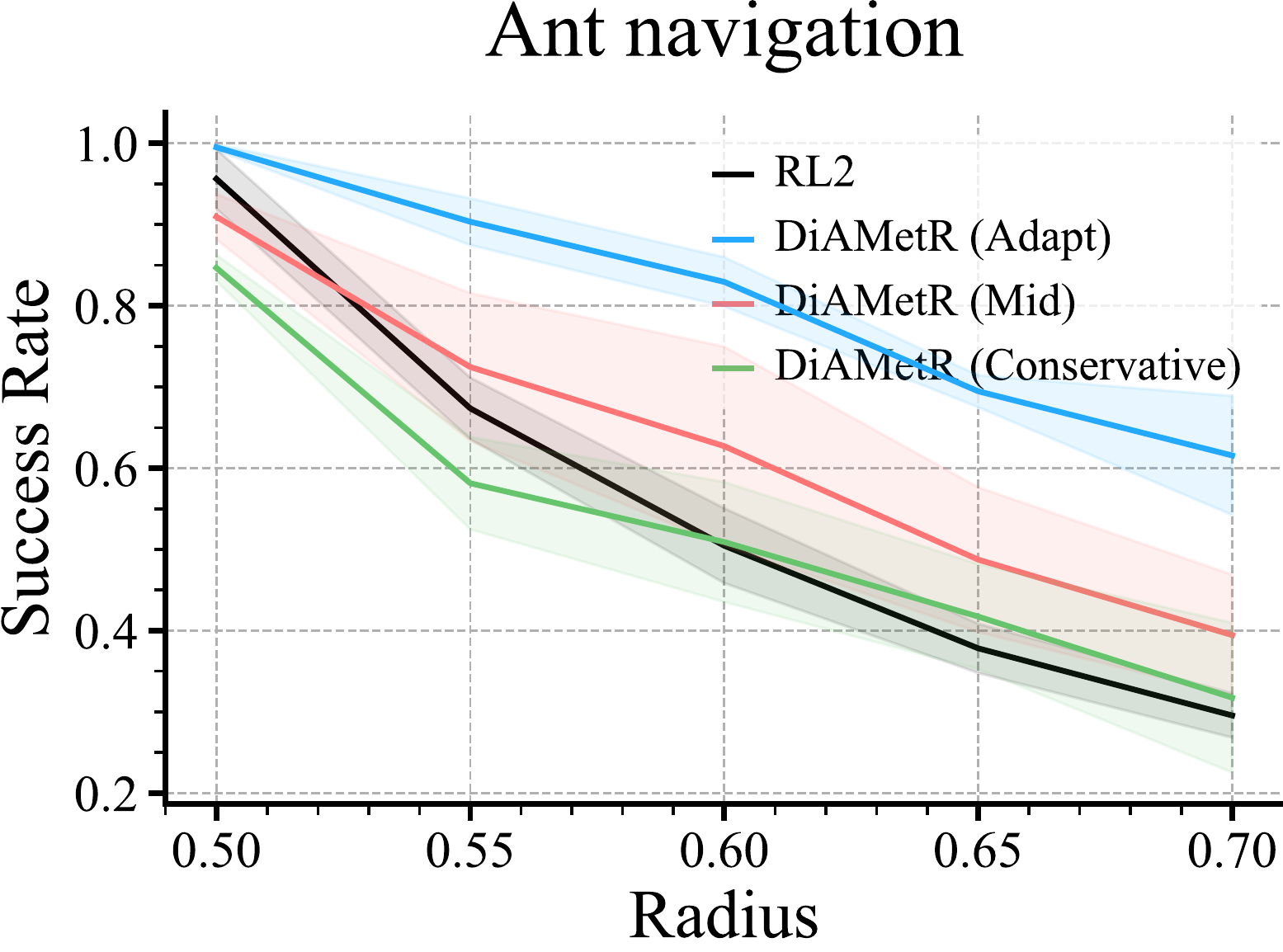}
    \includegraphics[width=0.24\linewidth]{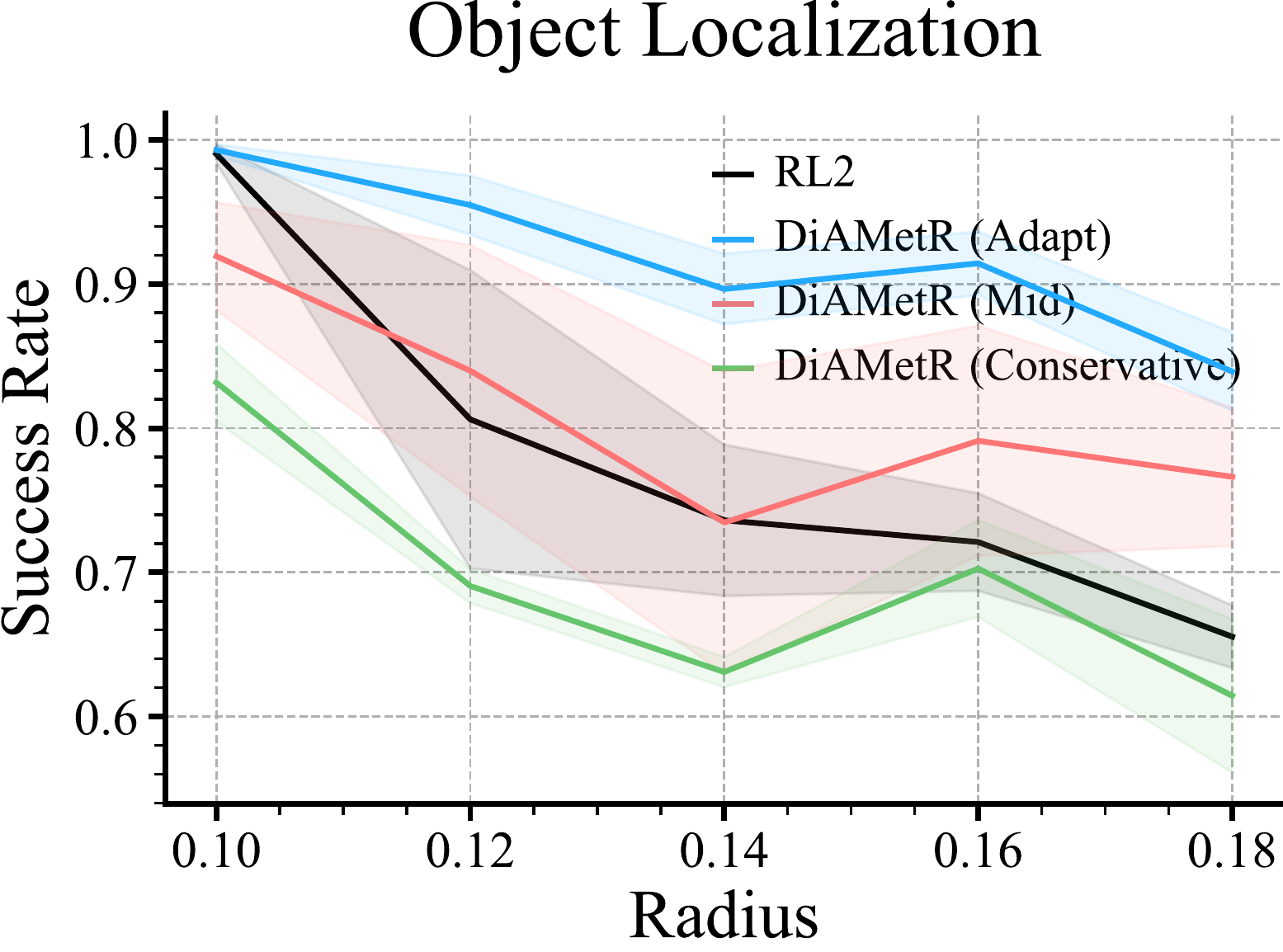}
    \includegraphics[width=0.24\linewidth]{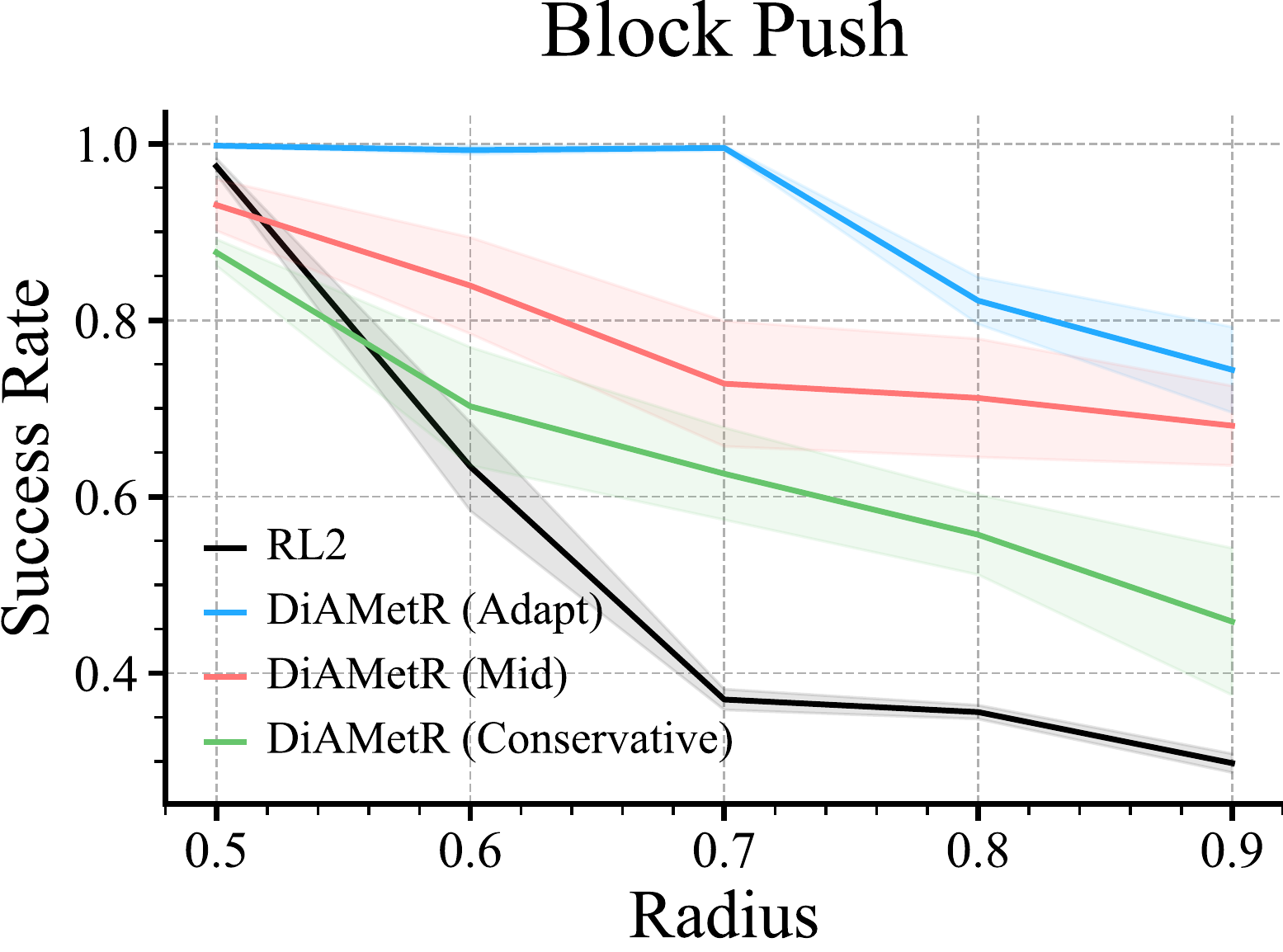}\\
    \includegraphics[width=0.24\linewidth]{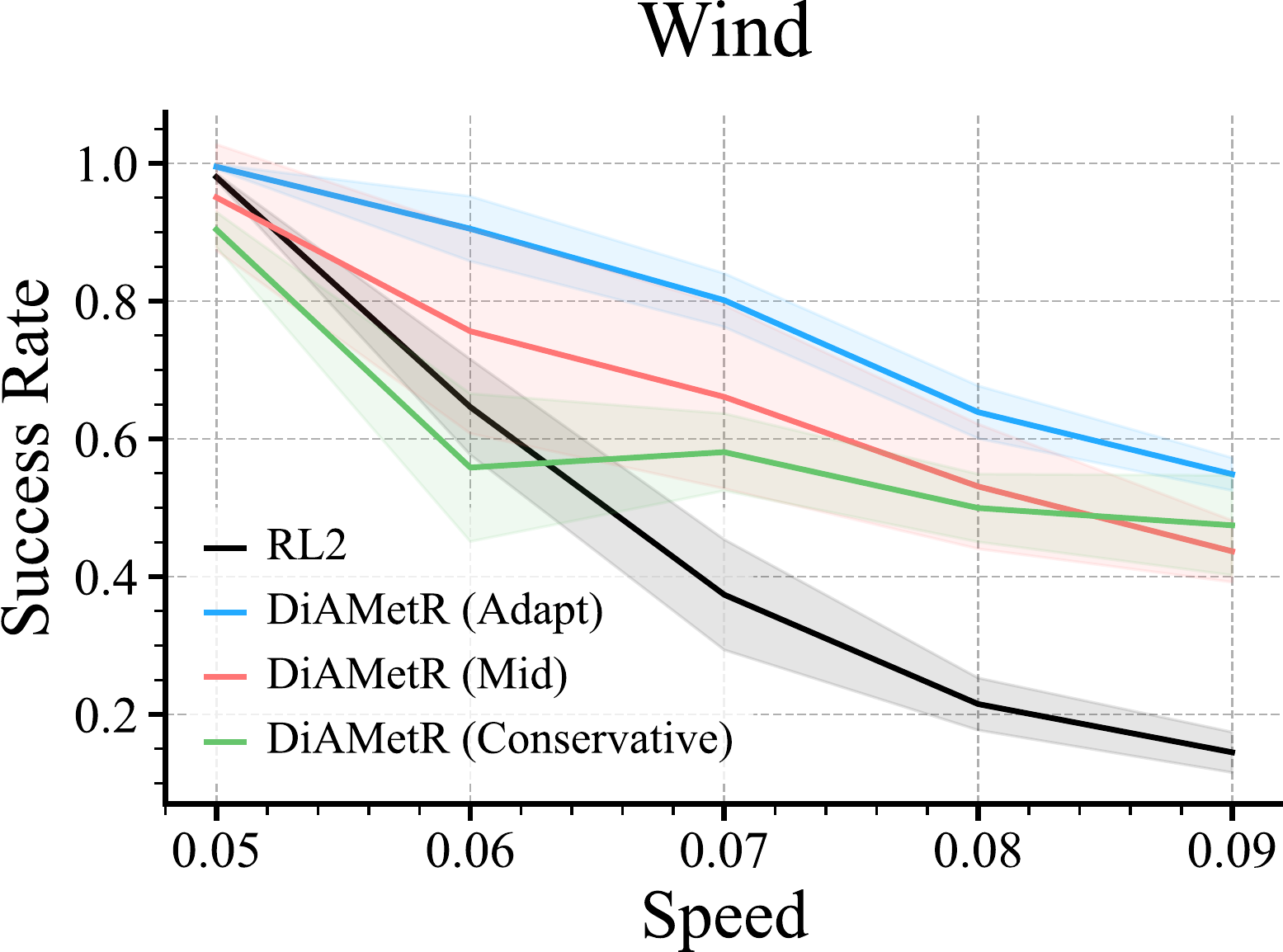}
    \includegraphics[width=0.24\linewidth]{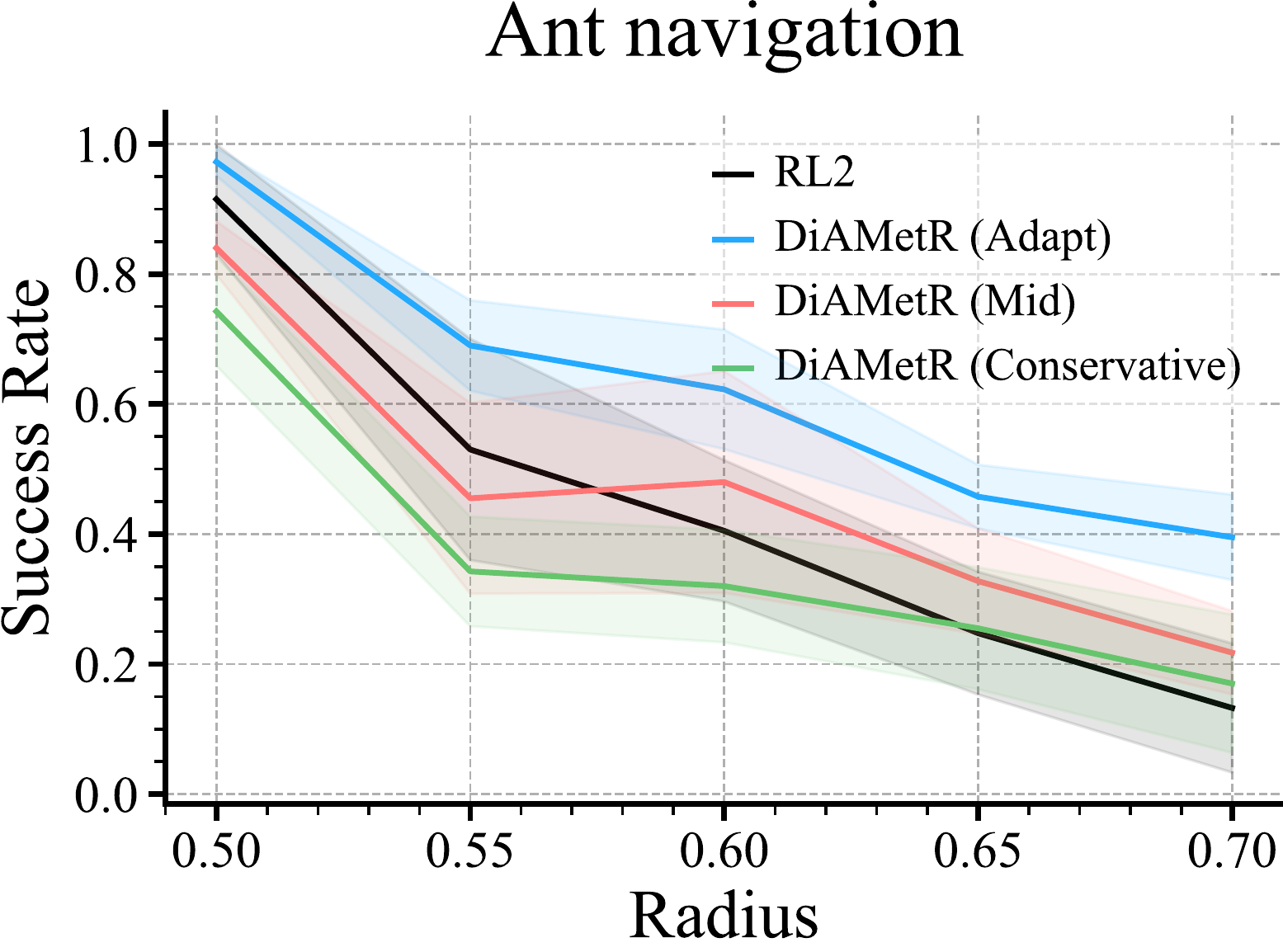}
    \includegraphics[width=0.24\linewidth]{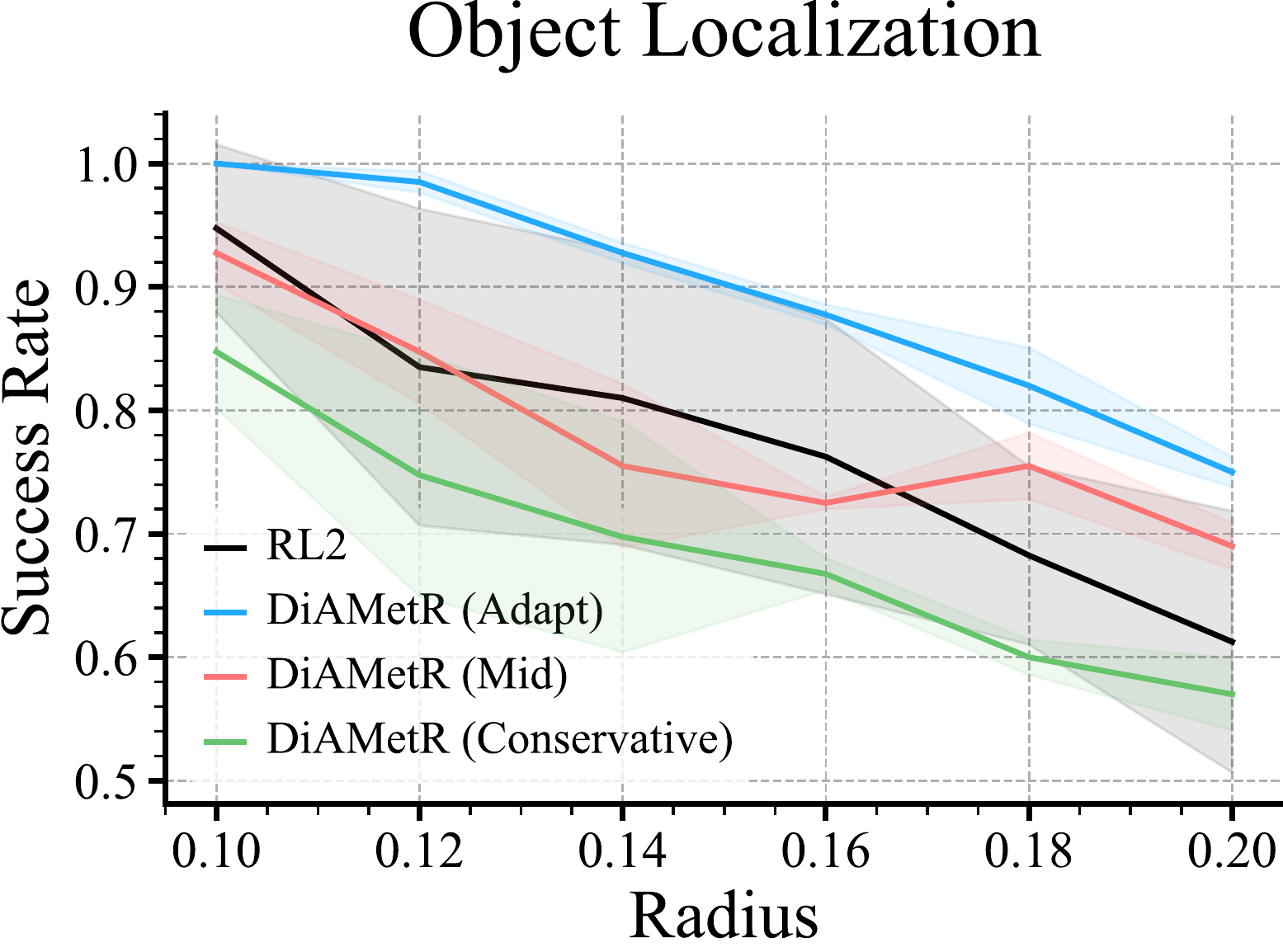}
    \includegraphics[width=0.24\linewidth]{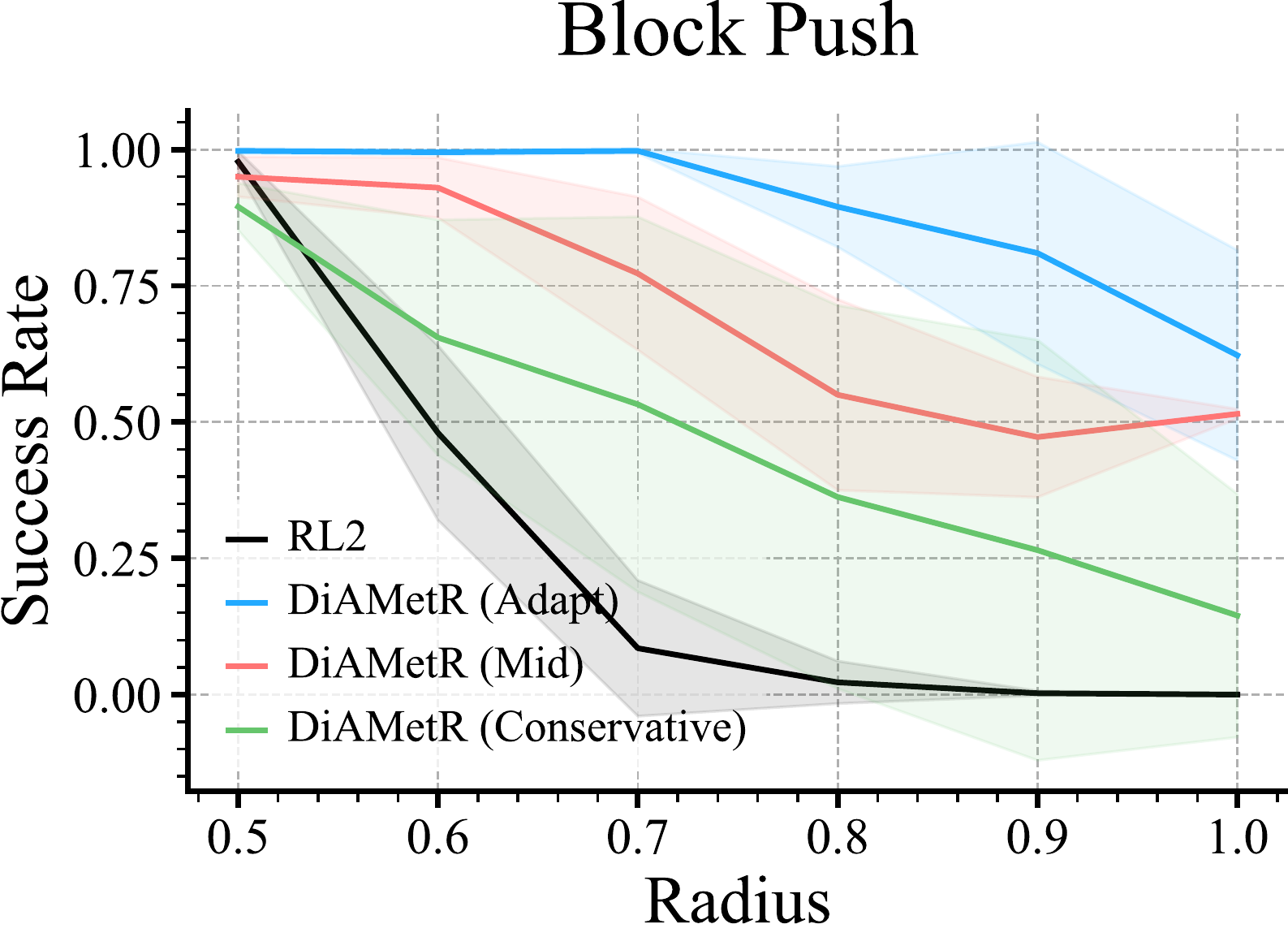}
    \caption{Adaptively choosing an uncertainty set for \rml~policy (Adapt) during test time allows it to better adapt to test time distribution shift than choosing an uncertainty set beforehand (Mid). Choosing a large uncertainty set for \rml~policy (Conservative) leads to a conservative behavior and hurts its performance when test time distribution shift is low. While top row contains \textit{in-support} task distribution shifts, the bottom row contains \textit{out-of-support} task distribution shifts. 
    % The first point $p_1$ on the horizontal axis indicates the task parameter ($\Delta$) distribution $\mathcal{U}(0, p_1)$ and the subsequent points $p_i$ indicate task parameter ($\Delta$) distribution $\mathcal{U}(p_{i-1}, p_i)$. While task parameter for \texttt{wind navigation} is wind velocity $w_\gT$, it is target location $s_\gT$ for other environments. Table~\ref{tbl:detail_task_dist} details the task distributions used in this evaluation.
    } 
    \label{fig:adapt_multi_set}
\end{figure}

\rml~meta-learns a family of adaptation policies, each conditioned on different uncertainty set. As discussed in section~\ref{sec:approach}, selecting a policy conditioned on a large uncertainty set would lead to overly conservative behavior. Furthermore, selecting a policy conditioned on a small uncertainty set would result in failure if the test time distribution shift is high. Therefore, we need to adaptively select an uncertainty set during test time. To validate this phenomenon empirically, we performed an ablation study in Figure~\ref{fig:adapt_multi_set}. As clearly visible, adaptively choosing an uncertainty set during test time allows for better test time distribution adaptation when compared to selecting an uncertainty set beforehand or selecting a large uncertainty set. These results suggest that a combination of training robust meta-learners and constructing various uncertainty sets allows for effective test-time adaptation under distribution shift. \rml~ is able to avoid both overly conservative behavior and under-exploration at test-time. 

% \subsection{[TODO method name] on zero-shot contextual generalization problems}

% Show results on just zero-shot performance with multiple uncertainty sets and test-time adaptation, no exploration. 

% \begin{figure}[!h]
%     \centering
%     \includegraphics[width=0.45\linewidth]{figs/placeholder.jpg}
%     \includegraphics[width=0.45\linewidth]{figs/placeholder.jpg}
%     \caption{Zero-shot and Mutliple Level uncertainty}
%     \label{fig:my_label}
% \end{figure}

\section{Discussion} 
\label{sec:discussion}
\vspace{-0.3cm}
In this work, we discussed the challenge of distribution shift in meta-reinforcement learning and showed how we can build meta-reinforcement learning algorithms that are robust to varying levels of distribution shift. We show how we can build distributionally ``adaptive" reinforcement learning algorithms that can adapt to varying levels of distribution shift, retaining a tradeoff between fast learning and maintaining asymptotic performance. We then show we can instantiate this algorithm practically by parameterizing uncertainty sets with a learned generative model. We empirically showed that this allows for learning meta-learners robust to changes in task distribution.

There are several avenues for future work we are keen on exploring, for instance extending adaptive distributional robustness to more complex meta RL tasks, including those with differing transition dynamics. Another interesting direction would be to develop a more formal theory providing adaptive robustness guarantees in meta-RL problems under these inherent distribution shifts. 

\section*{Acknowledgements}
The authors thank the members of Improbable AI Lab \& RAIL for discussions and helpful feedback. We thank MIT Supercloud and the Lincoln Laboratory Supercomputing Center for providing compute resources. This research was supported by an NSF graduate fellowship, a DARPA Machine Common Sense grant, ARO MURI Grant Number W911NF-21-1-0328, an MIT-IBM grant, and ARO W911NF-21-1-0097.

This research was also partly sponsored by the United States Air Force Research Laboratory and the United States Air Force Artificial Intelligence Accelerator and was accomplished under Cooperative Agreement Number FA8750-19- 2-1000. The views and conclusions contained in this document are those of the authors and should not be interpreted as representing the official policies, either expressed or implied, of the United States Air Force or the U.S. Government. The U.S. Government is authorized to reproduce and distribute reprints for Government purposes, notwithstanding any copyright notation herein.

\section*{Author Contributions}
\textbf{Anurag Ajay} helped in coming up with the framework of distributionally adaptive meta RL, implemented the DiaMetR algorithm, ran the experiments, helped with writing the paper and led the project.

\textbf{Abhishek Gupta} came up with the framework of distributionally adaptive meta RL, ran initial proof-of-concept experiments, helped with writing the paper and co-led the project.

\textbf{Dibya Ghosh} analyzed the framework of distributionally adaptive meta RL (Section 4.3), wrote all the related proofs, participated in research discussions and helped with writing the paper.

\textbf{Sergey Levine} provided feedback on the work, and participated in research discussions.

\textbf{Pulkit Agrawal} participated in research discussions and advised on the project.

\normalsize
%\newpage
\bibliographystyle{abbrvnat}
\bibliography{references}

\begin{thebibliography}{49}
\providecommand{\natexlab}[1]{#1}
\providecommand{\url}[1]{\texttt{#1}}
\expandafter\ifx\csname urlstyle\endcsname\relax
  \providecommand{\doi}[1]{doi: #1}\else
  \providecommand{\doi}{doi: \begingroup \urlstyle{rm}\Url}\fi

\bibitem[Chen et~al.(2022)Chen, Xu, and Agrawal]{pmlr-v164-chen22a}
T.~Chen, J.~Xu, and P.~Agrawal.
\newblock A system for general in-hand object re-orientation.
\newblock In A.~Faust, D.~Hsu, and G.~Neumann, editors, \emph{Proceedings of
  the 5th Conference on Robot Learning}, volume 164 of \emph{Proceedings of
  Machine Learning Research}, pages 297--307. PMLR, 08--11 Nov 2022.
\newblock URL \url{https://proceedings.mlr.press/v164/chen22a.html}.

\bibitem[Cohen et~al.(2019)Cohen, Rosenfeld, and Kolter]{cohen2019certified}
J.~Cohen, E.~Rosenfeld, and Z.~Kolter.
\newblock Certified adversarial robustness via randomized smoothing.
\newblock In \emph{International Conference on Machine Learning}, pages
  1310--1320. PMLR, 2019.

\bibitem[Collins et~al.(2020)Collins, Mokhtari, and
  Shakkottai]{collins2020trmaml}
L.~Collins, A.~Mokhtari, and S.~Shakkottai.
\newblock Distribution-agnostic model-agnostic meta-learning.
\newblock \emph{CoRR}, abs/2002.04766, 2020.
\newblock URL \url{https://arxiv.org/abs/2002.04766}.

\bibitem[De~Boer et~al.(2005)De~Boer, Kroese, Mannor, and
  Rubinstein]{de2005tutorial}
P.-T. De~Boer, D.~P. Kroese, S.~Mannor, and R.~Y. Rubinstein.
\newblock A tutorial on the cross-entropy method.
\newblock \emph{Annals of operations research}, 134\penalty0 (1):\penalty0
  19--67, 2005.

\bibitem[Deleu and Bengio(2018)]{deleu2018effects}
T.~Deleu and Y.~Bengio.
\newblock The effects of negative adaptation in model-agnostic meta-learning.
\newblock \emph{arXiv preprint arXiv:1812.02159}, 2018.

\bibitem[Dorfman et~al.(2020)Dorfman, Shenfeld, and Tamar]{dorfman2020offline}
R.~Dorfman, I.~Shenfeld, and A.~Tamar.
\newblock Offline meta learning of exploration.
\newblock \emph{arXiv preprint arXiv:2008.02598}, 2020.

\bibitem[Duan et~al.(2016)Duan, Schulman, Chen, Bartlett, Sutskever, and
  Abbeel]{duan2016rl}
Y.~Duan, J.~Schulman, X.~Chen, P.~L. Bartlett, I.~Sutskever, and P.~Abbeel.
\newblock Rl2: Fast reinforcement learning via slow reinforcement learning.
\newblock \emph{arXiv preprint arXiv:1611.02779}, 2016.

\bibitem[Fallah et~al.(2021)Fallah, Mokhtari, and
  Ozdaglar]{fallah2021generalization}
A.~Fallah, A.~Mokhtari, and A.~Ozdaglar.
\newblock Generalization of model-agnostic meta-learning algorithms: Recurring
  and unseen tasks.
\newblock \emph{Advances in Neural Information Processing Systems}, 34, 2021.

\bibitem[Filos et~al.(2020)Filos, Tigkas, McAllister, Rhinehart, Levine, and
  Gal]{filos2020can}
A.~Filos, P.~Tigkas, R.~McAllister, N.~Rhinehart, S.~Levine, and Y.~Gal.
\newblock Can autonomous vehicles identify, recover from, and adapt to
  distribution shifts?
\newblock In \emph{International Conference on Machine Learning}, pages
  3145--3153. PMLR, 2020.

\bibitem[Finn et~al.(2017)Finn, Abbeel, and Levine]{finn2017model}
C.~Finn, P.~Abbeel, and S.~Levine.
\newblock Model-agnostic meta-learning for fast adaptation of deep networks.
\newblock In \emph{International conference on machine learning}, pages
  1126--1135. PMLR, 2017.

\bibitem[Fu et~al.(2021)Fu, Tang, Hao, Chen, Feng, Li, and Liu]{fu2021towards}
H.~Fu, H.~Tang, J.~Hao, C.~Chen, X.~Feng, D.~Li, and W.~Liu.
\newblock Towards effective context for meta-reinforcement learning: an
  approach based on contrastive learning.
\newblock In \emph{Proceedings of the AAAI Conference on Artificial
  Intelligence}, volume~35, pages 7457--7465, 2021.

\bibitem[Gupta et~al.(2018{\natexlab{a}})Gupta, Eysenbach, Finn, and
  Levine]{gupta2018unsupervised}
A.~Gupta, B.~Eysenbach, C.~Finn, and S.~Levine.
\newblock Unsupervised meta-learning for reinforcement learning.
\newblock \emph{arXiv preprint arXiv:1806.04640}, 2018{\natexlab{a}}.

\bibitem[Gupta et~al.(2018{\natexlab{b}})Gupta, Mendonca, Liu, Abbeel, and
  Levine]{gupta2018meta}
A.~Gupta, R.~Mendonca, Y.~Liu, P.~Abbeel, and S.~Levine.
\newblock Meta-reinforcement learning of structured exploration strategies.
\newblock \emph{Advances in neural information processing systems}, 31,
  2018{\natexlab{b}}.

\bibitem[Haarnoja et~al.(2018)Haarnoja, Zhou, Abbeel, and
  Levine]{haarnoja2018soft}
T.~Haarnoja, A.~Zhou, P.~Abbeel, and S.~Levine.
\newblock Soft actor-critic: Off-policy maximum entropy deep reinforcement
  learning with a stochastic actor.
\newblock \emph{arXiv preprint arXiv:1801.01290}, 2018.

\bibitem[Hong et~al.(2021)Hong, Wang, Wang, and Zhou]{hong2021federated}
J.~Hong, H.~Wang, Z.~Wang, and J.~Zhou.
\newblock Federated robustness propagation: Sharing adversarial robustness in
  federated learning.
\newblock \emph{arXiv preprint arXiv:2106.10196}, 2021.

\bibitem[Jabri et~al.(2019)Jabri, Hsu, Gupta, Eysenbach, Levine, and
  Finn]{jabri2019unsupervised}
A.~Jabri, K.~Hsu, A.~Gupta, B.~Eysenbach, S.~Levine, and C.~Finn.
\newblock Unsupervised curricula for visual meta-reinforcement learning.
\newblock \emph{Advances in Neural Information Processing Systems}, 32, 2019.

\bibitem[Ke et~al.(2021)Ke, Wang, Bhattacharjee, Boots, and
  Srinivasa]{ke2021grasping}
L.~Ke, J.~Wang, T.~Bhattacharjee, B.~Boots, and S.~Srinivasa.
\newblock Grasping with chopsticks: Combating covariate shift in model-free
  imitation learning for fine manipulation.
\newblock In \emph{2021 IEEE International Conference on Robotics and
  Automation (ICRA)}, pages 6185--6191. IEEE, 2021.

\bibitem[Kumar et~al.(2021)Kumar, Fu, Pathak, and Malik]{kumar2021rma}
A.~Kumar, Z.~Fu, D.~Pathak, and J.~Malik.
\newblock Rma: Rapid motor adaptation for legged robots.
\newblock \emph{arXiv preprint arXiv:2107.04034}, 2021.

\bibitem[Lee et~al.(2019)Lee, Eysenbach, Parisotto, Xing, Levine, and
  Salakhutdinov]{lee2019smm}
L.~Lee, B.~Eysenbach, E.~Parisotto, E.~P. Xing, S.~Levine, and
  R.~Salakhutdinov.
\newblock Efficient exploration via state marginal matching.
\newblock \emph{CoRR}, abs/1906.05274, 2019.
\newblock URL \url{http://arxiv.org/abs/1906.05274}.

\bibitem[Lin et~al.(2020)Lin, Thomas, Yang, and Ma]{lin2021robustmeta}
Z.~Lin, G.~Thomas, G.~Yang, and T.~Ma.
\newblock Model-based adversarial meta-reinforcement learning.
\newblock In H.~Larochelle, M.~Ranzato, R.~Hadsell, M.~Balcan, and H.~Lin,
  editors, \emph{Advances in Neural Information Processing Systems 33: Annual
  Conference on Neural Information Processing Systems 2020, NeurIPS 2020,
  December 6-12, 2020, virtual}, 2020.
\newblock URL
  \url{https://proceedings.neurips.cc/paper/2020/hash/73634c1dcbe056c1f7dcf5969da406c8-Abstract.html}.

\bibitem[Madry et~al.(2017)Madry, Makelov, Schmidt, Tsipras, and
  Vladu]{madry2017towards}
A.~Madry, A.~Makelov, L.~Schmidt, D.~Tsipras, and A.~Vladu.
\newblock Towards deep learning models resistant to adversarial attacks.
\newblock \emph{arXiv preprint arXiv:1706.06083}, 2017.

\bibitem[Margolis et~al.(2022)Margolis, Yang, Paigwar, Chen, and
  Agrawal]{margolis2022rapid}
G.~B. Margolis, G.~Yang, K.~Paigwar, T.~Chen, and P.~Agrawal.
\newblock Rapid locomotion via reinforcement learning.
\newblock \emph{arXiv preprint arXiv:2205.02824}, 2022.

\bibitem[Mendonca et~al.(2020)Mendonca, Geng, Finn, and
  Levine]{mendonca2020mier}
R.~Mendonca, X.~Geng, C.~Finn, and S.~Levine.
\newblock Meta-reinforcement learning robust to distributional shift via model
  identification and experience relabeling.
\newblock \emph{CoRR}, abs/2006.07178, 2020.
\newblock URL \url{https://arxiv.org/abs/2006.07178}.

\bibitem[Miki et~al.(2022)Miki, Lee, Hwangbo, Wellhausen, Koltun, and
  Hutter]{miki2022learning}
T.~Miki, J.~Lee, J.~Hwangbo, L.~Wellhausen, V.~Koltun, and M.~Hutter.
\newblock Learning robust perceptive locomotion for quadrupedal robots in the
  wild.
\newblock \emph{Science Robotics}, 7\penalty0 (62):\penalty0 eabk2822, 2022.

\bibitem[Mishra et~al.(2017)Mishra, Rohaninejad, Chen, and
  Abbeel]{mishra2017simple}
N.~Mishra, M.~Rohaninejad, X.~Chen, and P.~Abbeel.
\newblock A simple neural attentive meta-learner.
\newblock \emph{arXiv preprint arXiv:1707.03141}, 2017.

\bibitem[Mitchell et~al.(2021)Mitchell, Rafailov, Peng, Levine, and
  Finn]{mitchell2021offline}
E.~Mitchell, R.~Rafailov, X.~B. Peng, S.~Levine, and C.~Finn.
\newblock Offline meta-reinforcement learning with advantage weighting.
\newblock In \emph{International Conference on Machine Learning}, pages
  7780--7791. PMLR, 2021.

\bibitem[Nagabandi et~al.(2018)Nagabandi, Clavera, Liu, Fearing, Abbeel,
  Levine, and Finn]{nagabandi2018learning}
A.~Nagabandi, I.~Clavera, S.~Liu, R.~S. Fearing, P.~Abbeel, S.~Levine, and
  C.~Finn.
\newblock Learning to adapt in dynamic, real-world environments through
  meta-reinforcement learning.
\newblock \emph{arXiv preprint arXiv:1803.11347}, 2018.

\bibitem[Nair et~al.(2020)Nair, Gupta, Dalal, and Levine]{nair2020awac}
A.~Nair, A.~Gupta, M.~Dalal, and S.~Levine.
\newblock Awac: Accelerating online reinforcement learning with offline
  datasets.
\newblock \emph{arXiv preprint arXiv:2006.09359}, 2020.

\bibitem[Nesterov(2009)]{nesterov2009primal}
Y.~Nesterov.
\newblock Primal-dual subgradient methods for convex problems.
\newblock \emph{Mathematical programming}, 120\penalty0 (1):\penalty0 221--259,
  2009.

\bibitem[Ni et~al.(2022)Ni, Eysenbach, Levine, and
  Salakhutdinov]{ni2022recurrent}
T.~Ni, B.~Eysenbach, S.~Levine, and R.~Salakhutdinov.
\newblock Recurrent model-free {RL} is a strong baseline for many {POMDP}s,
  2022.
\newblock URL \url{https://openreview.net/forum?id=E0zOKxQsZhN}.

\bibitem[Oikarinen et~al.(2021)Oikarinen, Zhang, Megretski, Daniel, and
  Weng]{oikarinen2021robbust}
T.~P. Oikarinen, W.~Zhang, A.~Megretski, L.~Daniel, and T.~Weng.
\newblock Robust deep reinforcement learning through adversarial loss.
\newblock In M.~Ranzato, A.~Beygelzimer, Y.~N. Dauphin, P.~Liang, and J.~W.
  Vaughan, editors, \emph{Advances in Neural Information Processing Systems 34:
  Annual Conference on Neural Information Processing Systems 2021, NeurIPS
  2021, December 6-14, 2021, virtual}, pages 26156--26167, 2021.
\newblock URL
  \url{https://proceedings.neurips.cc/paper/2021/hash/dbb422937d7ff56e049d61da730b3e11-Abstract.html}.

\bibitem[Pinto et~al.(2017)Pinto, Davidson, Sukthankar, and
  Gupta]{pinto2017rarl}
L.~Pinto, J.~Davidson, R.~Sukthankar, and A.~Gupta.
\newblock Robust adversarial reinforcement learning.
\newblock In D.~Precup and Y.~W. Teh, editors, \emph{Proceedings of the 34th
  International Conference on Machine Learning, {ICML} 2017, Sydney, NSW,
  Australia, 6-11 August 2017}, volume~70 of \emph{Proceedings of Machine
  Learning Research}, pages 2817--2826. {PMLR}, 2017.
\newblock URL \url{http://proceedings.mlr.press/v70/pinto17a.html}.

\bibitem[Rakelly et~al.(2019)Rakelly, Zhou, Finn, Levine, and
  Quillen]{rakelly2019efficient}
K.~Rakelly, A.~Zhou, C.~Finn, S.~Levine, and D.~Quillen.
\newblock Efficient off-policy meta-reinforcement learning via probabilistic
  context variables.
\newblock In \emph{International conference on machine learning}, pages
  5331--5340. PMLR, 2019.

\bibitem[R{\'e}nyi(1961)]{renyi1961measures}
A.~R{\'e}nyi.
\newblock On measures of entropy and information.
\newblock In \emph{Proceedings of the Fourth Berkeley Symposium on Mathematical
  Statistics and Probability, Volume 1: Contributions to the Theory of
  Statistics}, volume~4, pages 547--562. University of California Press, 1961.

\bibitem[Rothfuss et~al.(2018)Rothfuss, Lee, Clavera, Asfour, and
  Abbeel]{rothfuss2018promp}
J.~Rothfuss, D.~Lee, I.~Clavera, T.~Asfour, and P.~Abbeel.
\newblock Promp: Proximal meta-policy search.
\newblock \emph{arXiv preprint arXiv:1810.06784}, 2018.

\bibitem[Schulman et~al.(2017)Schulman, Wolski, Dhariwal, Radford, and
  Klimov]{schulman2017ppo}
J.~Schulman, F.~Wolski, P.~Dhariwal, A.~Radford, and O.~Klimov.
\newblock Proximal policy optimization algorithms.
\newblock \emph{CoRR}, abs/1707.06347, 2017.
\newblock URL
  \url{http://dblp.uni-trier.de/db/journals/corr/corr1707.html#SchulmanWDRK17}.

\bibitem[Sinha et~al.(2017)Sinha, Namkoong, Volpi, and
  Duchi]{sinha2017certifying}
A.~Sinha, H.~Namkoong, R.~Volpi, and J.~Duchi.
\newblock Certifying some distributional robustness with principled adversarial
  training.
\newblock \emph{arXiv preprint arXiv:1710.10571}, 2017.

\bibitem[Sutton et~al.(1999)Sutton, McAllester, Singh, and
  Mansour]{sutton1999policy}
R.~S. Sutton, D.~McAllester, S.~Singh, and Y.~Mansour.
\newblock Policy gradient methods for reinforcement learning with function
  approximation.
\newblock \emph{Advances in neural information processing systems}, 12, 1999.

\bibitem[Thrun and Pratt(1998)]{thrun98metalearning}
S.~Thrun and L.~Y. Pratt, editors.
\newblock \emph{Learning to Learn}.
\newblock Springer, 1998.
\newblock ISBN 978-1-4613-7527-2.
\newblock \doi{10.1007/978-1-4615-5529-2}.
\newblock URL \url{https://doi.org/10.1007/978-1-4615-5529-2}.

\bibitem[Vaserstein(1969)]{vaserstein1969markov}
L.~N. Vaserstein.
\newblock Markov processes over denumerable products of spaces, describing
  large systems of automata.
\newblock \emph{Problemy Peredachi Informatsii}, 5\penalty0 (3):\penalty0
  64--72, 1969.

\bibitem[Vinitsky et~al.(2020)Vinitsky, Du, Parvate, Jang, Abbeel, and
  Bayen]{vinitsky2020robust}
E.~Vinitsky, Y.~Du, K.~Parvate, K.~Jang, P.~Abbeel, and A.~M. Bayen.
\newblock Robust reinforcement learning using adversarial populations.
\newblock \emph{CoRR}, abs/2008.01825, 2020.
\newblock URL \url{https://arxiv.org/abs/2008.01825}.

\bibitem[Wolpert and Macready(1997)]{585893}
D.~Wolpert and W.~Macready.
\newblock No free lunch theorems for optimization.
\newblock \emph{IEEE Transactions on Evolutionary Computation}, 1\penalty0
  (1):\penalty0 67--82, 1997.
\newblock \doi{10.1109/4235.585893}.

\bibitem[Wu et~al.(2021)Wu, Goodman, Piech, and Finn]{wu2021prototransformer}
M.~Wu, N.~Goodman, C.~Piech, and C.~Finn.
\newblock Prototransformer: A meta-learning approach to providing student
  feedback.
\newblock \emph{arXiv preprint arXiv:2107.14035}, 2021.

\bibitem[Xie et~al.(2022)Xie, Sodhani, Finn, Pineau, and Zhang]{xie2022robust}
A.~Xie, S.~Sodhani, C.~Finn, J.~Pineau, and A.~Zhang.
\newblock Robust policy learning over multiple uncertainty sets.
\newblock \emph{arXiv preprint arXiv:2202.07013}, 2022.

\bibitem[Zhang et~al.(2021{\natexlab{a}})Zhang, Chen, Boning, and
  Hsieh]{zhang2021robust}
H.~Zhang, H.~Chen, D.~S. Boning, and C.~Hsieh.
\newblock Robust reinforcement learning on state observations with learned
  optimal adversary.
\newblock In \emph{9th International Conference on Learning Representations,
  {ICLR} 2021, Virtual Event, Austria, May 3-7, 2021}. OpenReview.net,
  2021{\natexlab{a}}.
\newblock URL \url{https://openreview.net/forum?id=sCZbhBvqQaU}.

\bibitem[Zhang et~al.(2021{\natexlab{b}})Zhang, Wang, Hu, Chen, Chen, Fan, and
  Zhang]{zhang2021metacure}
J.~Zhang, J.~Wang, H.~Hu, T.~Chen, Y.~Chen, C.~Fan, and C.~Zhang.
\newblock Metacure: Meta reinforcement learning with empowerment-driven
  exploration.
\newblock In \emph{International Conference on Machine Learning}, pages
  12600--12610. PMLR, 2021{\natexlab{b}}.

\bibitem[Zhao et~al.(2020)Zhao, Nagabandi, Rakelly, Finn, and
  Levine]{zhao2020meld}
T.~Z. Zhao, A.~Nagabandi, K.~Rakelly, C.~Finn, and S.~Levine.
\newblock Meld: Meta-reinforcement learning from images via latent state
  models.
\newblock \emph{arXiv preprint arXiv:2010.13957}, 2020.

\bibitem[Zintgraf et~al.(2019)Zintgraf, Shiarlis, Igl, Schulze, Gal, Hofmann,
  and Whiteson]{zintgraf2019varibad}
L.~Zintgraf, K.~Shiarlis, M.~Igl, S.~Schulze, Y.~Gal, K.~Hofmann, and
  S.~Whiteson.
\newblock Varibad: A very good method for bayes-adaptive deep rl via
  meta-learning.
\newblock \emph{arXiv preprint arXiv:1910.08348}, 2019.

\bibitem[Zintgraf et~al.(2021)Zintgraf, Feng, Lu, Igl, Hartikainen, Hofmann,
  and Whiteson]{zintgraf2021exploration}
L.~M. Zintgraf, L.~Feng, C.~Lu, M.~Igl, K.~Hartikainen, K.~Hofmann, and
  S.~Whiteson.
\newblock Exploration in approximate hyper-state space for meta reinforcement
  learning.
\newblock In \emph{International Conference on Machine Learning}, pages
  12991--13001. PMLR, 2021.

\end{thebibliography}

%%%%%%%%%%%%%%%%%%%%%%%%%%%%%%%%%%%%%%%%%%%%%%%%%%%%%%%%%%%%
\section*{Checklist}

%%% BEGIN INSTRUCTIONS %%%
% The checklist follows the references.  Please
% read the checklist guidelines carefully for information on how to answer these
% questions.  For each question, change the default \answerTODO{} to \answerYes{},
% \answerNo{}, or \answerNA{}.  You are strongly encouraged to include a {\bf
% justification to your answer}, either by referencing the appropriate section of
% your paper or providing a brief inline description.  For example:
% \begin{itemize}
%   \item Did you include the license to the code and datasets? \answerYes{See Section [].}
%   \item Did you include the license to the code and datasets? \answerNo{The code and the data are proprietary.}
%   \item Did you include the license to the code and datasets? \answerNA{}
% \end{itemize}
% Please do not modify the questions and only use the provided macros for your
% answers.  Note that the Checklist section does not count towards the page
% limit.  In your paper, please delete this instructions block and only keep the
% Checklist section heading above along with the questions/answers below.
%%% END INSTRUCTIONS %%%

\begin{enumerate}

\item For all authors...
\begin{enumerate}
  \item Do the main claims made in the abstract and introduction accurately reflect the paper's contributions and scope?
    \answerYes{}
  \item Did you describe the limitations of your work?
    \answerYes{See Section~\ref{sec:discussion}}
  \item Did you discuss any potential negative societal impacts of your work?
    \answerNA{Our work is done in simulation and won't have any negative societal impact.}
  \item Have you read the ethics review guidelines and ensured that your paper conforms to them?
    \answerYes{This work does not actually use human subjects, and is done in simulation. We have reviewed ethics guidelines and ensured that our paper conforms to them.}
\end{enumerate}

\item If you are including theoretical results...
\begin{enumerate}
  \item Did you state the full set of assumptions of all theoretical results?
    \answerNA{Math is used as a theory/formalism, but we don't make any provable claims about it.}
	\item Did you include complete proofs of all theoretical results?
    \answerNA{}
\end{enumerate}

\item If you ran experiments...
\begin{enumerate}
  \item Did you include the code, data, and instructions needed to reproduce the main experimental results (either in the supplemental material or as a URL)?
    \answerYes{We have included the code along with a README in the supplemental material}
  \item Did you specify all the training details (e.g., data splits, hyperparameters, how they were chosen)?
    \answerYes{See Appendix~\ref{sec:hyperparams}} 
	\item Did you report error bars (e.g., with respect to the random seed after running experiments multiple times)?
    \answerYes{All plots were created with 4 random seeds with std error bars.} 
	\item Did you include the total amount of compute and the type of resources used (e.g., type of GPUs, internal cluster, or cloud provider)?
    \answerYes{See Appendix~\ref{sec:hyperparams}}
\end{enumerate}

\item If you are using existing assets (e.g., code, data, models) or curating/releasing new assets...
\begin{enumerate}
  \item If your work uses existing assets, did you cite the creators?
    \answerYes{Environments we used are cited in section \ref{sec:experiments}. Codebase used are cited in Appendix~\ref{sec:hyperparams}}
  \item Did you mention the license of the assets?
    \answerNA{}
  \item Did you include any new assets either in the supplemental material or as a URL?
    \answerYes{We published the code and included all environments and assets as a part of this}
  \item Did you discuss whether and how consent was obtained from people whose data you're using/curating?
    \answerYes{Environments and codebases we used are open-source.}
  \item Did you discuss whether the data you are using/curating contains personally identifiable information or offensive content?
    \answerNA{}
\end{enumerate}

\item If you used crowdsourcing or conducted research with human subjects...
\begin{enumerate}
  \item Did you include the full text of instructions given to participants and screenshots, if applicable?
    \answerNA{}
  \item Did you describe any potential participant risks, with links to Institutional Review Board (IRB) approvals, if applicable?
    \answerNA{}
  \item Did you include the estimated hourly wage paid to participants and the total amount spent on participant compensation?
    \answerNA{}
\end{enumerate}

\end{enumerate}

\newpage
\appendix
\section{\rml~for handling out-of-support task distribution shifts}
\label{sec:rml_vae_oos}
To handle \textit{out-of-support} task distribution shifts, we parameterized new task distribution as latent space distributions $q_\phi(z)$ and measure the divergence between train and test task distribution via KL divergence in latent space $D(p_{\text{train}}(z)||q_\phi(z))$ in Section~\ref{sec:practical}. Using this parameterization, equation~\ref{eq:robust_meta_train} becomes
\begin{align}
     \max_\theta &\min_\phi \E_{z \sim q_\phi(z)}\left[\E_{\pi_{\text{meta},\theta}^\epsilon,\gP}\left[\frac{1}{k}\sum_{i=1}^k\sum_{t=1}^T \textcolor{purple}{r_\omega(s_t^{(i)}, a_t^{(i)},z)}\right]\right] \;\;\; \textcolor{purple}{\text{when rewards differ}}\nonumber\\
    \max_\theta &\min_\phi \E_{z \sim q_\phi(z)}\left[\E_{\pi_{\text{meta},\theta}^\epsilon,\textcolor{brown}{p_\omega(\cdot,\cdot,z)}}\left[\frac{1}{k}\sum_{i=1}^k\sum_{t=1}^T r_t^{(i)}\right]\right] \;\;\; \textcolor{brown}{\text{when dynamics differ}}\nonumber\\
    &~D_{\text{KL}}(p_{\text{train}}(z) || q_\phi(z)) \leq \epsilon
\label{eq:gen_robust_meta_train}
\end{align}
This objective function is solved in Algorithm~\ref{alg:meta_train_code} for different values of $\epsilon$. To imagine \textit{out-of-support} distributionally shifted task (i.e. reward or dynamics) distributions, \rml~leverages structured VAE which we describe in subsequent subsections.

\subsection{Structured VAE for modeling reward distributions}
\label{sec:structure_vae}
\begin{wrapfigure}{r}{0.4\linewidth}
    \centering
    \includegraphics[width=0.6\textwidth]{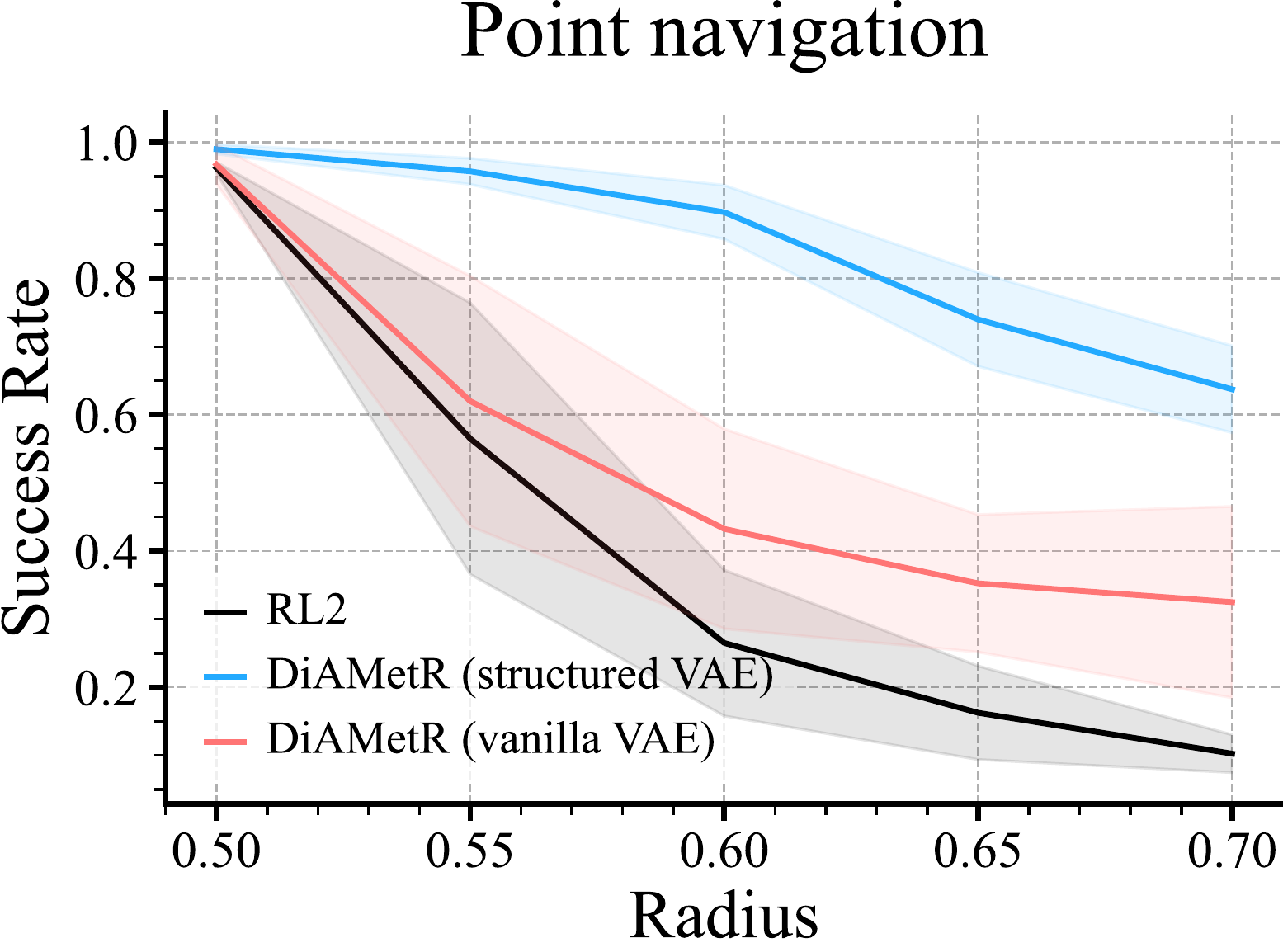}
    \caption{Using a vanilla VAE, in lieu of a structured VAE, to model task distribution hurts \rml's performance on test-task distributions.}
    \label{fig:vae_abl}
\end{wrapfigure}
We leverage the sparsity in reward functions (i.e. $0/1$ rewards) in the environments used and describe a structured VAE to model $r_\omega(s,a,z)$ with $p(z) = \gN(0,I)$ and KL-divergence for $D(\cdot||\cdot)$. Let $\bar{h} = (\sum_{t=1}^T r_t s_t)/(\sum_{t=1}^T r_t)$ be the mean of states achieving a $+1$ reward in trajectory $h$. The encoder $z \sim q_\psi(z|\bar{h})$ encodes $\bar{h}$ into a latent vector $z$. The reward model $r_\omega(s,a,z)$ consists of 2 components: (i) latent decoder $\hat{\bar{h}} = r^{h}_\omega(z)$ which reconstructs $\bar{h}$ and (ii) reward predictor $r^{\text{rew}}_\omega(s, \hat{\bar{h}}) = \exp(-\|M\odot(s-\hat{\bar{h}}) \|_2^2/\sigma^2)$ which predicts reward for a state given the decoded latent vector. $M$ is a masking function and $\sigma$ is a learned parameter. The training objective becomes
\begin{equation}
    \min_{\omega, \psi} \E_{h\sim\gD}\left[\E_{z \sim q_\psi(z|\bar{h})}\left[\left\|\bar{h} - r^{h}_\omega(z)\right\|_2^2 + \sum_{t=1}^T\left\|r^{\text{rew}}_\omega(s_t, r^{h}_\omega(z)) - r_t\right\|^2\right] + D_{\text{KL}}(q_\psi(z|\bar{h}) || p(z))\right]
    \label{eq:prac_vae}
\end{equation}
The structure in the VAE helps in extrapolating reward functions when $z \sim q_\phi(z)$. This can be further verified by reduction in \rml's performance on test-task distributions when using vanilla VAE (see Figure~\ref{fig:vae_abl}).

\subsection{Structured VAE for modelling dynamics distributions}
\label{subsec:struct_dyn_vae}
We describe our structured VAE architecture for modelling dynamics distribution $p_\omega(s,a,z)$ with $p(z)=\gN(0,I)$ and KL-divergence for $D(\cdot||\cdot)$. It handles \textit{out-of-support} shifted test task distributions where only dynamics vary across tasks. We leverage the fact that dynamics differ by an additive term. The encoder $z \sim q_\phi(z|(s_t, a_t)_{t=1}^T)$ encodes the state action trajectory to a latent vector $z$. The dynamics model takes the form $p_\omega(s,a,z) = W_\omega z + p_\omega^{\text{dyn}}(s,a)$ where $W_\omega$ is a parameter. The training objective becomes:
\begin{align}
    \min_{\omega, \psi} \E_{(s_t, a_t)_{t=1}^T\sim\gD}\left[\E_{z \sim q_\psi(z|(s_t, a_t)_{t=1}^T)}\left[ \sum_{t=1}^{T-1}\left\|p^{\text{dyn}}_\omega(s_t, a_t) + W_\omega z - s_{t+1}\right\|^2\right]
    + D_{\text{KL}}(q_\psi(z|(s_t, a_t)_{t=1}^T) || p(z))\right]
    \label{eq:prac_vae_dyn}
\end{align}
The structure in the VAE helps in extrapolating dynamics when $z \sim q_\phi(z)$.

\section{\rml~for handling in-support task distribution shifts}
% \begin{figure}[t]
%     \centering
%     \includegraphics[width=0.95\linewidth]{figs/diametr_train_test_comb.pdf}
%     \caption{\textcolor{violet}{During meta-train phase, \rml~learns a family of meta-policies robust to varying levels of distribution shift (as characterized by $\epsilon_i$). During meta-test phase, given a potentially shifted test-time distribution of tasks, \rml~chooses the meta-policy with the most appropriate level of robustness and use it to perform fast adaptation for new tasks sampled from the same shifted test task distribution.}}
%     \label{fig:diametr_diagram}
% \end{figure}
To handle \textit{in-support} task distribution shifts, we parameterized new task distribution as re-weighted (empirical) training task distribution $q_\phi(\gT) \propto w_\gT p_{\train}(\gT)$ (where $\phi=\{w_{\gT_i}\}_{i=1}^{n_\text{tr}}$) in Section~\ref{sec:practical}. Using this parameterization, equation~\ref{eq:robust_meta_train} becomes
\begin{align}
    \max_\theta &\min_\phi \E_{\gT\sim p_{\train}(\gT)}\left[\E_{\pi_{\text{meta},\theta}^\epsilon,\gP}\left[\frac{nw_\gT}{\sum_{i=1}^{n_{\text{tr}}} w_{\gT_i}}\frac{1}{k}\sum_{i=1}^k\sum_{t=1}^T r_t^{(i)}\right]\right] \nonumber\\
    &~D_{\text{KL}}(p_{\text{train}}(\gT) || q_\phi(\gT)) \leq \epsilon
\label{eq:weight_robust_meta_train}
\end{align}
This objective function is solved in Algorithm~\ref{alg:meta_train_code} for different values of $\epsilon$. For \textit{in-support} task distribution shifts, shifts in dynamics distribution and reward distribution don't require separate treatment.
\section{Test time Meta Policy Selection}

\begin{algorithm}[t]
    \small
    \caption{(Detailed) \textbf{\rml}:Meta-training phase}
    \label{alg:detail_meta_train_code}
    \begin{algorithmic}[1]
    \STATE Given: $p_{\text{train}}(\gT)$, Return: $\{\pi_{\text{meta}, \theta}^{\epsilon_i}\}_{i=1}^M$
    \STATE $\pi_{\text{meta}, \theta}^{\epsilon_1}$, $\gD_{\text{Replay-Buffer}} \leftarrow$ Solve equation~\ref{eq:meta_train} with off-policy $\text{RL}^2$
    \STATE \textcolor{purple}{Reward distribution $r_\omega$} / \textcolor{brown}{Dynamic distribution $p_\omega$}, prior $p_\train(z) \leftarrow$ Solve eq~\ref{eq:prac_vae_main} using $\gD_{\text{Replay-Buffer}}$ 
    % \STATE \textcolor{brown}{Dynamic distribution $p_\omega$, prior $p_\train(z) \leftarrow$ Solve eq~\ref{eq:prac_vae} using $\gD_{\text{Replay-Buffer}}$}
    \FOR{$\epsilon$ in $\{\epsilon_2,\ldots,\epsilon_M\}$}
        \STATE Initialize $q_\phi$, $\pi_{\text{meta}, \theta}^\epsilon$ and $\lambda \geq 0$.
      	\FOR{iteration $n=1, 2, ...$}
      	    \STATE \textbf{Meta-policy:} Update $\pi_{\text{meta},\theta}^\epsilon$ using off-policy $\text{RL}^2$~\cite{ni2022recurrent}
                \vspace{-0.2cm}
                \textcolor{blue}{
                \begin{equation}
                    \theta \coloneqq \theta + \alpha\nabla_\theta \E_{\gT \sim p_\train(\gT)}[\E_{\pi_{\text{meta},\theta}^\epsilon,\gT}[\frac{nw_\gT}{k\sum_{i=1}^{n_\text{tr}}w_{\gT_i}}\sum_{i=1}^k\sum_{t=1}^T r_t^{(i)}]]\nonumber
                \end{equation}}
                \vspace{-0.2cm}
                \textcolor{purple}{
                \begin{equation}
                    \theta \coloneqq \theta + \alpha\nabla_\theta \E_{z \sim q_\phi(z)}[\E_{\pi_{\text{meta},\theta}^\epsilon,\gP}[\frac{1}{k}\sum_{i=1}^k\sum_{t=1}^T r_\omega(s_t^{(i)}, a_t^{(i)}, z)]]\nonumber
                \end{equation}}
                \vspace{-0.2cm}
                \textcolor{brown}{
                \begin{equation}
                    \theta \coloneqq \theta + \alpha\nabla_\theta \E_{z \sim q_\phi(z)}[\E_{\pi_{\text{meta},\theta}^\epsilon,p_\omega(\cdot, \cdot, z)}[\frac{1}{k}\sum_{i=1}^k\sum_{t=1}^T r_t^{(i)}]]\nonumber
                \end{equation}}
                \vspace{-0.2cm}
            \STATE \textbf{Adversarial task distribution:} Update $q_\phi$ using Reinforce~\cite{sutton1999policy}
            \textcolor{blue}{
            \begin{equation}
                \phi \coloneqq \phi - \alpha\nabla_\phi (\E_{\gT \sim p_\train(\gT)}[\E_{\pi_{\text{meta},\theta}^\epsilon,\gT}[\frac{nw_\gT}{k\sum_{i=1}^{n_\text{tr}}w_{\gT_i}}\sum_{i=1}^k\sum_{t=1}^T r_t^{(i)}]] + \lambda D_{\text{KL}}(p_\train(\gT)\|q_\phi(\gT))\nonumber
            \end{equation}}
            \textcolor{purple}{
            \begin{equation}
                \phi \coloneqq \phi - \alpha\nabla_\phi (\E_{z \sim q_\phi(z)}[\E_{\pi_{\text{meta},\theta}^\epsilon,\gP}[\frac{1}{k}\sum_{i=1}^k\sum_{t=1}^T r_\omega(s_t^{(i)}, a_t^{(i)}, z)]] + \lambda D_{\text{KL}}(p_\train(z)\|q_\phi(z))\nonumber
            \end{equation}}
            \textcolor{brown}{
            \begin{equation}
                \phi \coloneqq \phi - \alpha\nabla_\phi (\E_{z \sim q_\phi(z)}[\E_{\pi_{\text{meta},\theta}^\epsilon,p_\omega(\cdot, \cdot, z)}[\frac{1}{k}\sum_{i=1}^k\sum_{t=1}^T r_t^{(i)}]] + \lambda D_{\text{KL}}(p_\train(z)\|q_\phi(z))\nonumber
            \end{equation}}
      	    \STATE \textbf{Lagrange constraint multiplier:} Update $\lambda$ to enforce $D_{\text{KL}}(p_\train \|q_\phi) < \epsilon$,
      	    \textcolor{blue}{
      	    \begin{equation}
                \lambda \coloneqq_{\lambda \geq 0} \lambda + \alpha (D_{\text{KL}}(p_\train(\gT)\|q_\phi(\gT)) - \epsilon)\nonumber
            \end{equation}}
            \vspace{-0.5cm}
            \begin{equation}
                \textcolor{purple}{\lambda \coloneqq_{\lambda \geq 0} \lambda + \alpha (D_{\text{KL}}(p_\train(z)\|q_\phi(z)) - \epsilon)} \;\;\;\; \textcolor{brown}{\lambda \coloneqq_{\lambda \geq 0} \lambda + \alpha (D_{\text{KL}}(p_\train(z)\|q_\phi(z)) - \epsilon)}\nonumber
            \end{equation}
      	\ENDFOR
    %   	\STATE Add $\pi_{\text{meta}, \theta}^\epsilon$ to $\Pi$
    \ENDFOR
    \end{algorithmic}
    % \vspace{-0.5cm}
\end{algorithm}

\label{sec:test_time_select}
As discussed in Section~\ref{sec:approach}, to adapt to test time task distribution shifts, we train a family of meta-policies $\Pi = \{\pi_\text{meta}^{\epsilon_i}\}$ to be robust to varying degrees of distribution shifts. We then choose the appropriate meta-policy during test-time based on the inferred task distribution shift. In this section, we frame the test-time selection of meta-policy from the family $\Pi$ as a stochastic multi-arm bandit problem. Every iteration involves pulling an arm $i$ which corresponds to executing $\pi_\text{meta}^{\epsilon_i}$ for $1$ meta-episode ($k$ environment episodes) on a task $\gT\sim p_\testt(\gT)$. Let $R_i$ be the expected return for pulling arm $i$
\begin{equation}
    R_i = \E_{\pi_\text{meta}^{\epsilon_i}, \gT\sim p_\testt(\gT)}\left[\frac{1}{k}\sum_{i=1}^k\sum_{t=1}^T r_t^{(i)} \right]
\end{equation}
Let $R^* = \max_{i\in\{1,\ldots,M\}} R_i$ and $\pi_\text{meta}^{\epsilon}$ be the corresponding meta-policy. The goal of the stochastic bandit problem is to pull arms $i_1, \ldots, i_N \in \{1,\ldots,M\}$ such that the test-time regret $\gR_N$ is minimized
\begin{equation}
    \gR_N^\testt = NR^* - \sum_{t=1}^N R_{i_t}
\end{equation}
with constraint that $i_t$ can depend only on the information available prior to iteration $t$. We choose Thompson sampling, a zero-regret bandit algorithm, to solve this problem. In principle, Thompson sampling should learn to choose $\pi_\text{meta}^{\epsilon}$ after $N$ iterations.
\section{Environment Description}
\label{sec:env_desc}
We describe the environments used in the paper:
\begin{itemize}[leftmargin=*]
    \item \texttt{\{Point,Wheeled,Ant\}-navigation}: The rewards for each task correspond to reaching an unobserved target location $s_{\gT}$. The agent (i.e. Wheeled, Ant) must explore the environment to find the unobserved target location (Wheeled driving a differential drive robot, Ant controlling a four legged robotic quadruped). It receives a reward of $1$ once it gets within a small $\delta$ distance of the target $s_t$, as in ~\cite{gupta2018meta}.
    \item \texttt{Dense Ant-navigation}: The rewards for each task correspond to reaching an unobserved target location $s_{\gT}$. The Ant (a four legged robotic quadruped) must explore the environment to find the unobserved target location. It receives a reward of $-\|\text{agent}_{x,y} - s_\gT\|$ where $\text{agent}_{x,y}$ is the (x,y) position of the Ant.
    \item \texttt{Wind navigation}: The rewards for each task correspond to reaching an unobserved target location $s_t$. While $s_{\gT}$ is fixed across tasks (i.e. at $(1/\sqrt{2},1/\sqrt{2})$), the agent (a linear system robot) much navigate in the presence of wind (i.e. a noise vector $w_{\gT}$) that varies across tasks. It receives a reward of $1$ once it gets within a small $\delta$ distance of the target $s_{\gT}$.
    \item \texttt{Object localization}: Each task corresponds to using the gripper to localize an object kept at an unobserved target location $s_t$. The Fetch robot must move its gripper around and explore the environment to find the object. Once the gripper touches the object kept at target location $s_t$, it receives a reward of $1$.
    \item \texttt{Block push}: Each task corresponds to moving the block to an unobserved target location $s_t$. The robot arm must move the block around and explore the environment to find the unobserved target location. Once the block gets within a small $\delta$ distance of the target $s_t$, it receives a reward of $1$, as in ~\cite{gupta2018meta}.
\end{itemize}

Furthermore, Table~\ref{tbl:env_desc} describes the state space $\gS$, action space $\gA$, episodic horizon $H$, frameskip for each environment and $k$ (i.e. number of environment episodes in $1$ meta episode). Table~\ref{tbl:detail_task_dist} provides parameters for train and test task distributions for different meta-RL tasks used in the paper.

\begin{table}[H]
  \caption{Environment Description}
  \label{tbl:env_desc}
  \centering
  \begin{tabular}{lllllll}
    \toprule
    Name  & State space $\gS$ & Action space $\gA$ & Episodic Horizon $H$ & Frameskip & $k$\\
    \midrule
    \texttt{Point-navigation} & Box(2,) & Box(2,) & 60 & 1 & 2\\
    \texttt{Wind-navigation} & Box(2,) & Box(2,) & 25 & 1 & 1\\
    \texttt{Wheeled-navigation} & Box(12,) & Box(2,) & 60 & 10 & 2\\
    \texttt{Ant-navigation} & Box(29,) & Box(8,) & 200 & 5 & 2\\
    \texttt{Dense Ant-navigation} & Box(29,) & Box(8,) & 200 & 5 & 2\\
    \texttt{Object localization} & Box(17,) & Box(6,) & 50 & 10 & 2\\
    \texttt{Block Push} & Box(10,) & Box(4,) & 60 & 10 & 2\\
    \bottomrule
  \end{tabular}
\end{table}

\begin{table}[h]
  \caption{Parameters for train and test task distribution for \texttt{\{Point,Wheeled, Ant\}-navigation}, \texttt{Dense Ant-navigation}, \texttt{Wind-navigation}, \texttt{Object localization} and \texttt{Block-push}. While tasks in \texttt{Wind-navigation} vary in dynamics, tasks in other environments vary in reward function. The shifted test task distributions can be either \textit{in-support} or \textit{out-of-support} of the training task distribution. All these task distributions are determined by distributions of underlying task parameters (say target location $s_\gT$ or wind velocity $w_\gT$), which either determine the reward function or the dynamics function.}
  \label{tbl:detail_task_dist}
  \small
  \centering
    \begin{tabular}{|m{2.5cm}|m{2cm}|m{3cm}|m{2.5cm}|m{4cm}|}
    \hline
    Environment & Task type & Task parameter distribution & $p_\train(\gT)$ & $\{p_\testt^i(\gT)\}_{i=1}^K$\\
    \hline
    \shortstack{\texttt{Point, Wheeled,}\\\texttt{Ant-navigation}} & \shortstack{reward change, \\out-of-support\\ shift} & \shortstack{$s_\gT=(\Delta\cos\theta, \Delta\sin\theta)$\\ $\Delta\sim\mathcal{U}(\Delta_{\min}, \Delta_{\max})$\\ $\theta\sim\mathcal{U}(0, 2\pi)$} & \shortstack{$\Delta\sim\mathcal{U}(0, 0.5)$} & \shortstack{$\Delta\sim\mathcal{U}(0, 0.5), \mathcal{U}(0.5, 0.55)$\\$\mathcal{U}(0.55, 0.6),\mathcal{U}(0.6, 0.65)$ \\$\mathcal{U}(0.65, 0.7)$}\\
    \hline
    \shortstack{\texttt{Point, Wheeled,}\\\texttt{Ant-navigation}} & \shortstack{reward change, \\in-support shift} & \shortstack{$s_\gT=(\Delta\cos\theta, \Delta\sin\theta)$\\ $\Delta\sim\mathcal{U}(\Delta_{\min}, \Delta_{\max})$\\ $\theta\sim\mathcal{U}(0, 2\pi)$} & \shortstack{$\Delta\sim\text{Exp}(\lambda=5)$} & \shortstack{$\Delta\sim\mathcal{U}(0, 0.5), \mathcal{U}(0.5, 0.55)$\\$\mathcal{U}(0.55, 0.6),\mathcal{U}(0.6, 0.65)$ \\$\mathcal{U}(0.65, 0.7)$}\\
    \hline
    \shortstack{\texttt{Dense}\\\texttt{Ant-navigation}} & \shortstack{reward change, \\in-support shift} & \shortstack{$s_\gT=(\Delta\cos\theta, \Delta\sin\theta)$\\ $\Delta\sim\mathcal{U}(\Delta_{\min}, \Delta_{\max})$\\ $\theta\sim\mathcal{U}(0, 2\pi)$} & \shortstack{$\Delta\sim\text{Exp}(\lambda=5)$} & \shortstack{$\Delta\sim\mathcal{U}(0, 0.5), \mathcal{U}(0.5, 0.55)$\\$\mathcal{U}(0.55, 0.6),\mathcal{U}(0.6, 0.65)$ \\$\mathcal{U}(0.65, 0.7)$}\\
    \hline
    \shortstack{\texttt{Wind}\\\texttt{-navigation}} & \shortstack{dynamics change,\\out-of-support\\shift} & \shortstack{$w_\gT=(\Delta\cos\theta, \Delta\sin\theta)$\\ $\Delta\sim\mathcal{U}(\Delta_{\min}, \Delta_{\max})$\\ $\theta\sim\mathcal{U}(0, 2\pi)$} & \shortstack{$\Delta\sim\mathcal{U}(0, 0.05)$} & \shortstack{$\Delta\sim\mathcal{U}(0, 0.05), \mathcal{U}(0.05, 0.06)$\\$\mathcal{U}(0.06, 0.07), \mathcal{U}(0.07, 0.08)$\\$\mathcal{U}(0.08, 0.09)$}\\
    \hline
    \shortstack{\texttt{Wind}\\\texttt{-navigation}} & \shortstack{dynamics change,\\in-support shift} & \shortstack{$w_\gT=(\Delta\cos\theta, \Delta\sin\theta)$\\ $\Delta\sim\mathcal{U}(\Delta_{\min}, \Delta_{\max})$\\ $\theta\sim\mathcal{U}(0, 2\pi)$} & \shortstack{$\Delta\sim\text{Exp}(\lambda=40)$} & \shortstack{$\Delta\sim\mathcal{U}(0, 0.05), \mathcal{U}(0.05, 0.06)$\\$\mathcal{U}(0.06, 0.07), \mathcal{U}(0.07, 0.08)$\\$\mathcal{U}(0.08, 0.09)$}\\
    \hline
    \shortstack{\texttt{Object}\\\texttt{localization}} & \shortstack{reward change, \\out-of-support\\ shift} & \shortstack{$s_\gT=(\Delta\cos\theta, \Delta\sin\theta)$\\ $\Delta\sim\mathcal{U}(\Delta_{\min}, \Delta_{\max})$\\ $\theta\sim\mathcal{U}(0, 2\pi)$} & \shortstack{$\Delta\sim\mathcal{U}(0, 0.1)$} & \shortstack{$\Delta\sim\mathcal{U}(0, 0.1), \mathcal{U}(0.1, 0.12)$\\$\mathcal{U}(0.12, 0.14), \mathcal{U}(0.14, 0.16)$\\$\mathcal{U}(0.16, 0.18), \mathcal{U}(0.18, 0.2)$}\\
    \hline
    \shortstack{\texttt{Object}\\\texttt{localization}} & \shortstack{reward change, \\in-support shift} & \shortstack{$s_\gT=(\Delta\cos\theta, \Delta\sin\theta)$\\ $\Delta\sim\mathcal{U}(\Delta_{\min}, \Delta_{\max})$\\ $\theta\sim\mathcal{U}(0, 2\pi)$} & \shortstack{$\Delta\sim\text{Exp}(\lambda=20
    )$} & \shortstack{$\Delta\sim\mathcal{U}(0, 0.1), \mathcal{U}(0.1, 0.12)$\\$\mathcal{U}(0.12, 0.14), \mathcal{U}(0.14, 0.16)$\\$\mathcal{U}(0.16, 0.18)$}\\
    \hline
    \shortstack{\texttt{Block-push}} & \shortstack{reward change, \\out-of-support\\ shift} & \shortstack{$s_\gT=(\Delta\cos\theta, \Delta\sin\theta)$\\ $\Delta\sim\mathcal{U}(\Delta_{\min}, \Delta_{\max})$\\ $\theta\sim\mathcal{U}(0, \pi/2)$} & \shortstack{$\Delta\sim\mathcal{U}(0, 0.5)$} & \shortstack{$\Delta\sim\mathcal{U}(0, 0.5), \mathcal{U}(0.5, 0.6)$\\$\mathcal{U}(0.6, 0.7), \mathcal{U}(0.7, 0.8)$\\$\mathcal{U}(0.8, 0.9), \mathcal{U}(0.9, 1.0)$}\\
    \hline
    \shortstack{\texttt{Block-push}} & \shortstack{reward change, \\in-support shift} & \shortstack{$s_\gT=(\Delta\cos\theta, \Delta\sin\theta)$\\ $\Delta\sim\mathcal{U}(\Delta_{\min}, \Delta_{\max})$\\ $\theta\sim\mathcal{U}(0, \pi/2)$} & \shortstack{$\Delta\sim\text{Exp}(\lambda=4
    )$} & \shortstack{$\Delta\sim\mathcal{U}(0, 0.5), \mathcal{U}(0.5, 0.6)$\\$\mathcal{U}(0.6, 0.7), \mathcal{U}(0.7, 0.8)$\\$\mathcal{U}(0.8, 0.9), \mathcal{U}(0.9, 1.0)$}\\
    \hline
  \end{tabular}
\end{table}
\section{Experimental Evaluation on Wheeled and Point Robot Navigation}
\label{sec:point_exps}
In Section~\ref{sec:experiments}, we evaluated \rml~on \texttt{Wind-navigation}, \texttt{Ant-navigation}, \texttt{Fetch-reach} and \texttt{Block-push}. We continue the experimental evaluation of \rml~in this section and compare it to $\text{RL}^2$, VariBAD, and HyperX on train task distribution and different test task distributions of \texttt{Point navigation} and \texttt{Wheeled navigation}~\cite{gupta2018meta}. We see that \rml~either matches or outperforms the baselines on train task distribution and outperforms the baselines on test task distributions. Furthermore, adaptively selecting an uncertainty set during test time allows for better test time distribution adaptation when compared to selecting an uncertainty set beforehand or selecting a large uncertainty set.

\begin{figure}[!t]
    \centering
    \includegraphics[width=0.24\linewidth]{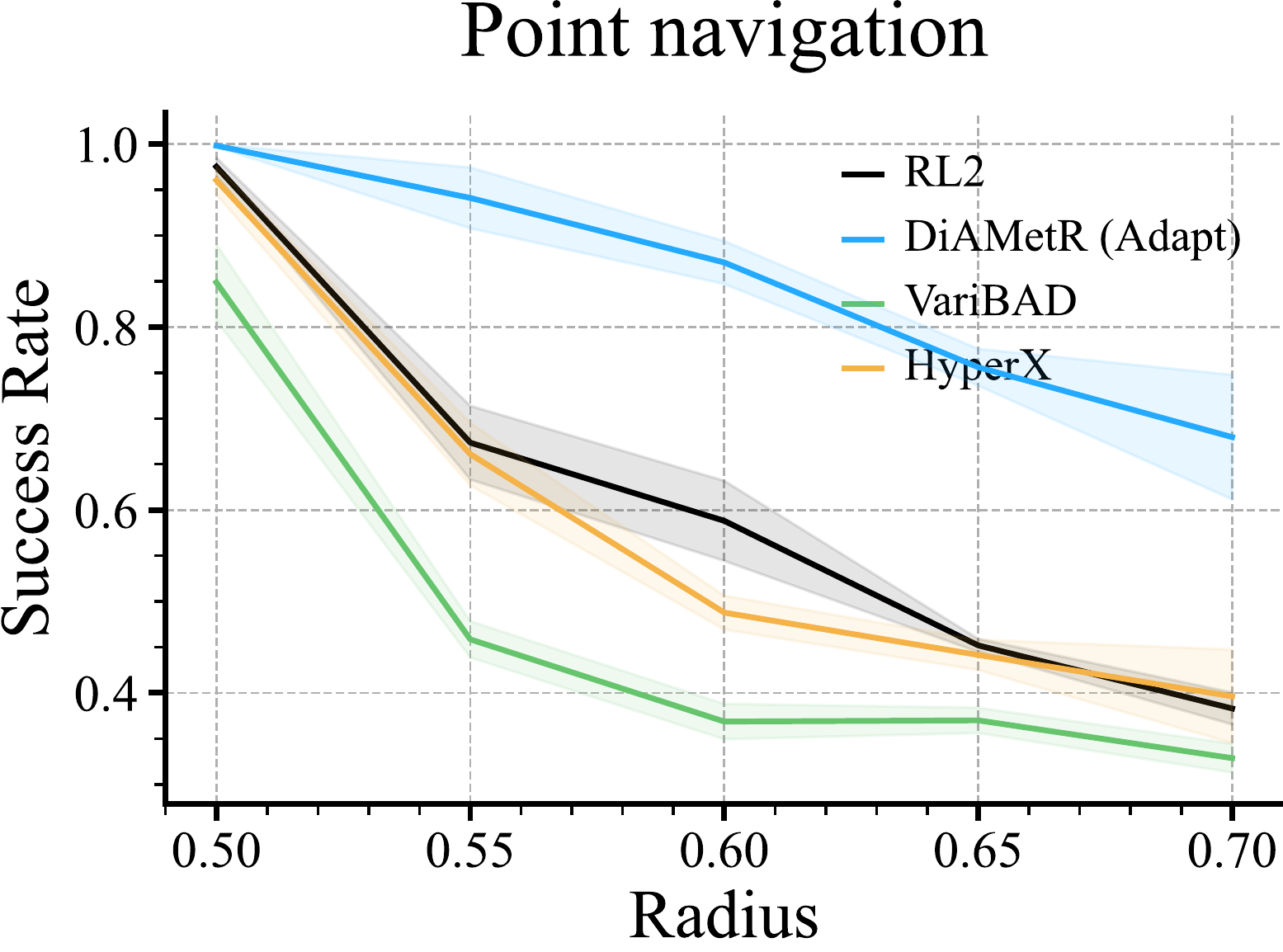}
    \includegraphics[width=0.24\linewidth]{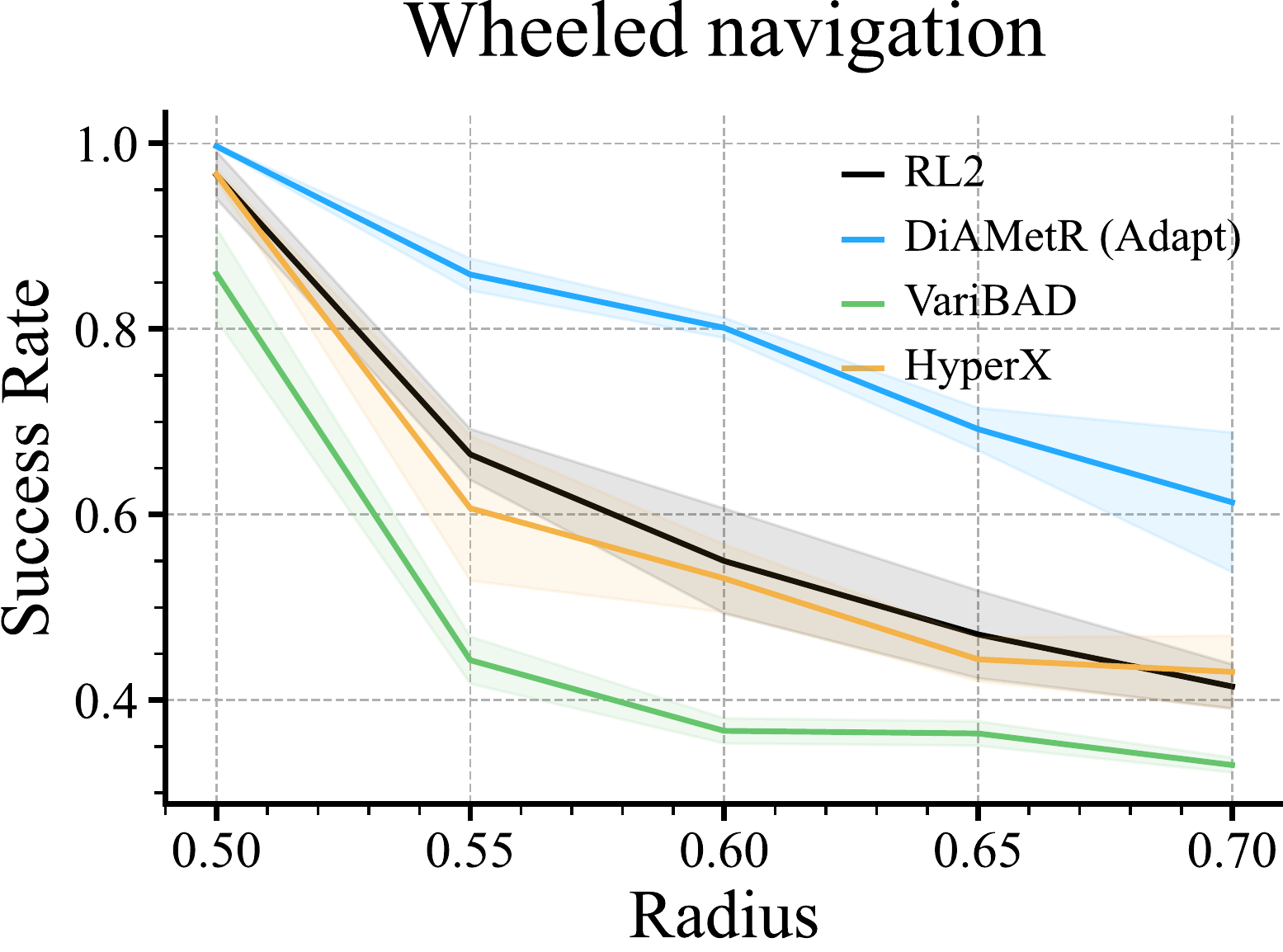}
    \includegraphics[width=0.24\linewidth]{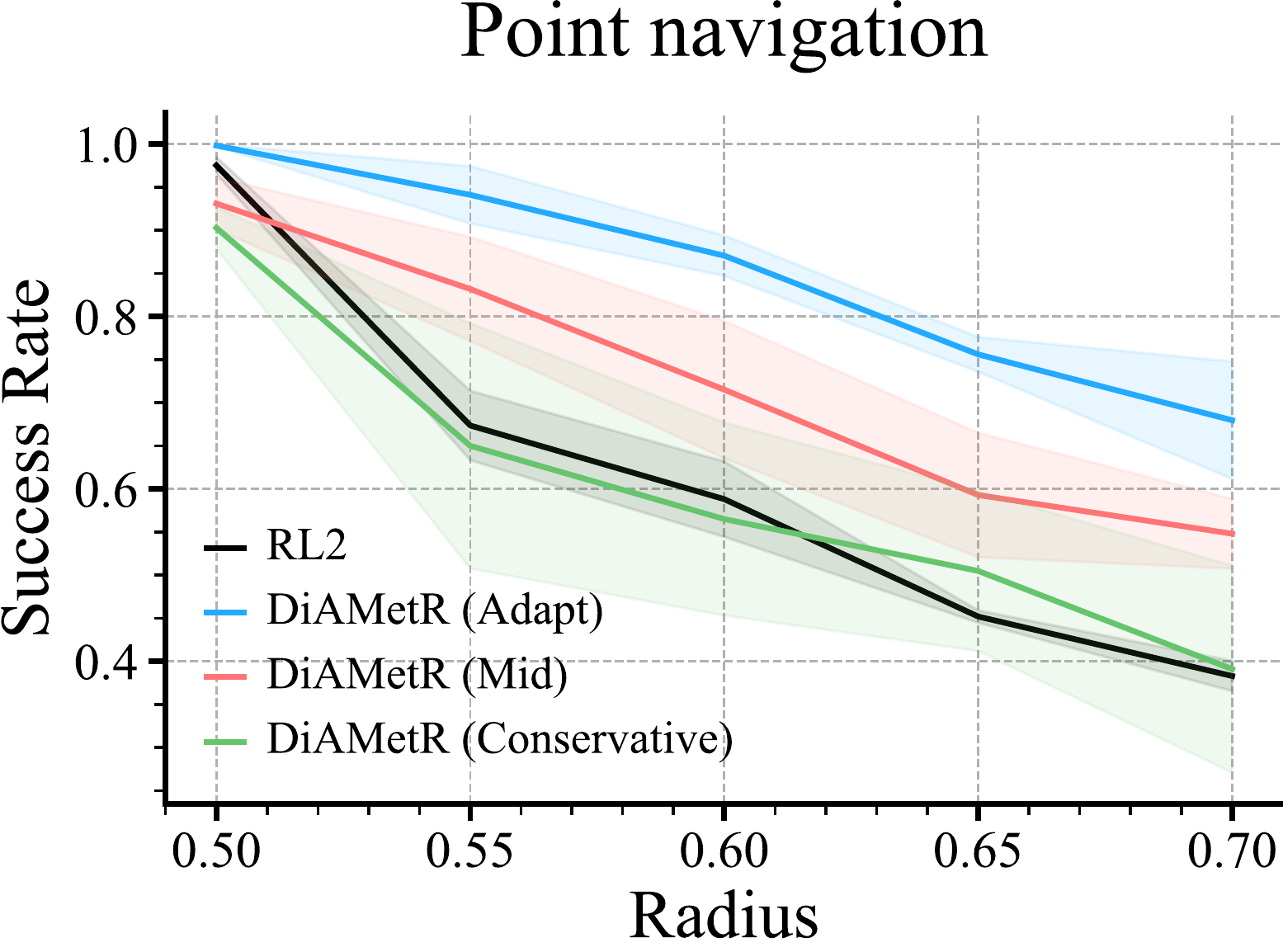}
    \includegraphics[width=0.24\linewidth]{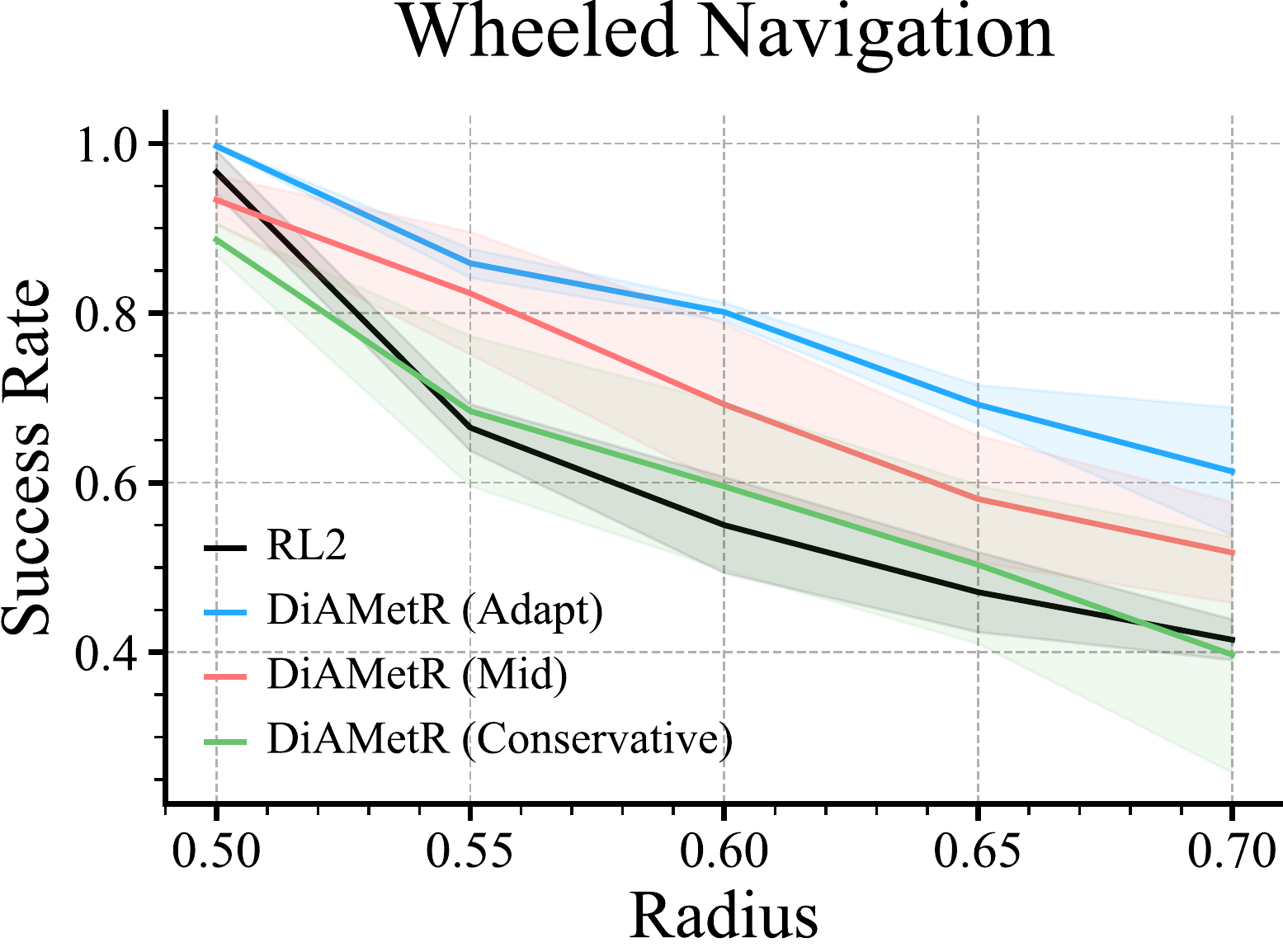}\\
    \includegraphics[width=0.24\linewidth]{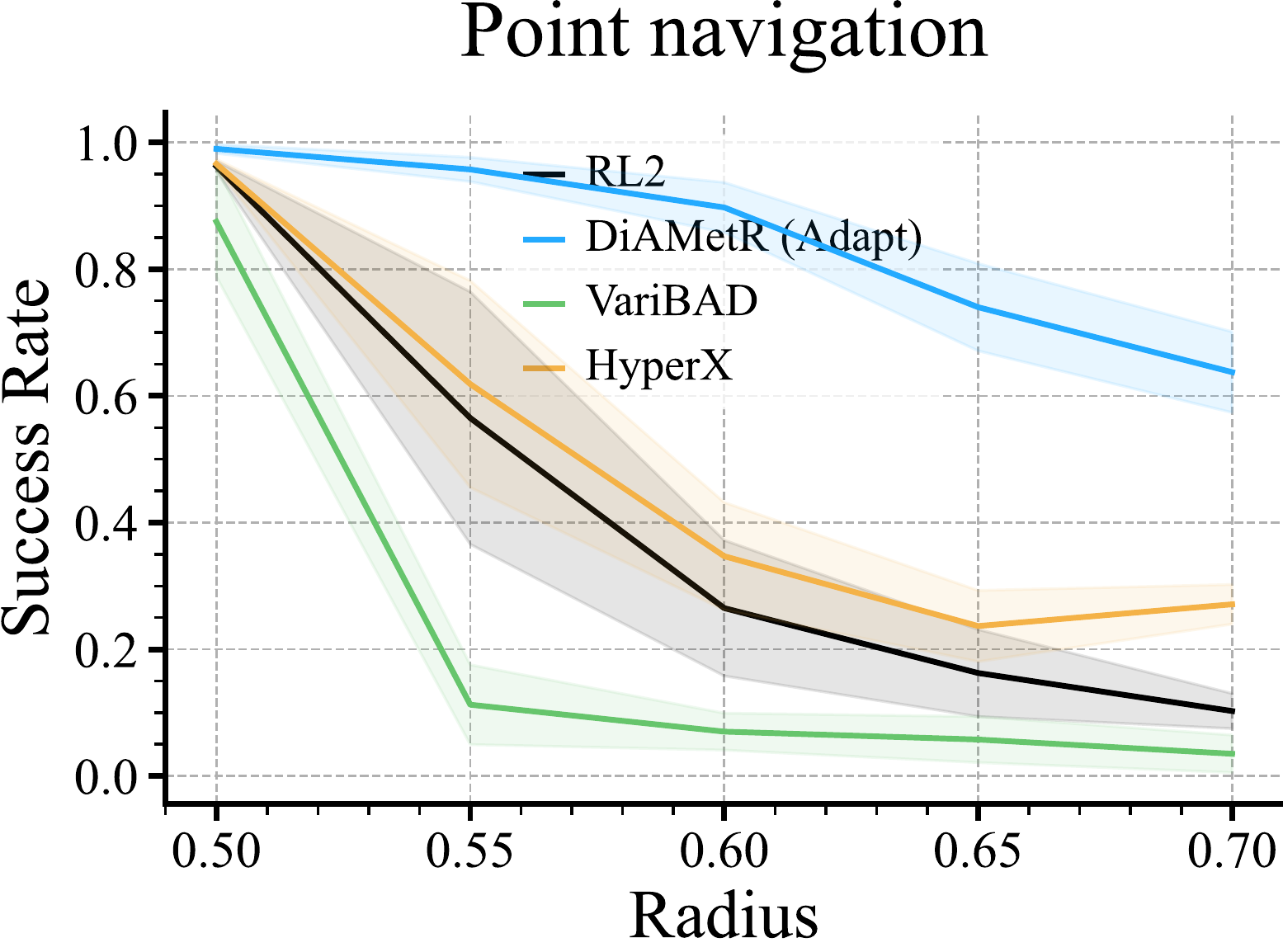}
    \includegraphics[width=0.24\linewidth]{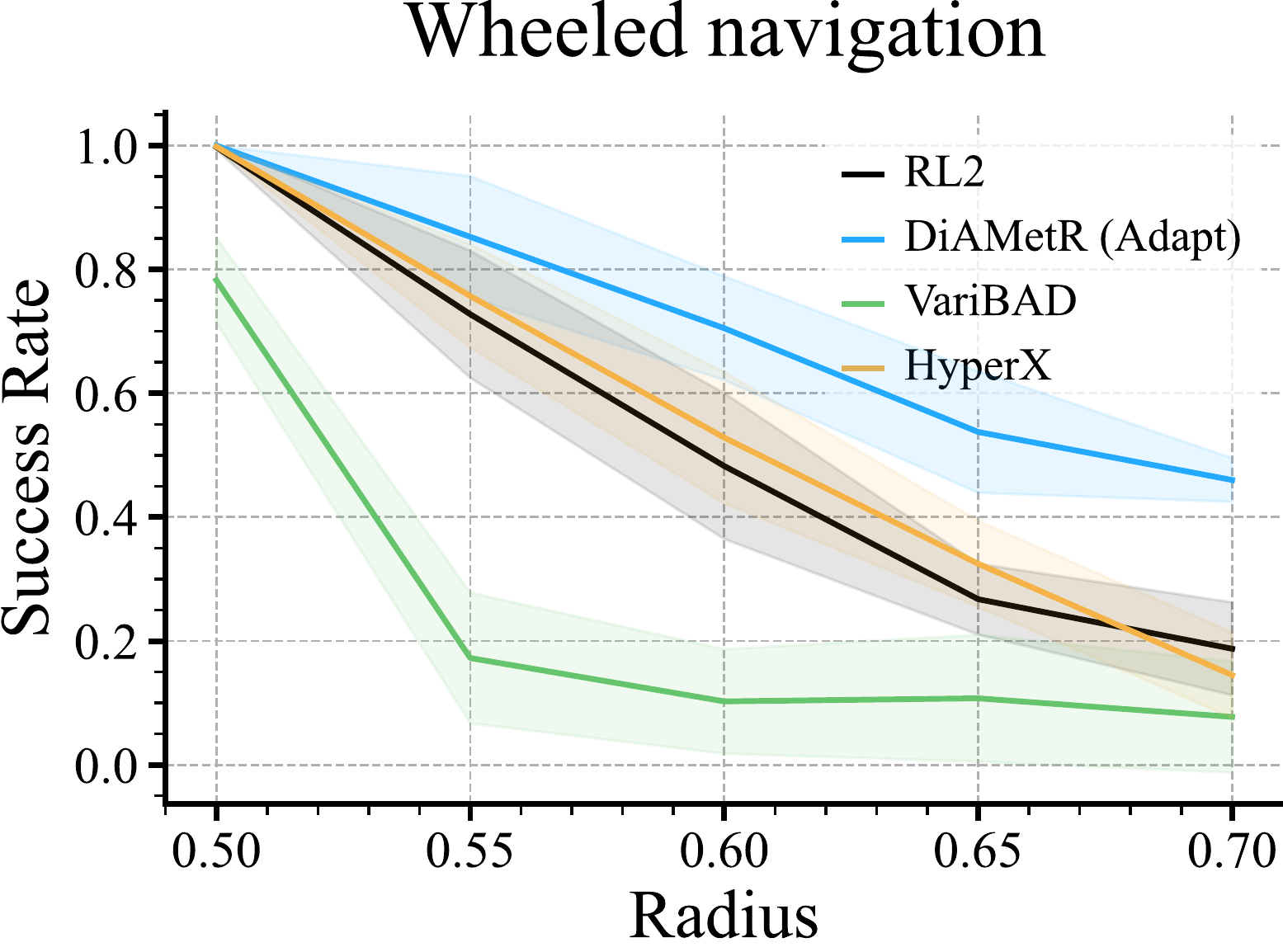}
    \includegraphics[width=0.24\linewidth]{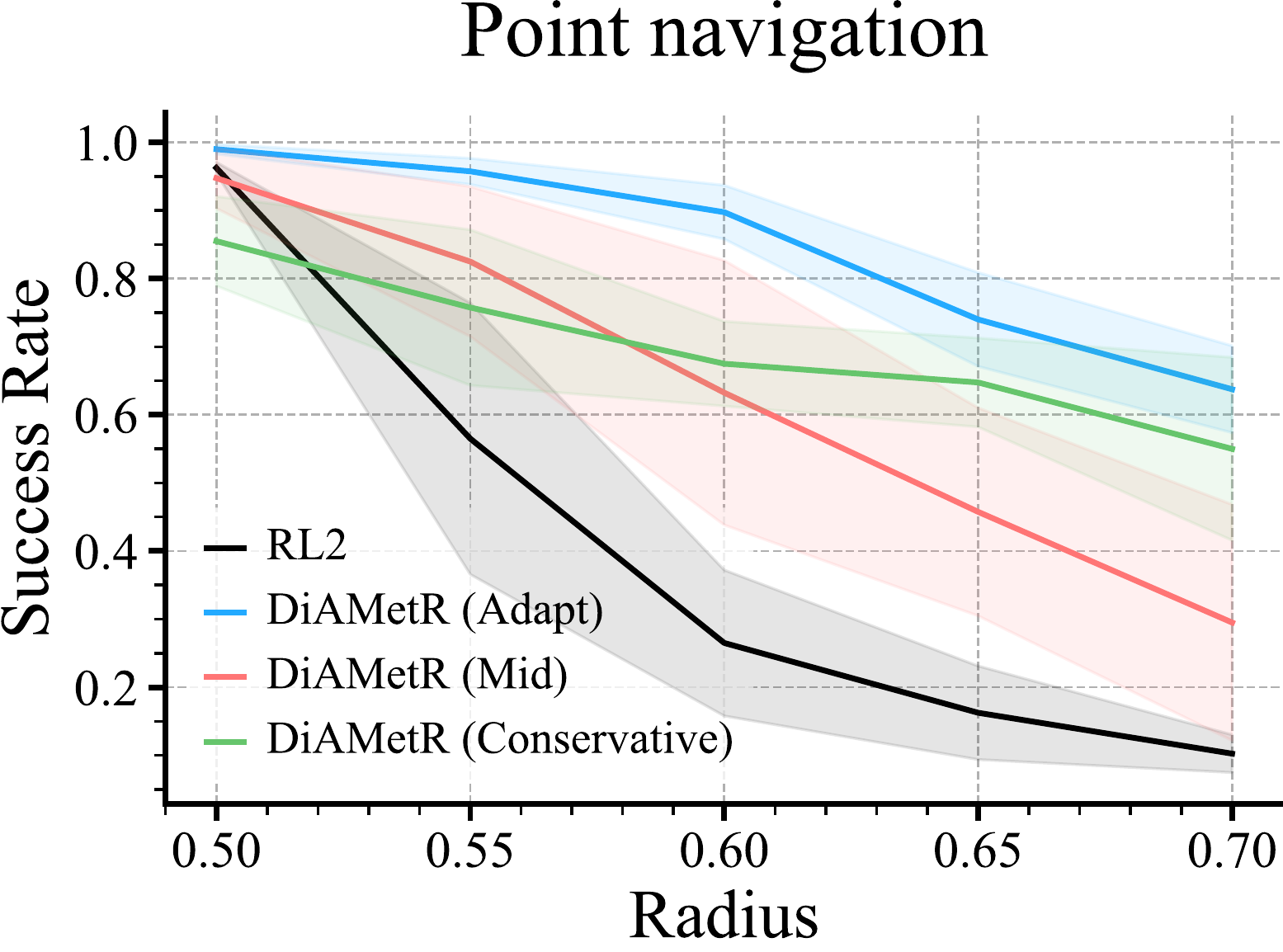}
    \includegraphics[width=0.24\linewidth]{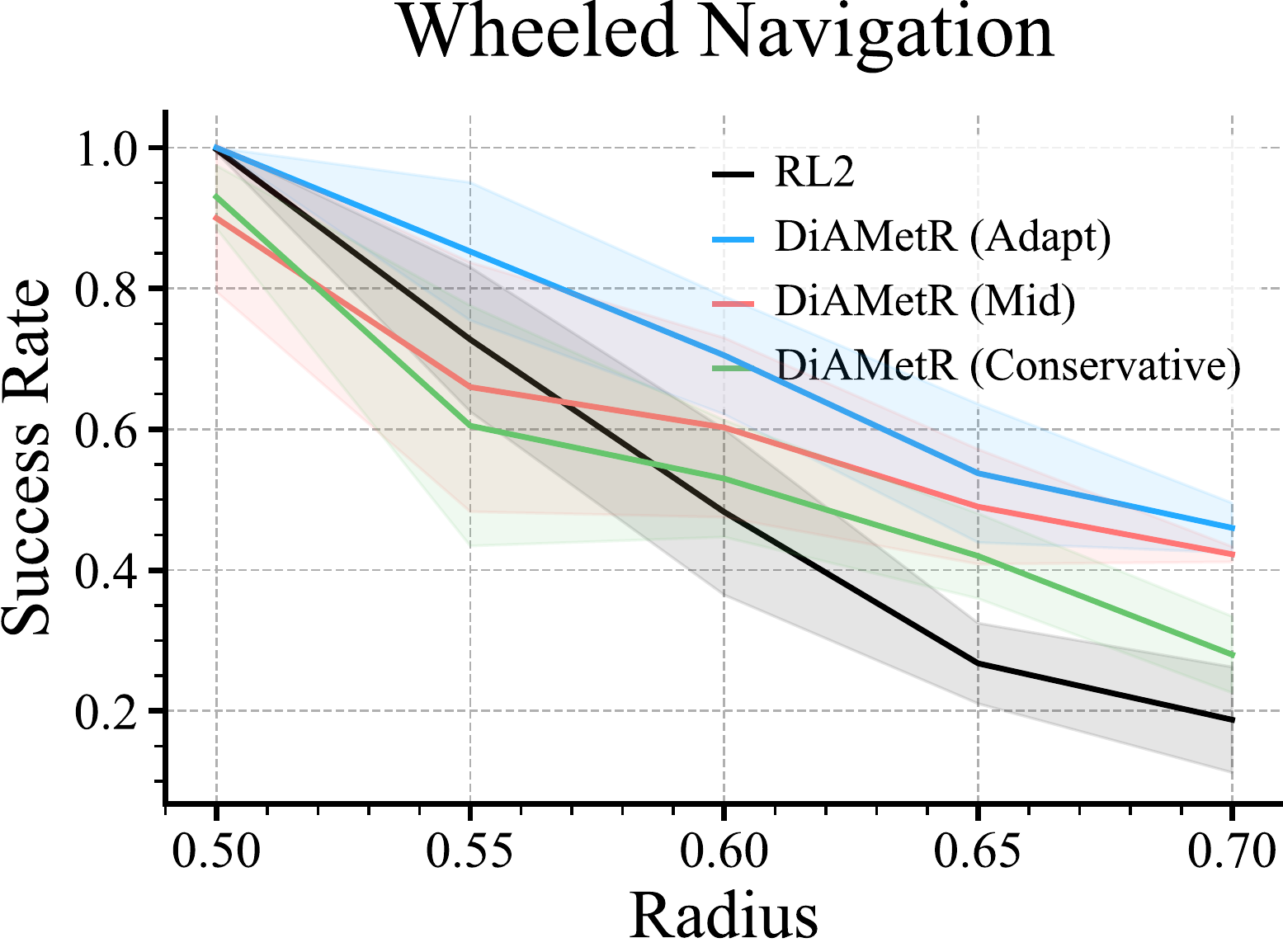}
    \caption{We evaluate \rml~and meta RL algorithms ($\text{RL}^2$, VariBAD and HyperX) on different \textit{in-support} and \textit{out-of-support} shifted test task distribution of \texttt{Point navigation} and \texttt{Wheeled navigation}. \rml~either matches or outperforms $\text{RL}^2$, VariBAD and HyperX on these task distributions. Furthermore, adaptively selecting an uncertainty set of \rml~policy (Adapt) during test time allows it to better adapt to test time distribution shift than choosing an uncertainty set beforehand (Mid). Choosing a large uncertainty set of \rml~policy (Conservative) leads to a conservative behavior and hurts its performance when test time distribution shift is low. The first point $p_1$ on the horizontal axis indicates the task parameter ($\Delta$) distribution $\mathcal{U}(0, p_1)$ and the subsequent points $p_i$ indicate task parameter ($\Delta$) distribution $\mathcal{U}(p_{i-1}, p_i)$. Here, task parameter is target location $s_\gT$. Table~\ref{tbl:detail_task_dist} details the task distributions used in this evaluation.
    } 
    \label{fig:point_exps}
\end{figure}
\section{Evaluations on dense reward environments with in-support distribution shifts}
\begin{figure}[h]
    \centering
    \includegraphics[width=0.24\linewidth]{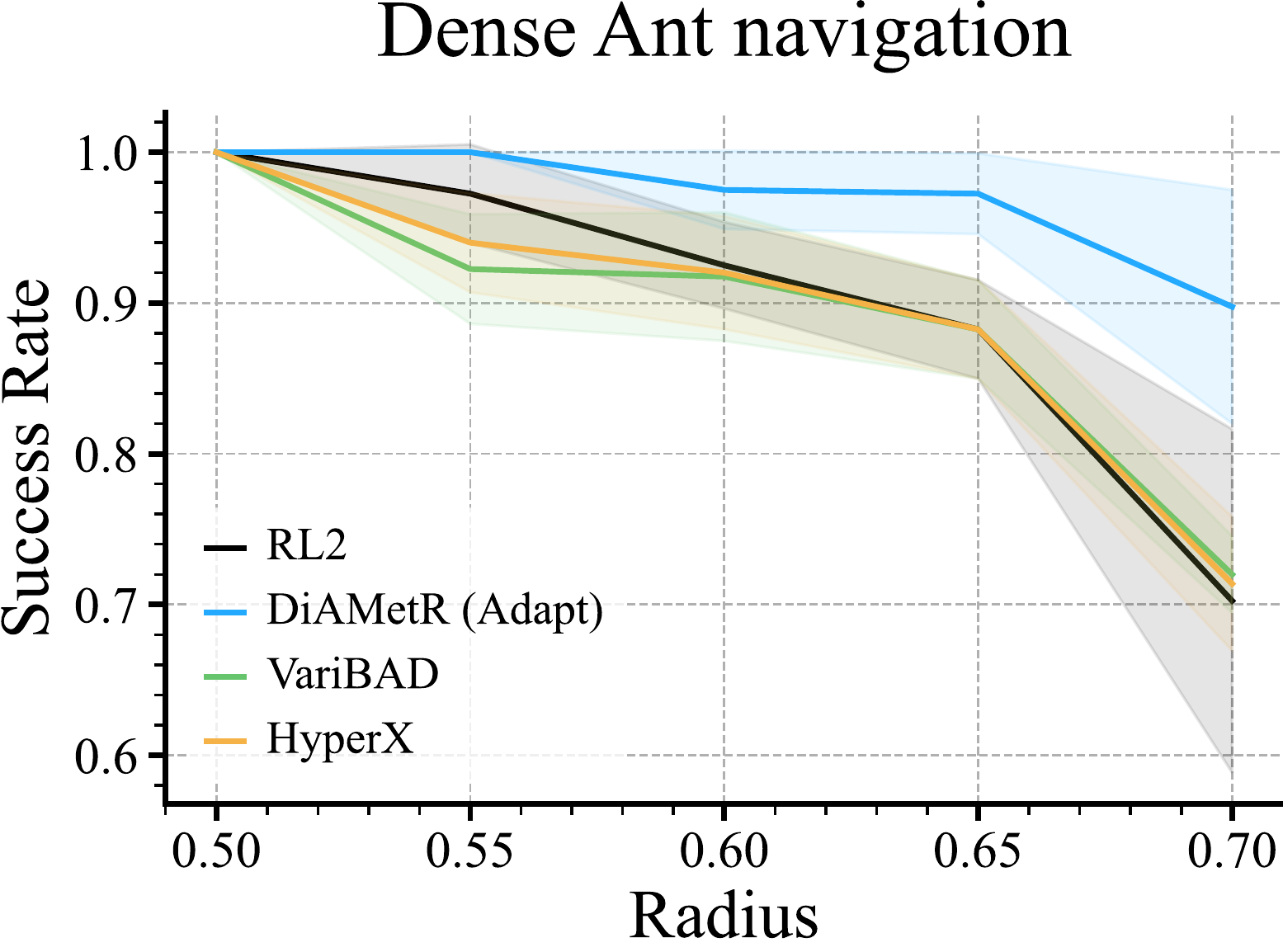}
    \includegraphics[width=0.24\linewidth]{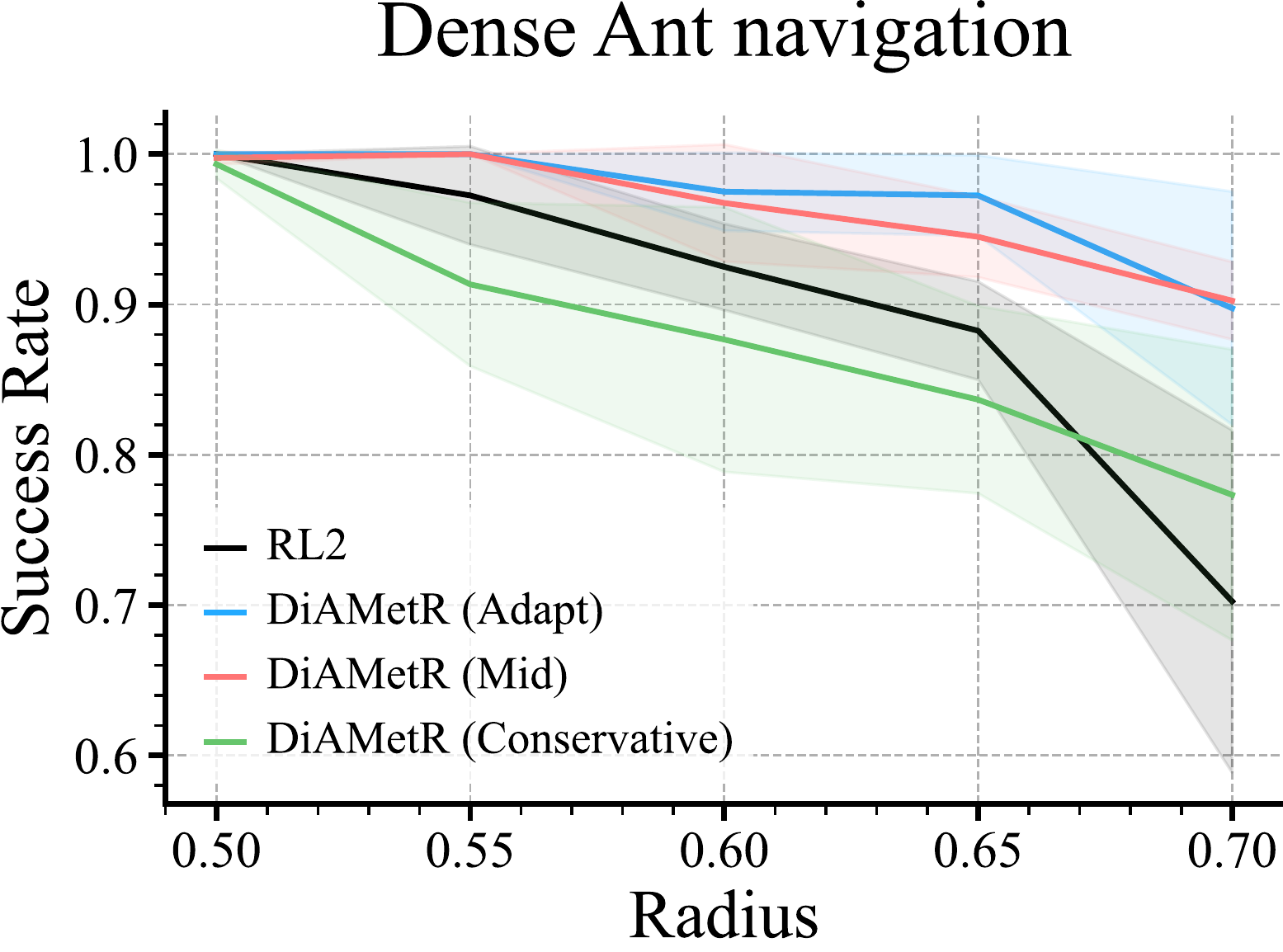}
    \caption{We evaluate \rml~and meta RL algorithms ($\text{RL}^2$, VariBAD and HyperX) on different \textit{in-support} shifted test task distributions of \texttt{Dense Ant-navigation}. \rml~either matches or outperforms $\text{RL}^2$, VariBAD and HyperX on these test task distributions. 
    Furthermore, selecting a single uncertainty set (that is neither too small nor too large) is sufficient in \texttt{Dense Ant-navigation} as \rml(Adapt) and \rml(Mid) have similar performances (within standard error). The first point $p_1$ on the horizontal axis indicates the task target distance ($\Delta$) distribution $\mathcal{U}(0, p_1)$ and the subsequent points $p_i$ indicate task target distance ($\Delta$) distribution $\mathcal{U}(p_{i-1}, p_i)$.}
    \label{fig:dense_rew_ant}
\end{figure}

We test the applicability of \rml~on an environment with dense rewards. We use a variant of \texttt{Ant navigation}, namely \texttt{Dense Ant-navigation} for this evaluation. Furthermore, the shifted test task distributions are \textit{in-support} of the training task distribution (see Table~\ref{tbl:detail_task_dist} for a detailed description of these task distributions). Figure~\ref{fig:dense_rew_ant} shows that \rml~still outperforms existing meta RL algorithms ($\text{RL}^2$, VariBAD, HyperX) on shifted test task distributions. However, the gap between \rml~and other meta RL algorithms is less than in sparse reward environments. Furthermore, Figure~\ref{fig:dense_rew_ant} shows that selecting a single uncertainty set (that is neither too small nor too large) is sufficient in \texttt{Dense Ant-navigation} as \rml(Adapt) and \rml(Mid) have similar performances (within standard error).
\section{Meta-policy Selection and Adaptation during Meta-test}
\label{sec:meta_test_adapt}
In this section, we show that \rml~is able adapt to various test task distributions across different environments by selecting an appropriate meta-policy based on the inferred test-time distribution shift and then quickly adapting the meta-policy to new tasks drawn from the same test-distribution. The performance of meta-RL baselines ($\text{RL}^2$, variBAD, HyperX) remains more or less the same after test-time finetuning showing that $10$ iteration (with $25$ rollouts per iteration) isn’t enough for the meta-RL baselines to adapt to a new task distribution. For comparison, these meta-RL baselines take 1500 iterations (with 25 meta-episodes per iteration) during training to learn a meta-policy for train task distribution.

\begin{figure}[H]
    \centering
    \begin{subfigure}[b]{0.24\textwidth}
        \includegraphics[width=\linewidth]{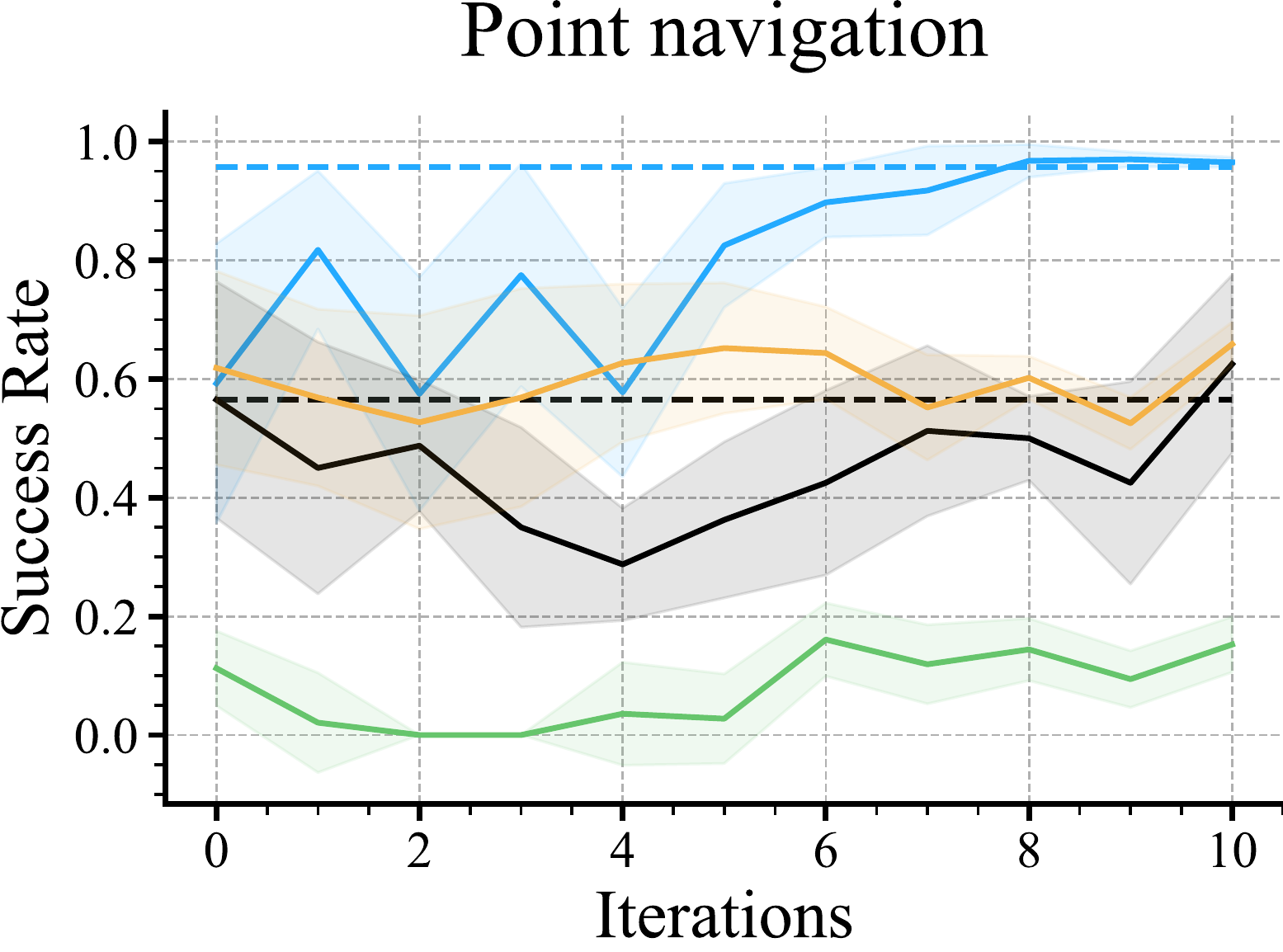}
        \caption{$\Delta \sim \mathcal{U}(0.5, 0.55)$}
    \end{subfigure}\hfill
    \begin{subfigure}[b]{0.24\textwidth}
        \includegraphics[width=\linewidth]{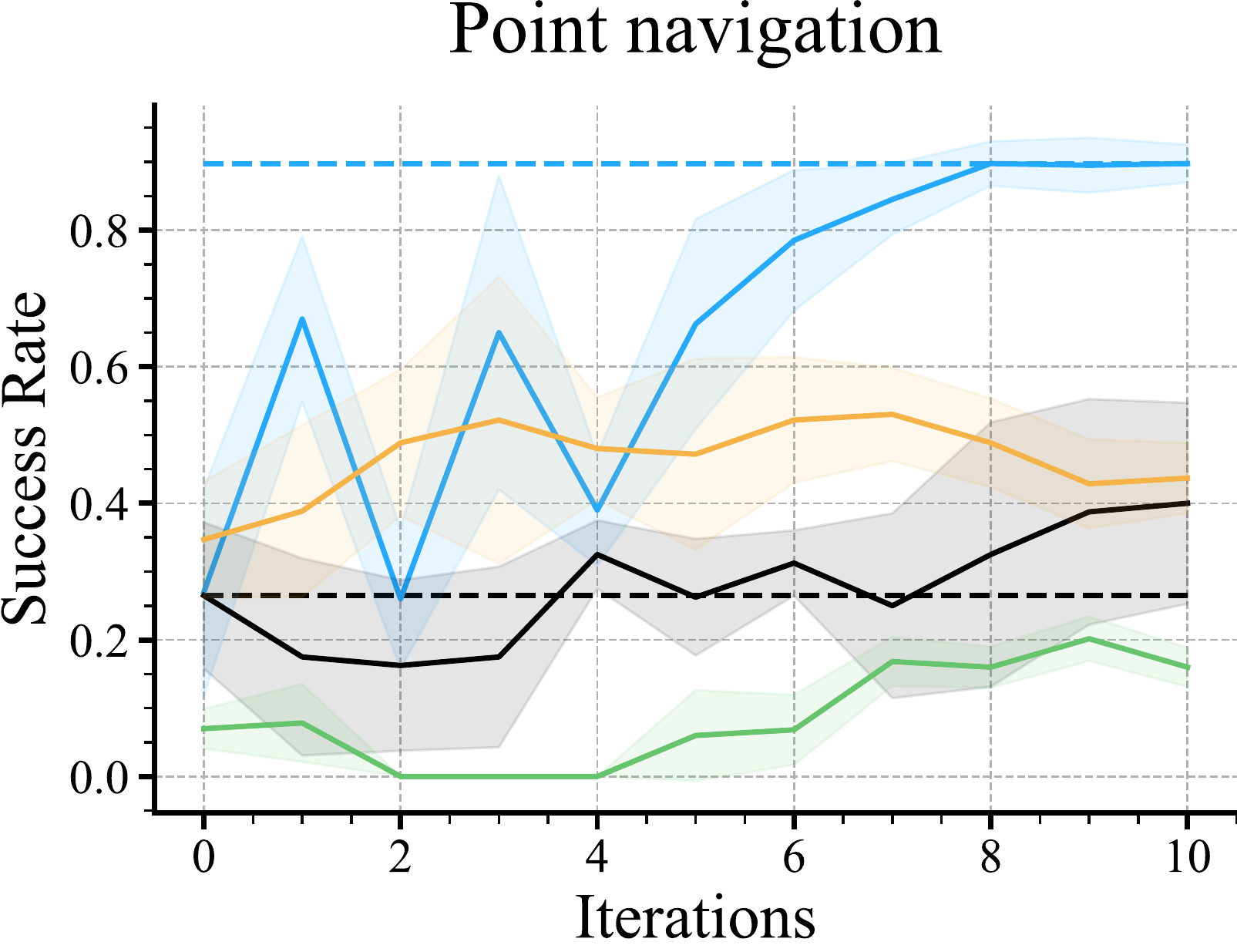}
        \caption{$\Delta \sim \mathcal{U}(0.55, 0.6)$}
    \end{subfigure}\hfill
    \begin{subfigure}[b]{0.24\textwidth}
        \includegraphics[width=\linewidth]{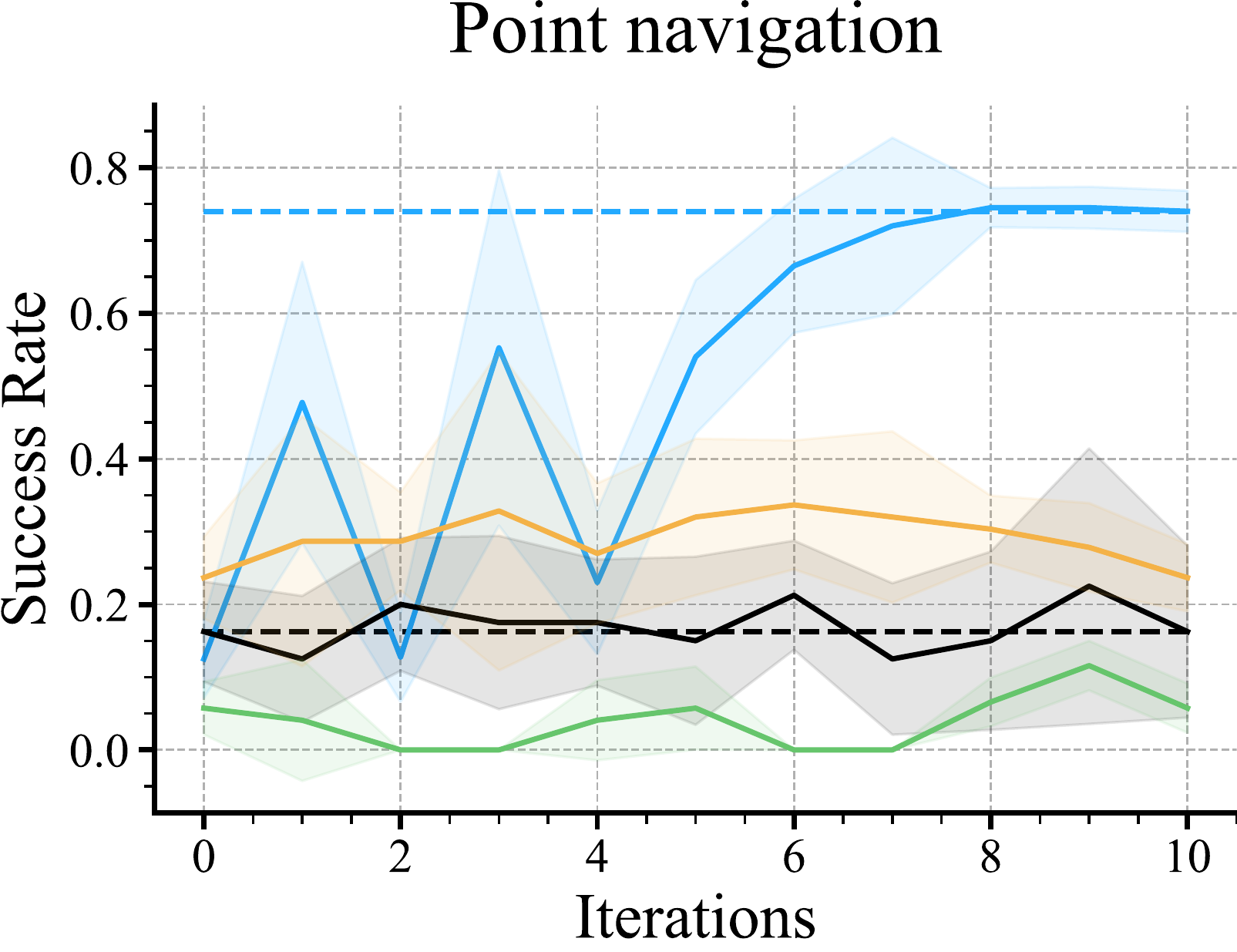}
        \caption{$\Delta \sim \mathcal{U}(0.6, 0.65)$}
    \end{subfigure}\hfill
    \begin{subfigure}[b]{0.24\textwidth}
        \includegraphics[width=\linewidth]{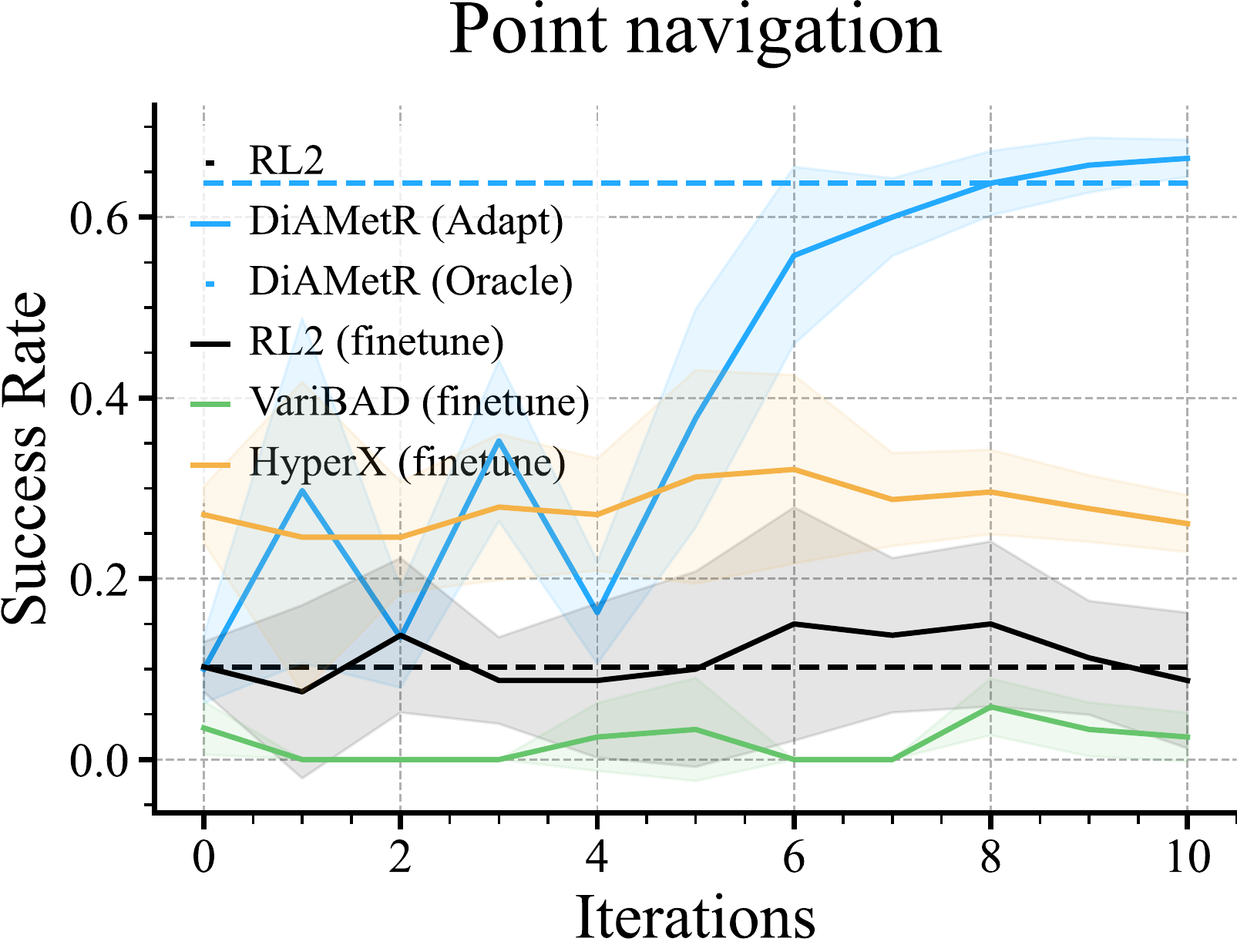}
        \caption{$\Delta \sim \mathcal{U}(0.65, 0.7)$}
    \end{subfigure}\hfill
    \caption{We compare test time adaptation of \rml~with test time finetuning of $\text{RL}^2$ on point robot navigation for various test task distributions. We run the adaptation procedure for $10$ iterations collecting $25$ rollouts per iteration.}
    \label{fig:point_robot_test_adapt}
\end{figure}

\begin{figure}[H]
    \centering
    \begin{subfigure}[b]{0.24\textwidth}
        \includegraphics[width=\linewidth]{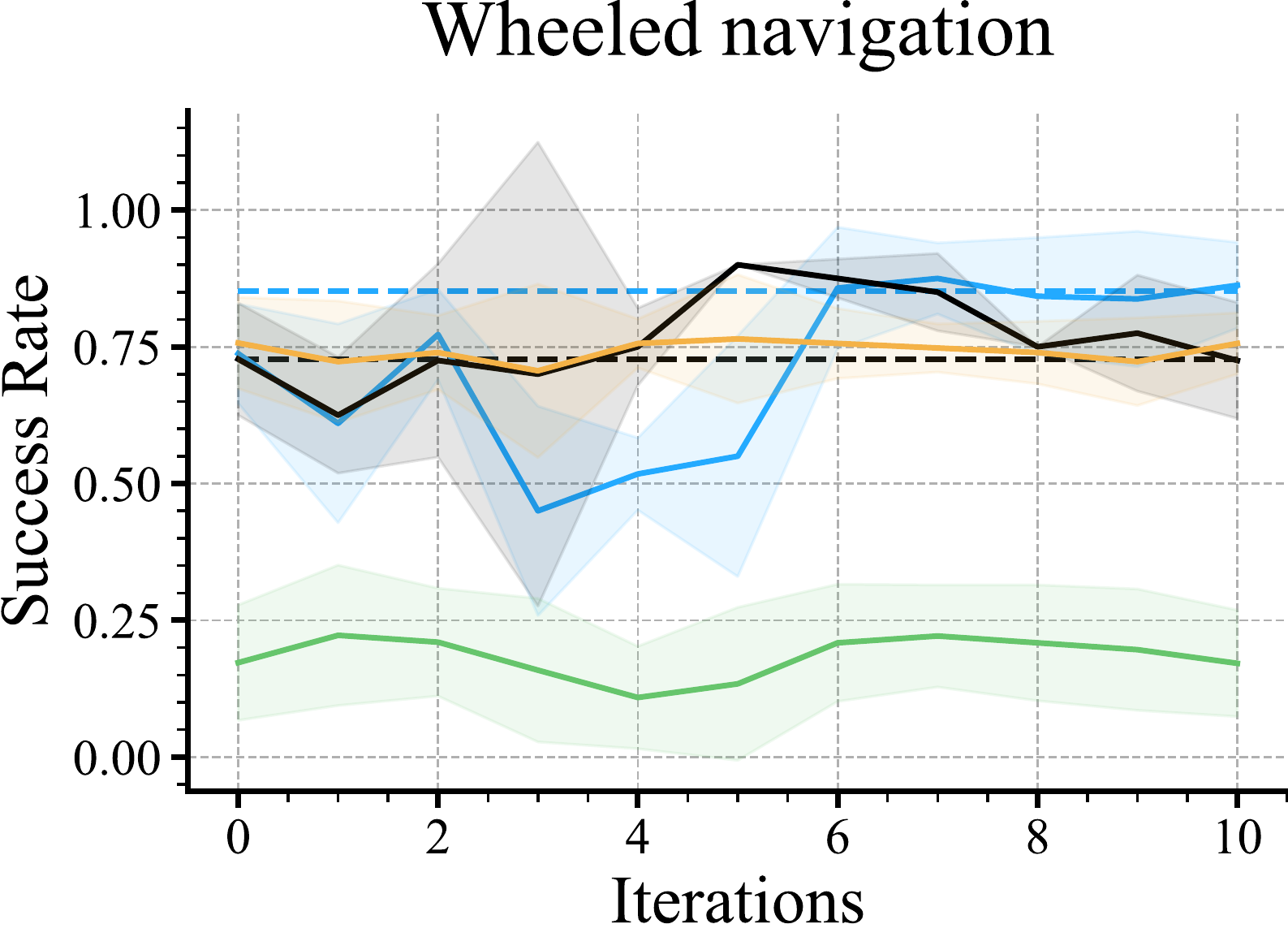}
        \caption{$\Delta \sim \mathcal{U}(0.5, 0.55)$}
    \end{subfigure}\hfill
    \begin{subfigure}[b]{0.24\textwidth}
        \includegraphics[width=\linewidth]{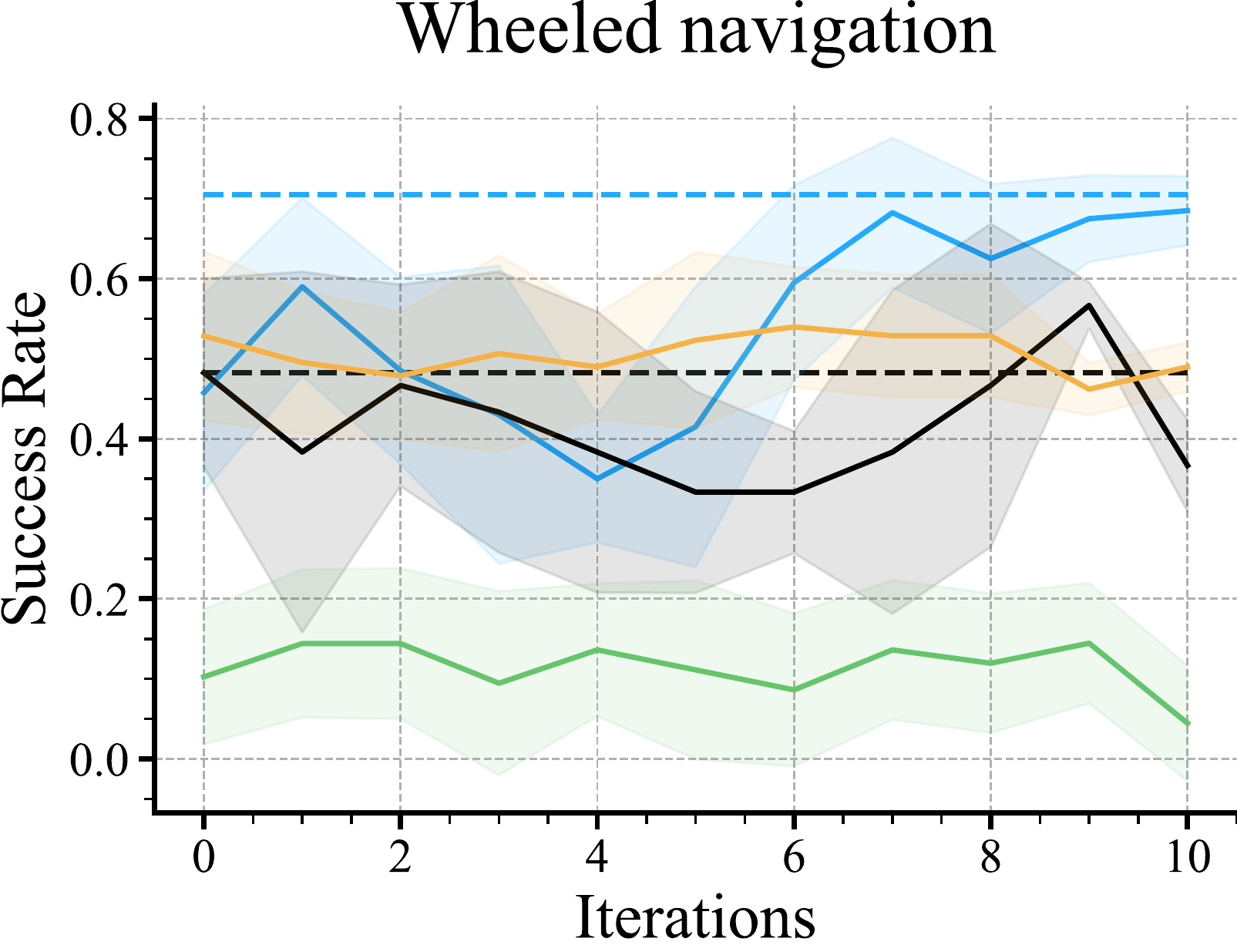}
        \caption{$\Delta \sim \mathcal{U}(0.55, 0.6)$}
    \end{subfigure}\hfill
    \begin{subfigure}[b]{0.24\textwidth}
        \includegraphics[width=\linewidth]{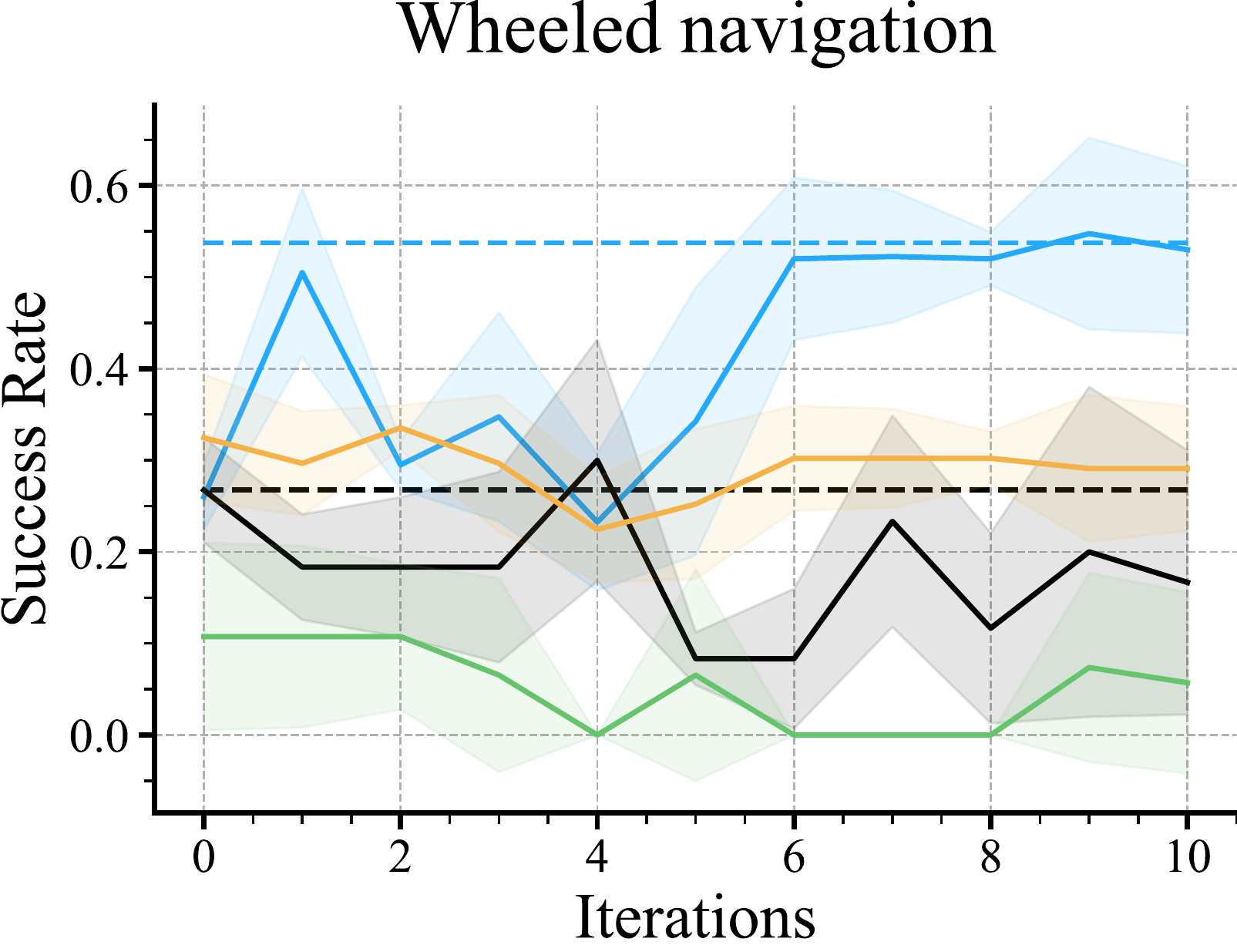}
        \caption{$\Delta \sim \mathcal{U}(0.6, 0.65)$}
    \end{subfigure}\hfill
    \begin{subfigure}[b]{0.24\textwidth}
        \includegraphics[width=\linewidth]{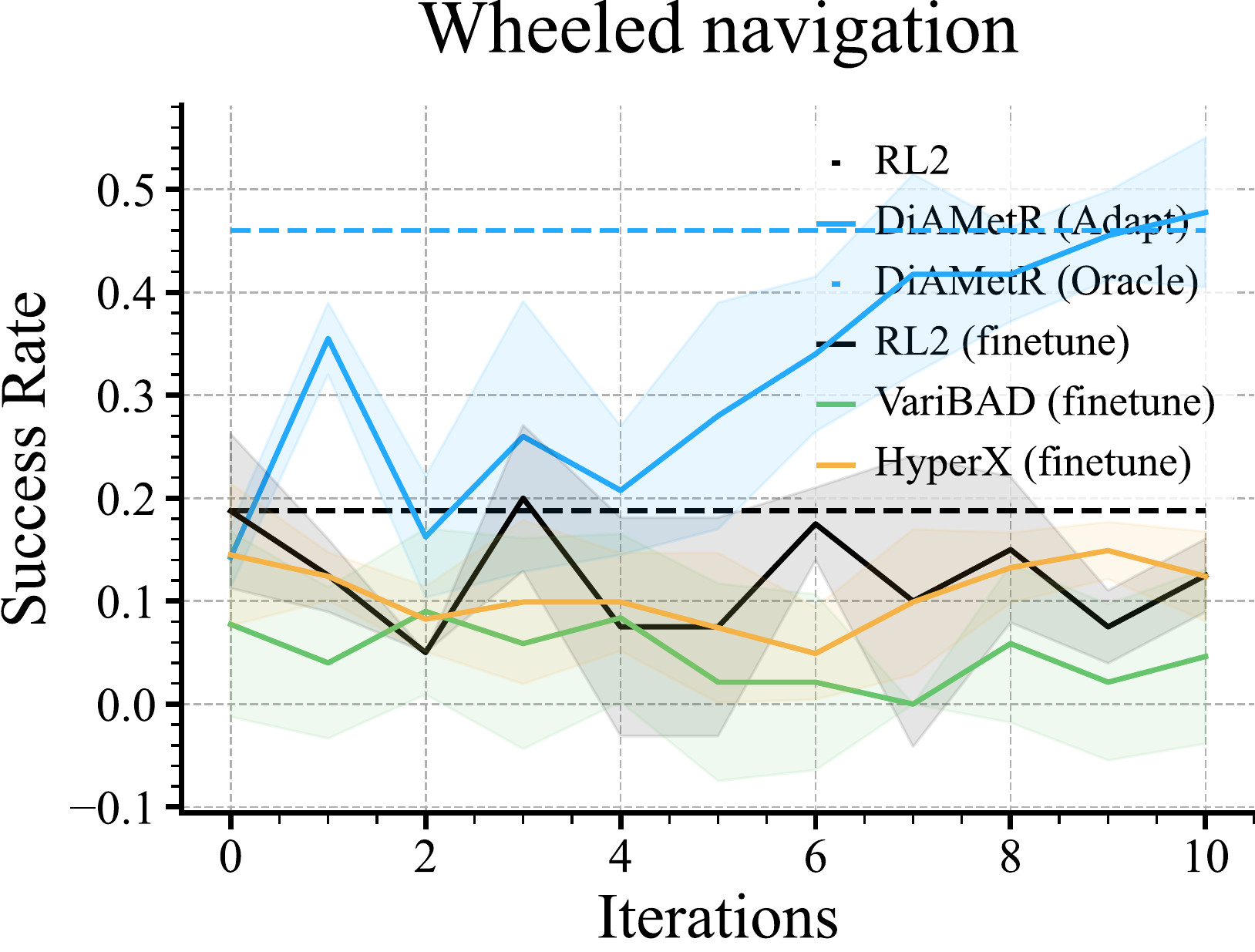}
        \caption{$\Delta \sim \mathcal{U}(0.65, 0.7)$}
    \end{subfigure}\hfill
    \caption{We compare test time adaptation of \rml~with test time finetuning of $\text{RL}^2$ on wheeled navigation for various test task distributions. We run the adaptation procedure for $10$ iterations collecting $25$ rollouts per iteration.}
    \label{fig:wheeled_test_adapt}
\end{figure}

\begin{figure}[H]
    \centering
    \begin{subfigure}[b]{0.24\textwidth}
        \includegraphics[width=\linewidth]{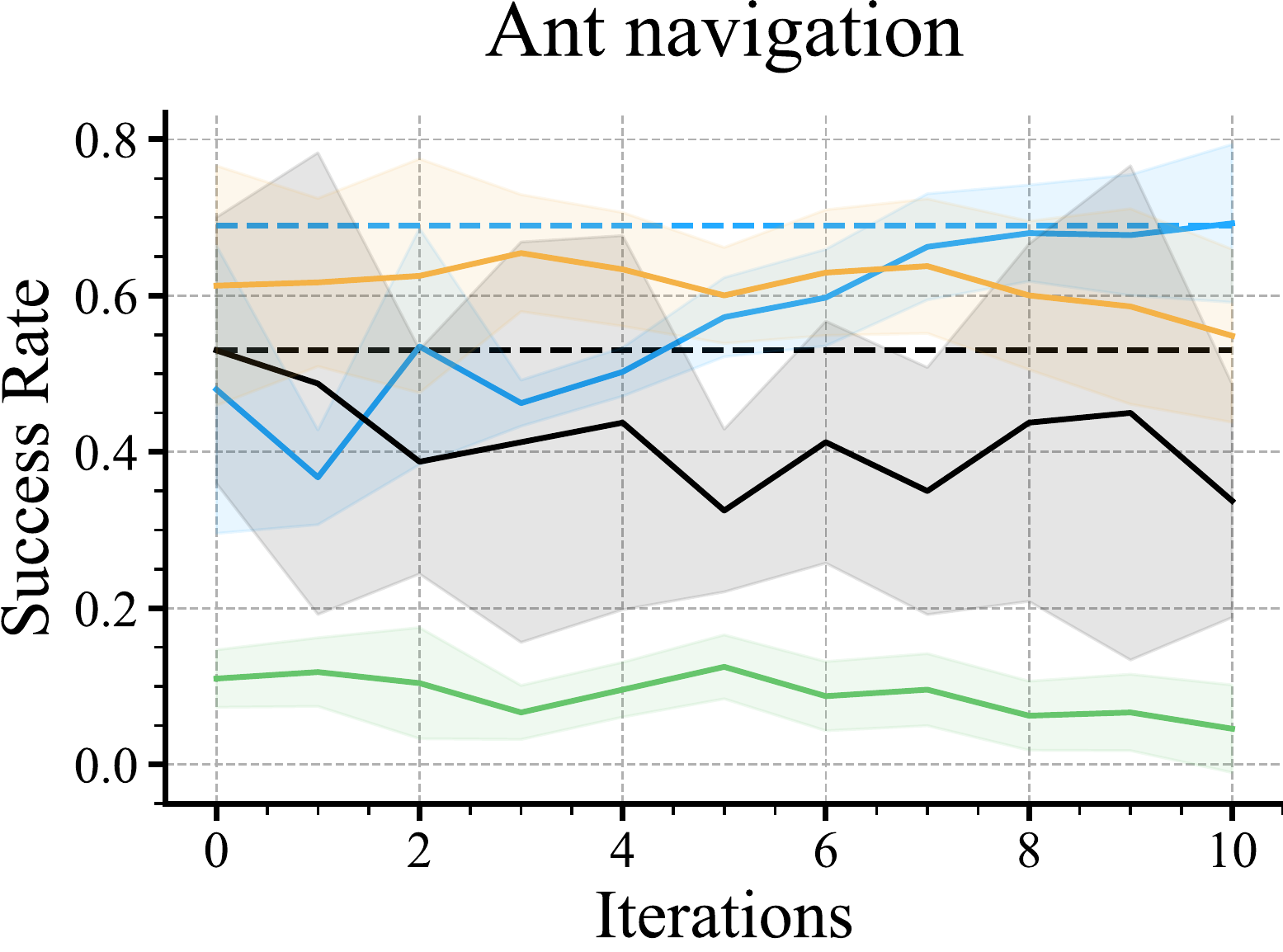}
        \caption{$\Delta \sim \mathcal{U}(0.5, 0.55)$}
    \end{subfigure}\hfill
    \begin{subfigure}[b]{0.24\textwidth}
        \includegraphics[width=\linewidth]{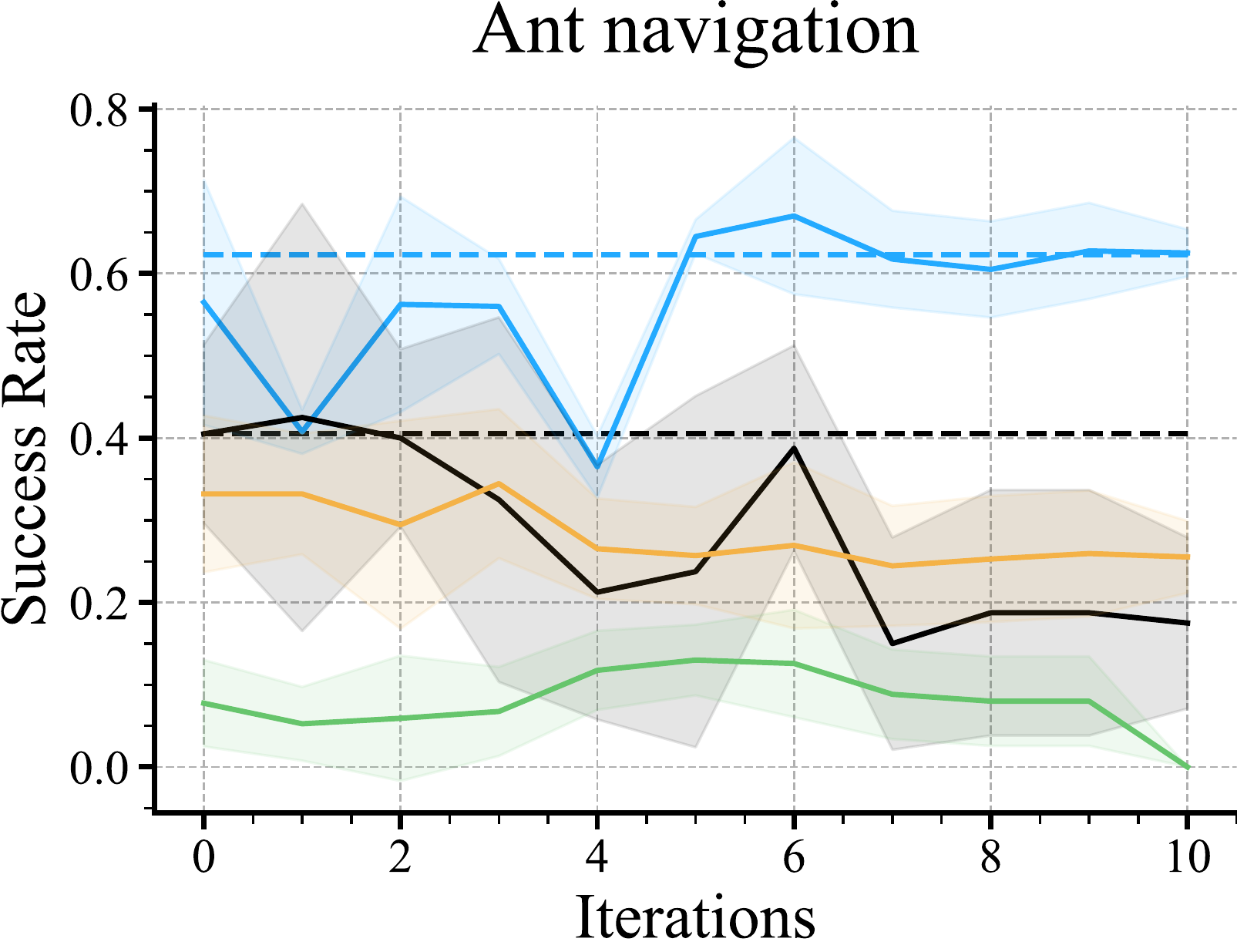}
        \caption{$\Delta \sim \mathcal{U}(0.55, 0.6)$}
    \end{subfigure}\hfill
    \begin{subfigure}[b]{0.24\textwidth}
        \includegraphics[width=\linewidth]{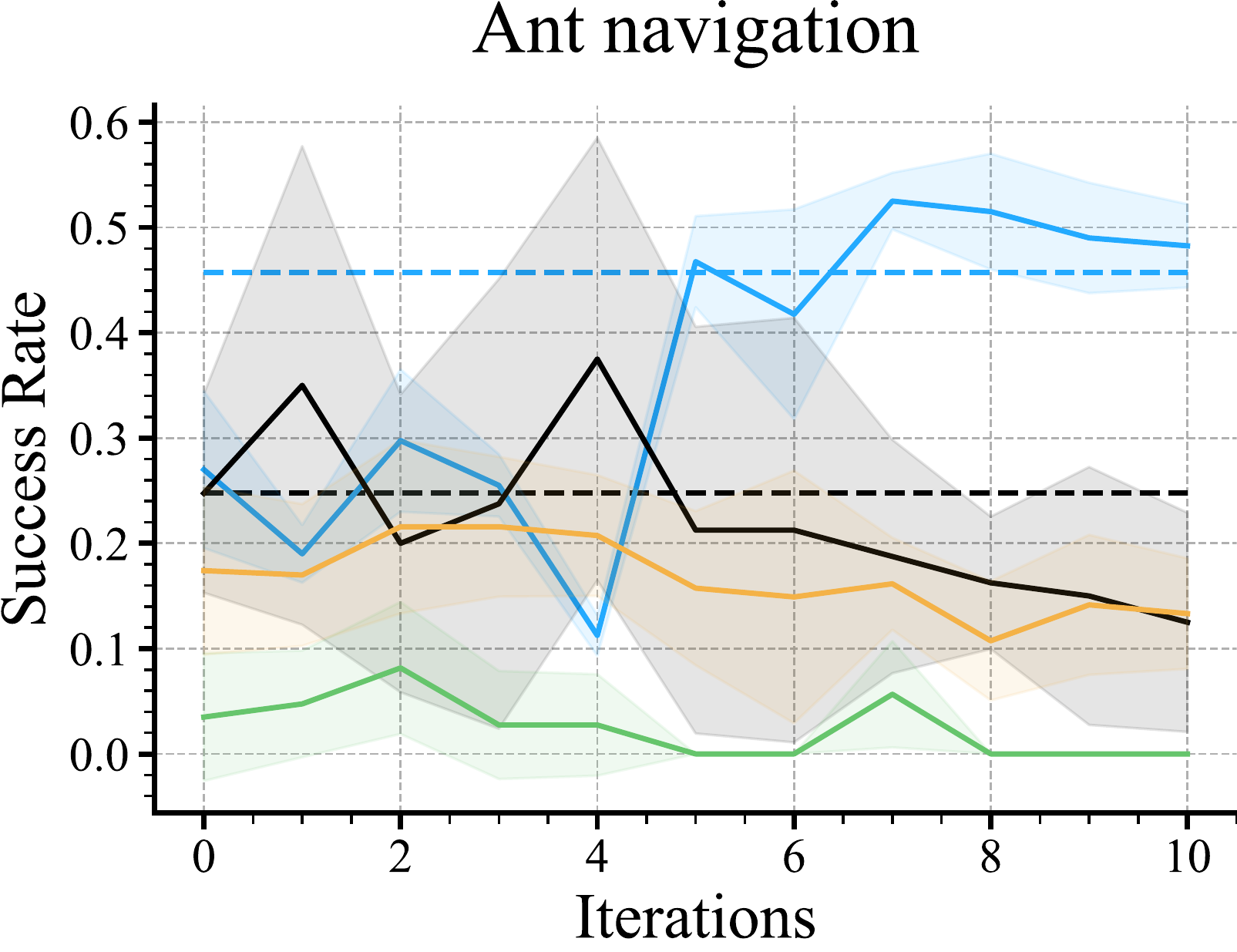}
        \caption{$\Delta \sim \mathcal{U}(0.6, 0.65)$}
    \end{subfigure}\hfill
    \begin{subfigure}[b]{0.24\textwidth}
        \includegraphics[width=\linewidth]{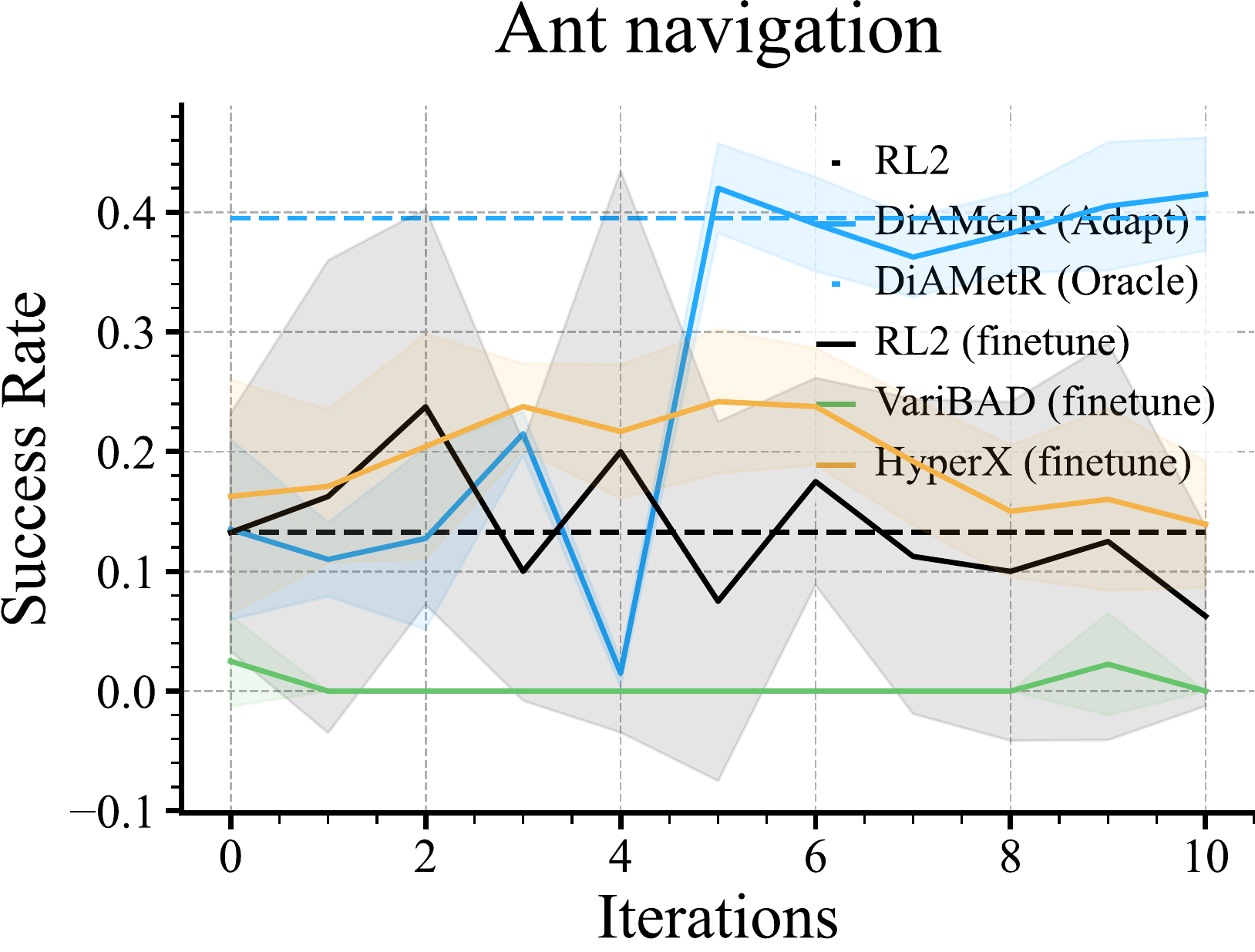}
        \caption{$\Delta \sim \mathcal{U}(0.65, 0.7)$}
    \end{subfigure}\hfill
    \caption{We compare test time adaptation of \rml~with test time finetuning of $\text{RL}^2$ on ant navigation for various test task distributions. We run the adaptation procedure for $10$ iterations collecting $25$ rollouts per iteration.}
    \label{fig:ant_goal_test_adapt}
\end{figure}

\begin{figure}[H]
    \centering
    \begin{subfigure}[b]{0.24\textwidth}
        \includegraphics[width=\linewidth]{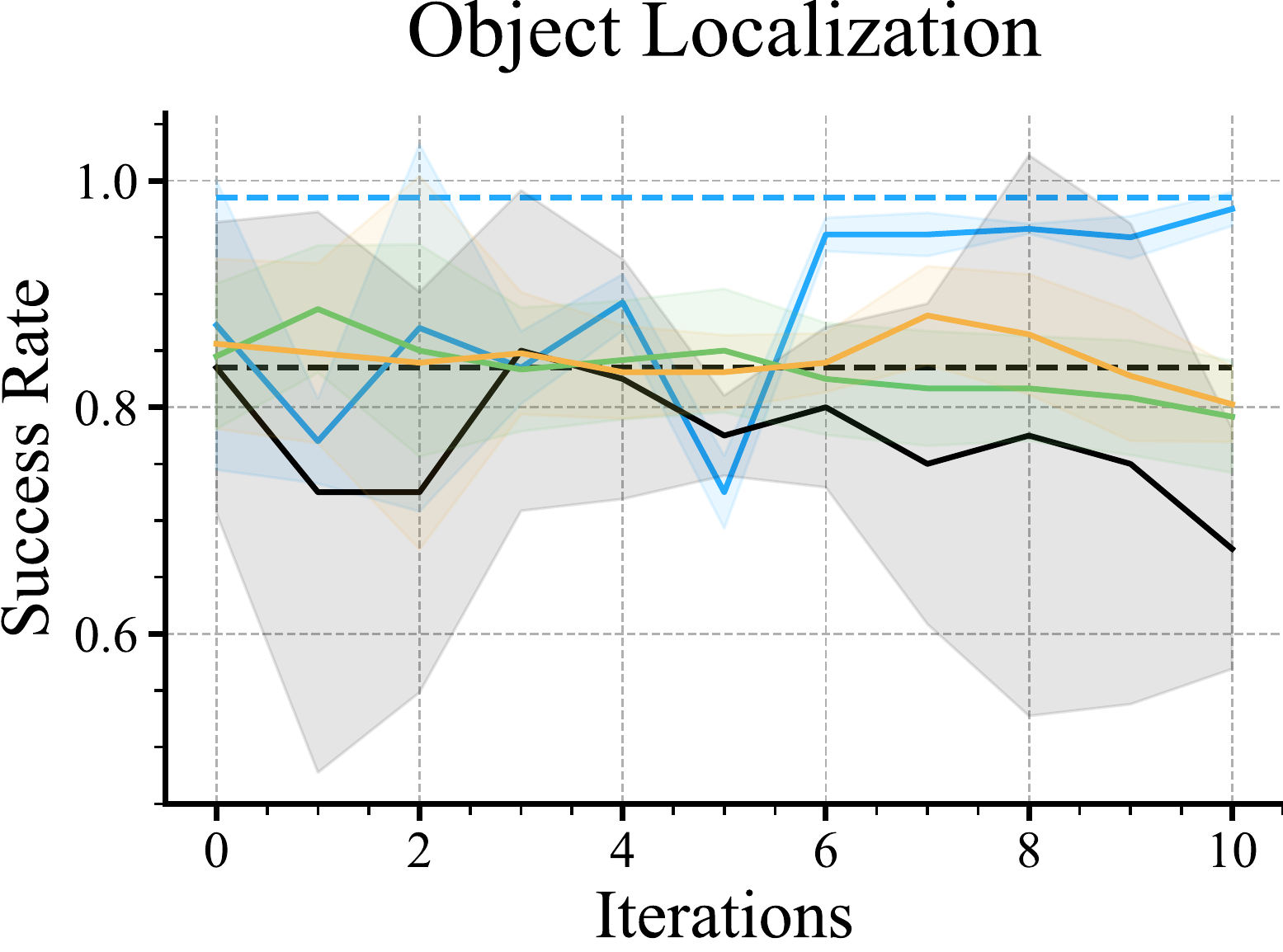}
        \caption{$\Delta \sim \mathcal{U}(0.1, 0.12)$}
    \end{subfigure}
    \begin{subfigure}[b]{0.24\textwidth}
        \includegraphics[width=\linewidth]{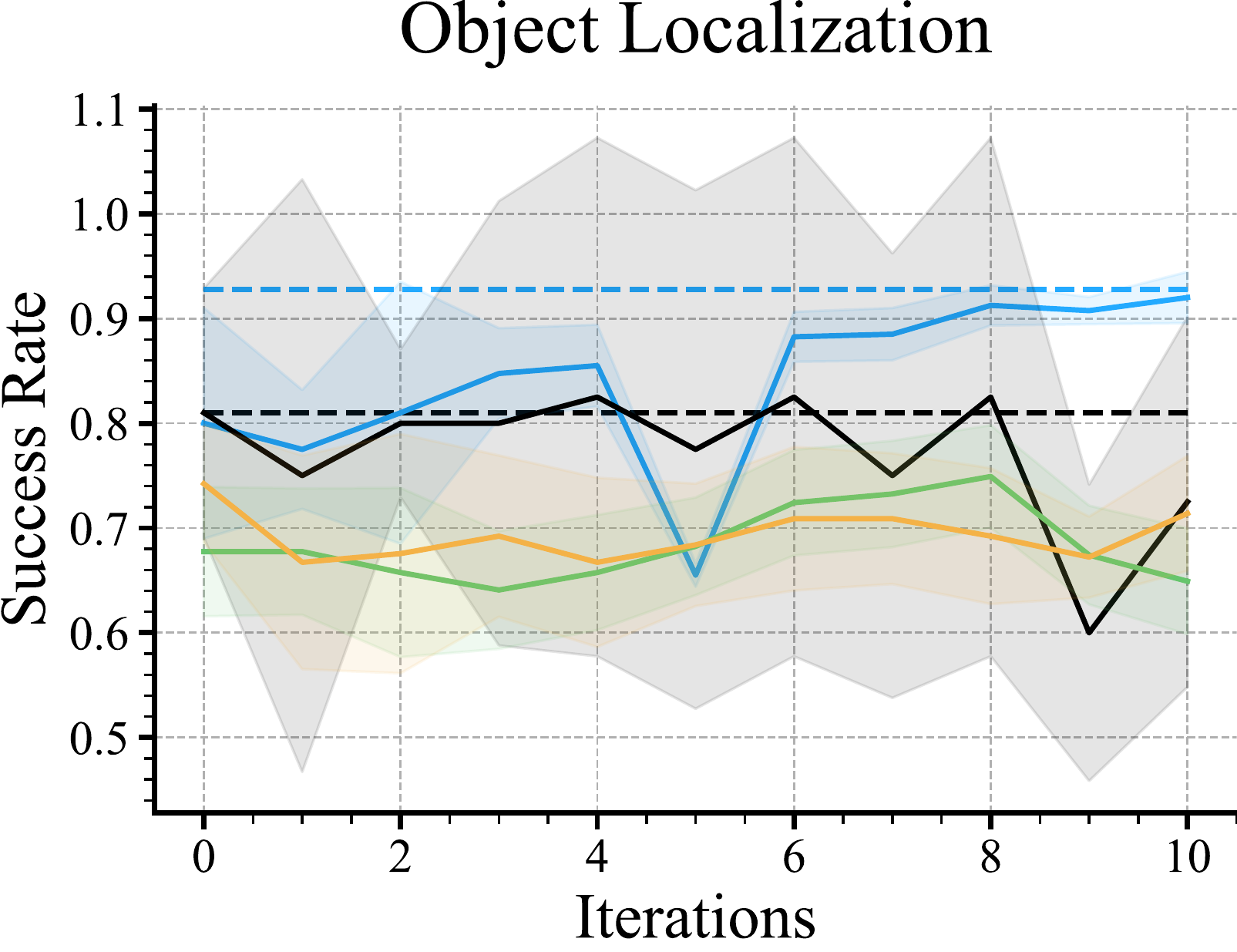}
        \caption{$\Delta \sim \mathcal{U}(0.12, 0.14)$}
    \end{subfigure}
    \begin{subfigure}[b]{0.24\textwidth}
        \includegraphics[width=\linewidth]{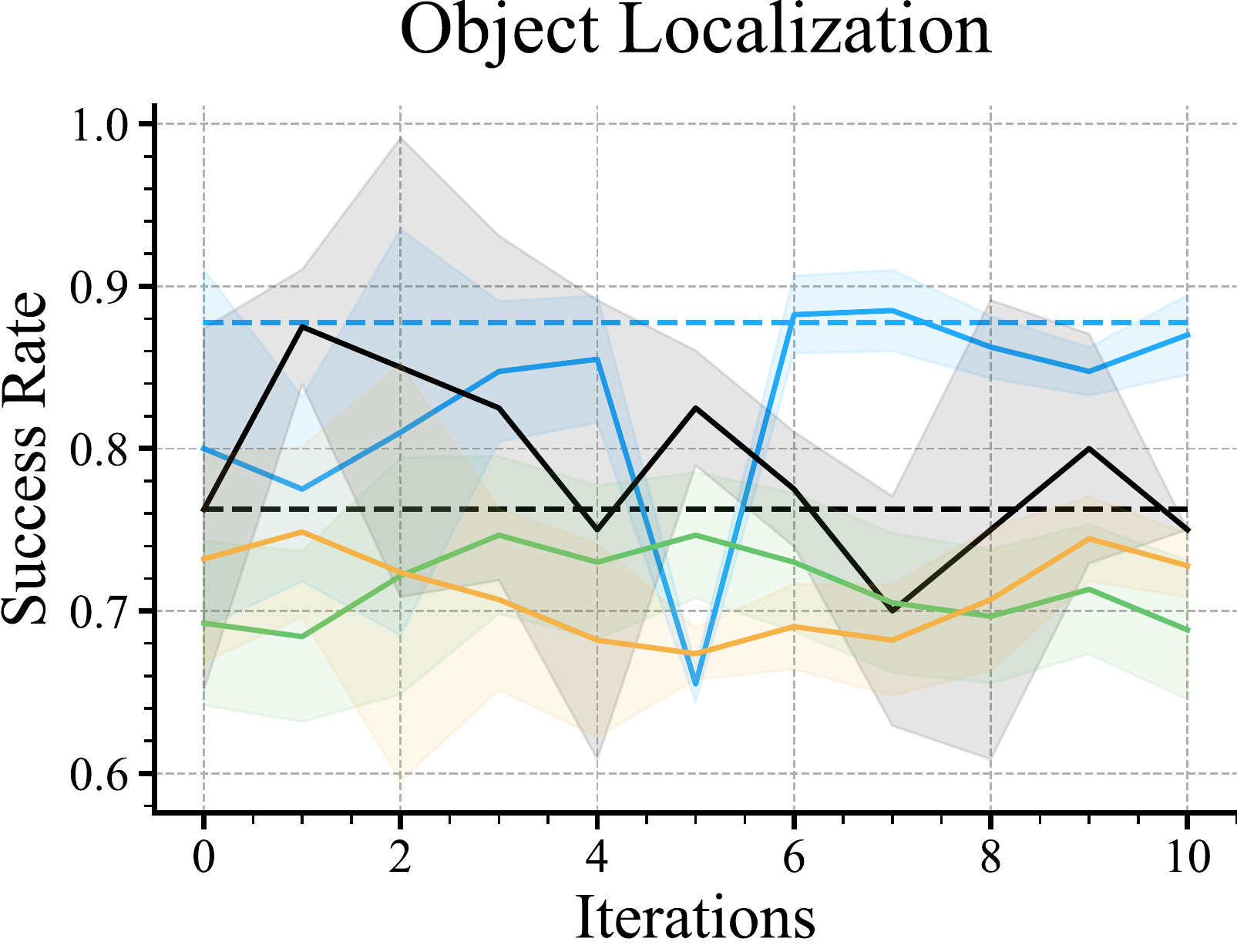}
        \caption{$\Delta \sim \mathcal{U}(0.14, 0.16)$}
    \end{subfigure}\\
    \begin{subfigure}[b]{0.24\textwidth}
        \includegraphics[width=\linewidth]{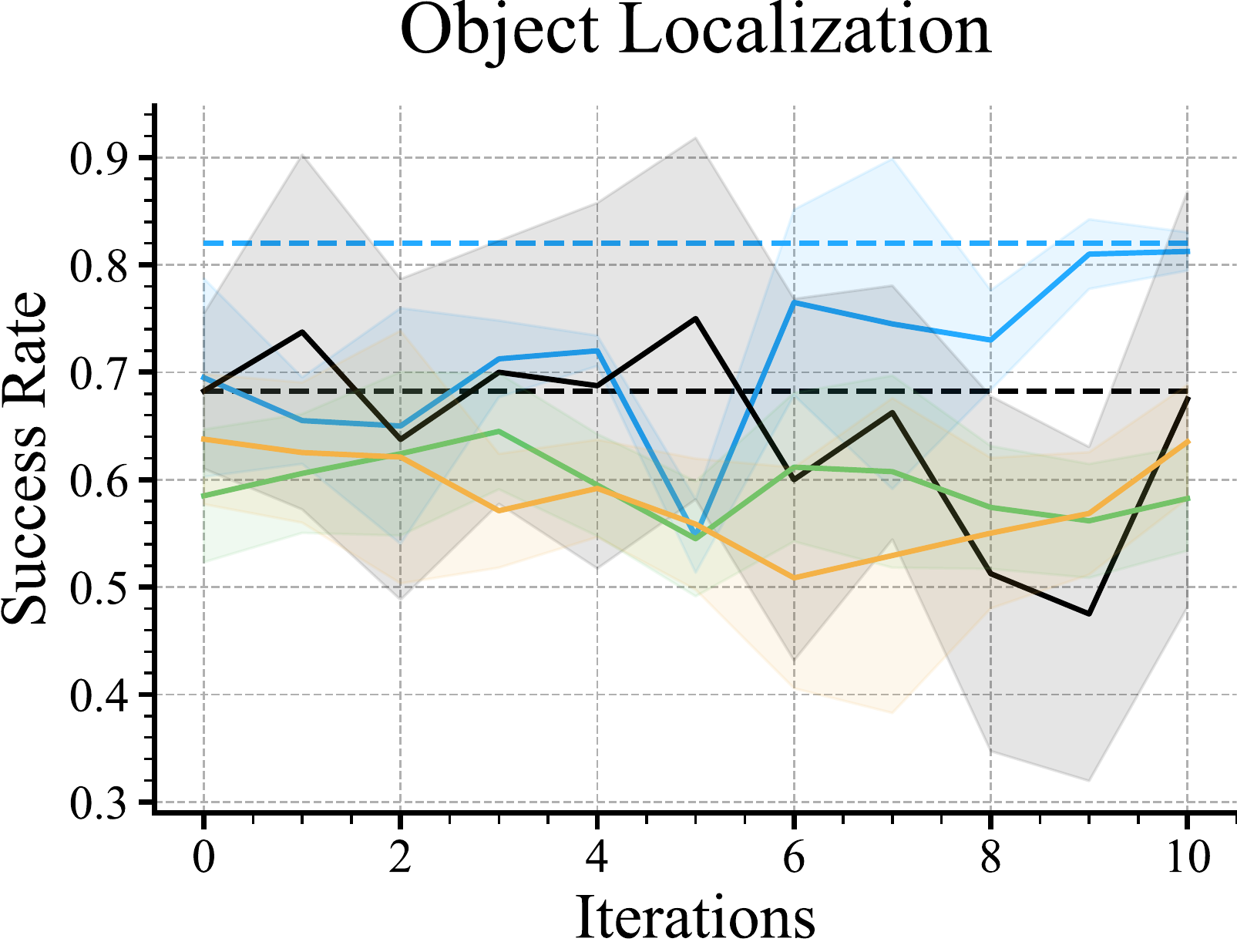}
        \caption{$\Delta \sim \mathcal{U}(0.16, 0.18)$}
    \end{subfigure}
    \begin{subfigure}[b]{0.24\textwidth}
        \includegraphics[width=\linewidth]{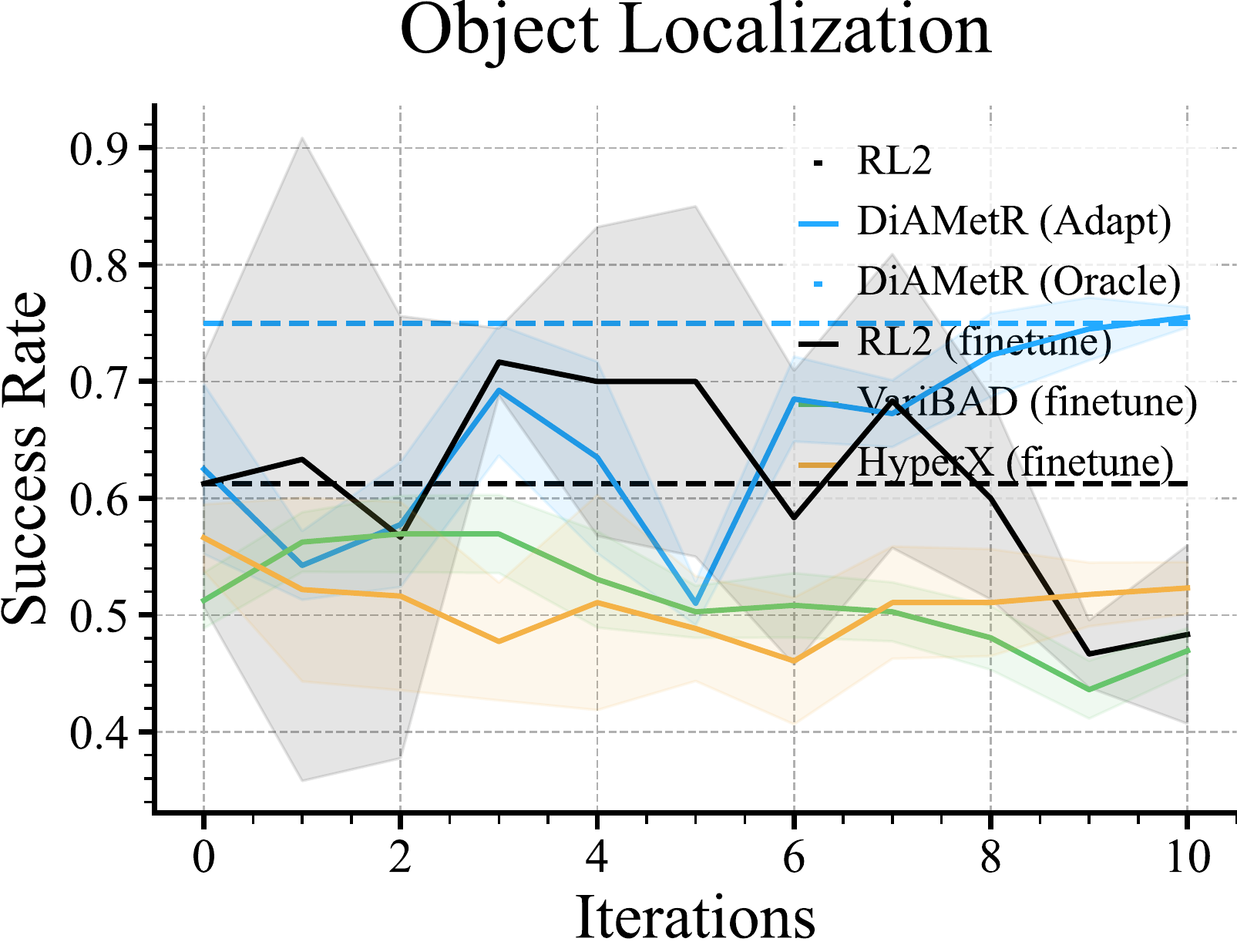}
        \caption{$\Delta \sim \mathcal{U}(0.18, 0.20)$}
    \end{subfigure}
    \caption{We compare test time adaptation of \rml~with test time finetuning of $\text{RL}^2$ on object localization for various test task distributions. We run the adaptation procedure for $10$ iterations collecting $25$ rollouts per iteration.}
    \label{fig:ant_goal_test_adapt}
\end{figure}

\begin{figure}[H]
    \centering
    \begin{subfigure}[b]{0.24\textwidth}
        \includegraphics[width=\linewidth]{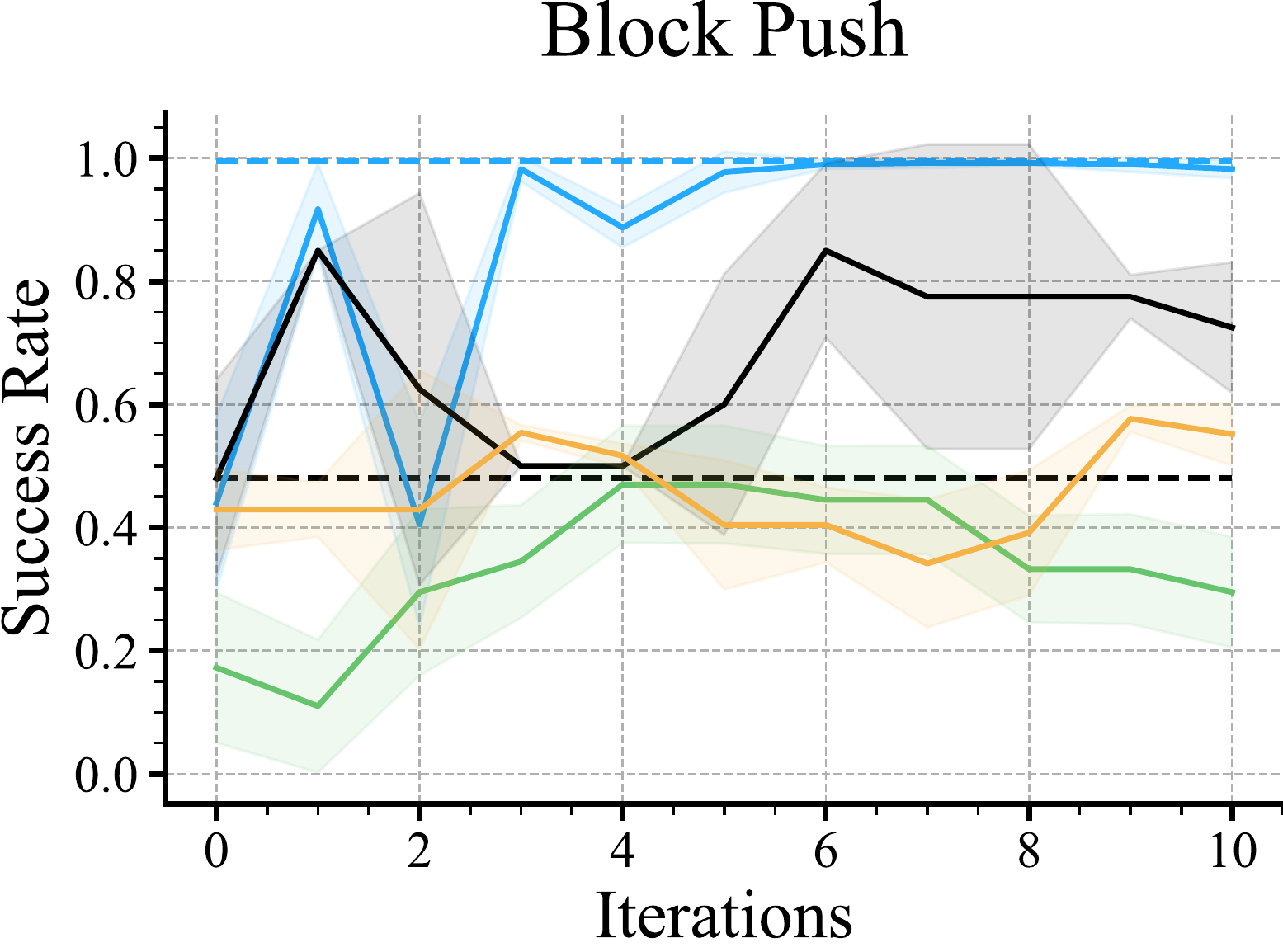}
        \caption{$\Delta \sim \mathcal{U}(0.5, 0.6)$}
    \end{subfigure}
    \begin{subfigure}[b]{0.24\textwidth}
        \includegraphics[width=\linewidth]{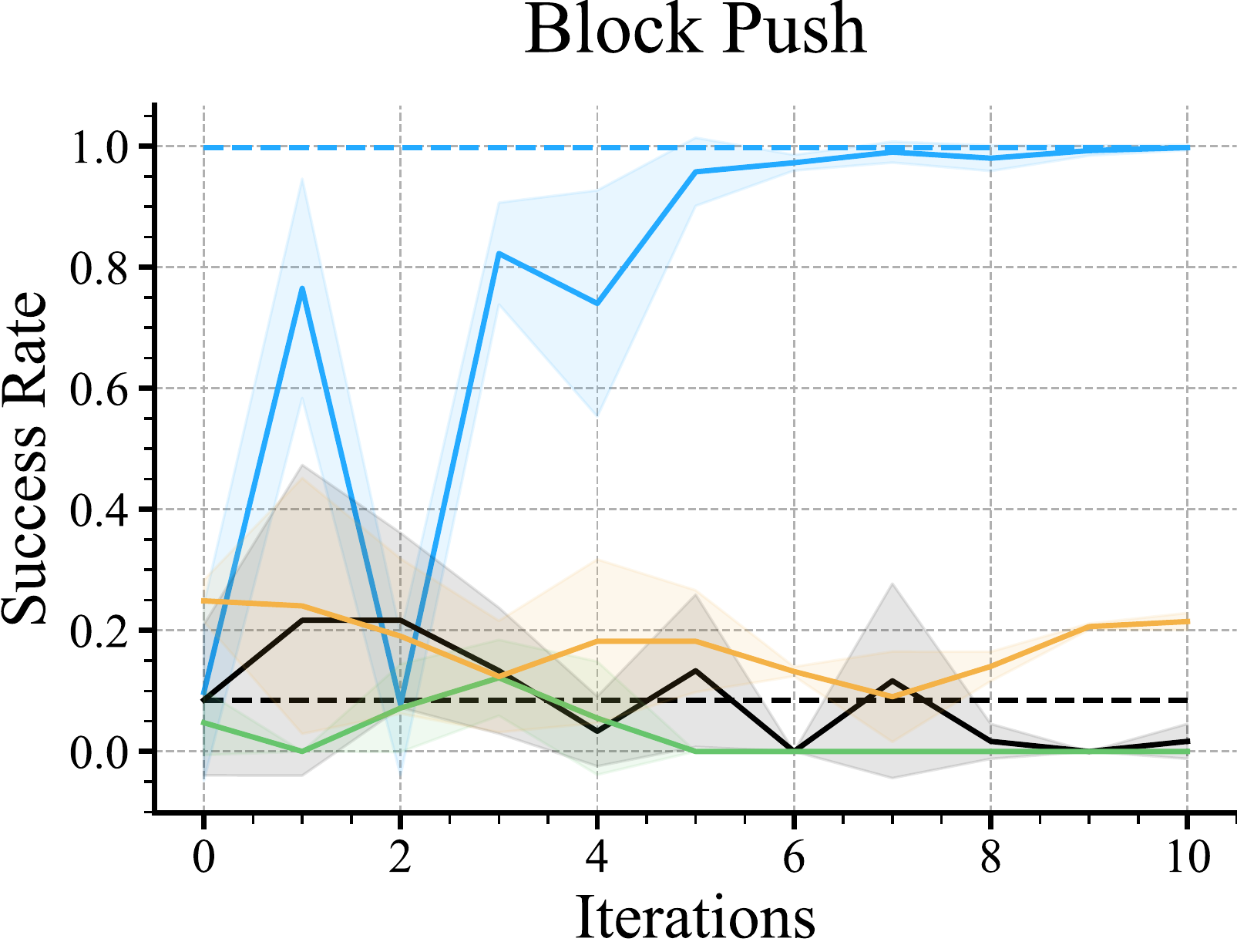}
        \caption{$\Delta \sim \mathcal{U}(0.6, 0.7)$}
    \end{subfigure}
    \begin{subfigure}[b]{0.24\textwidth}
        \includegraphics[width=\linewidth]{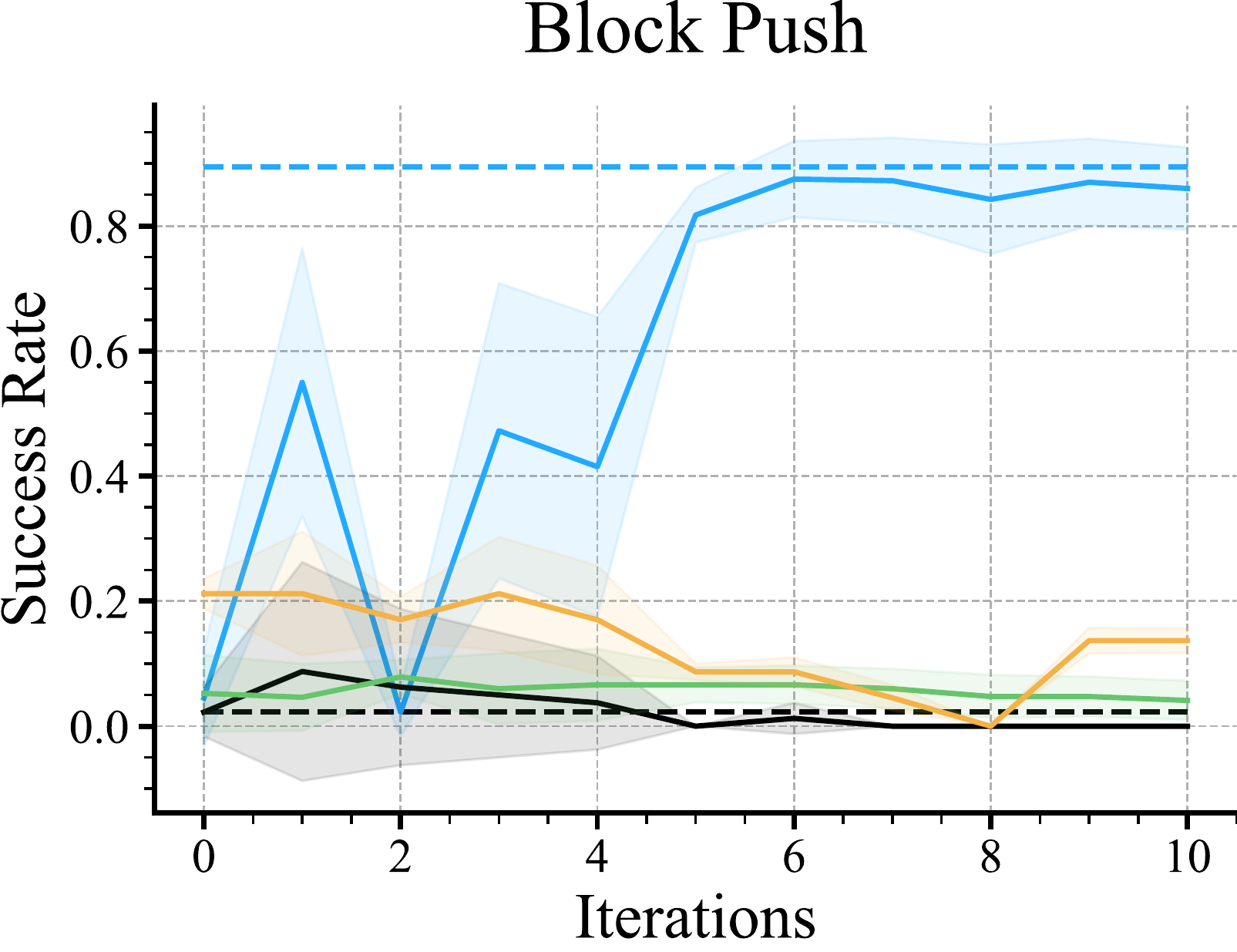}
        \caption{$\Delta \sim \mathcal{U}(0.7, 0.8)$}
    \end{subfigure}\\
    \begin{subfigure}[b]{0.24\textwidth}
        \includegraphics[width=\linewidth]{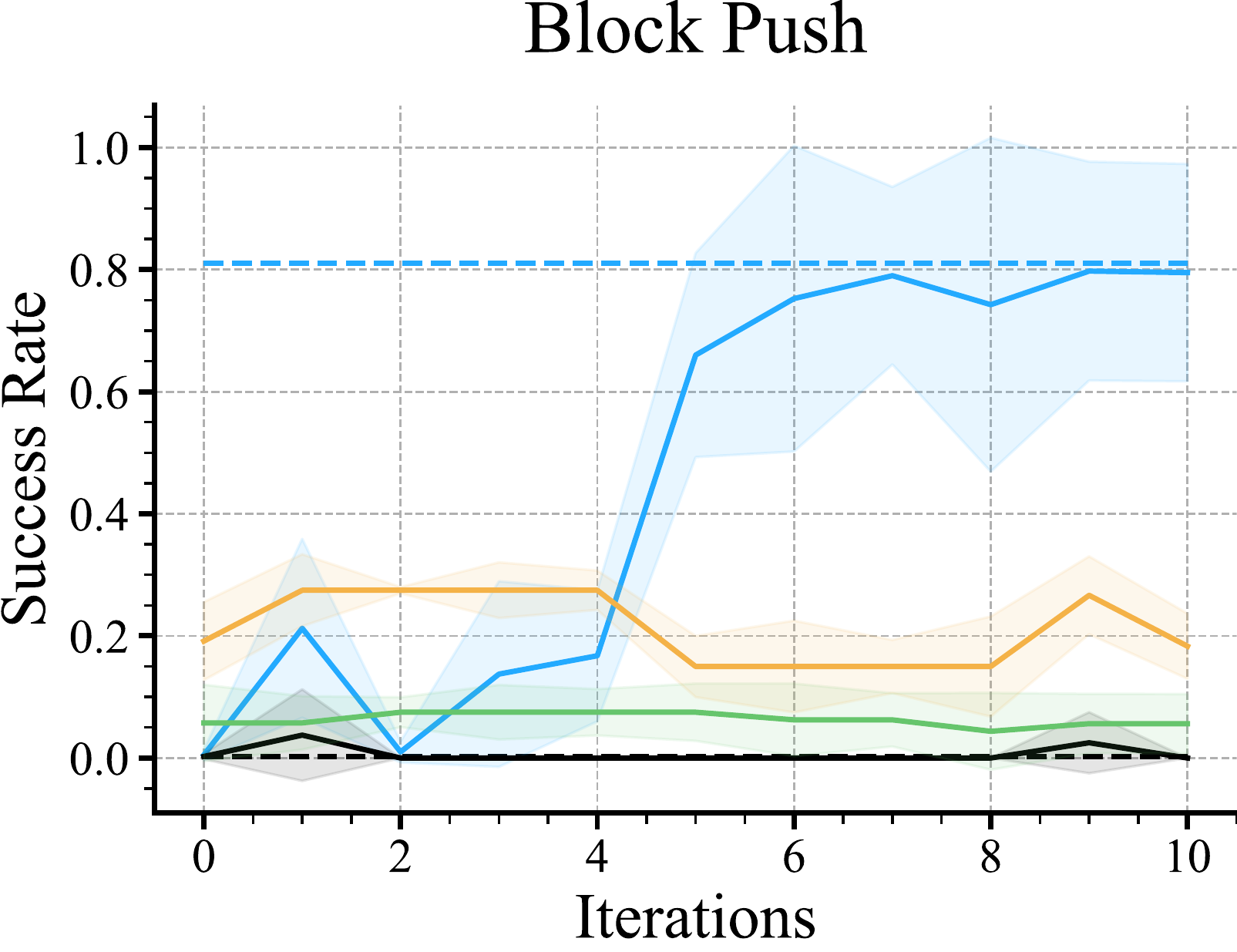}
        \caption{$\Delta \sim \mathcal{U}(0.8, 0.9)$}
    \end{subfigure}
    \begin{subfigure}[b]{0.24\textwidth}
        \includegraphics[width=\linewidth]{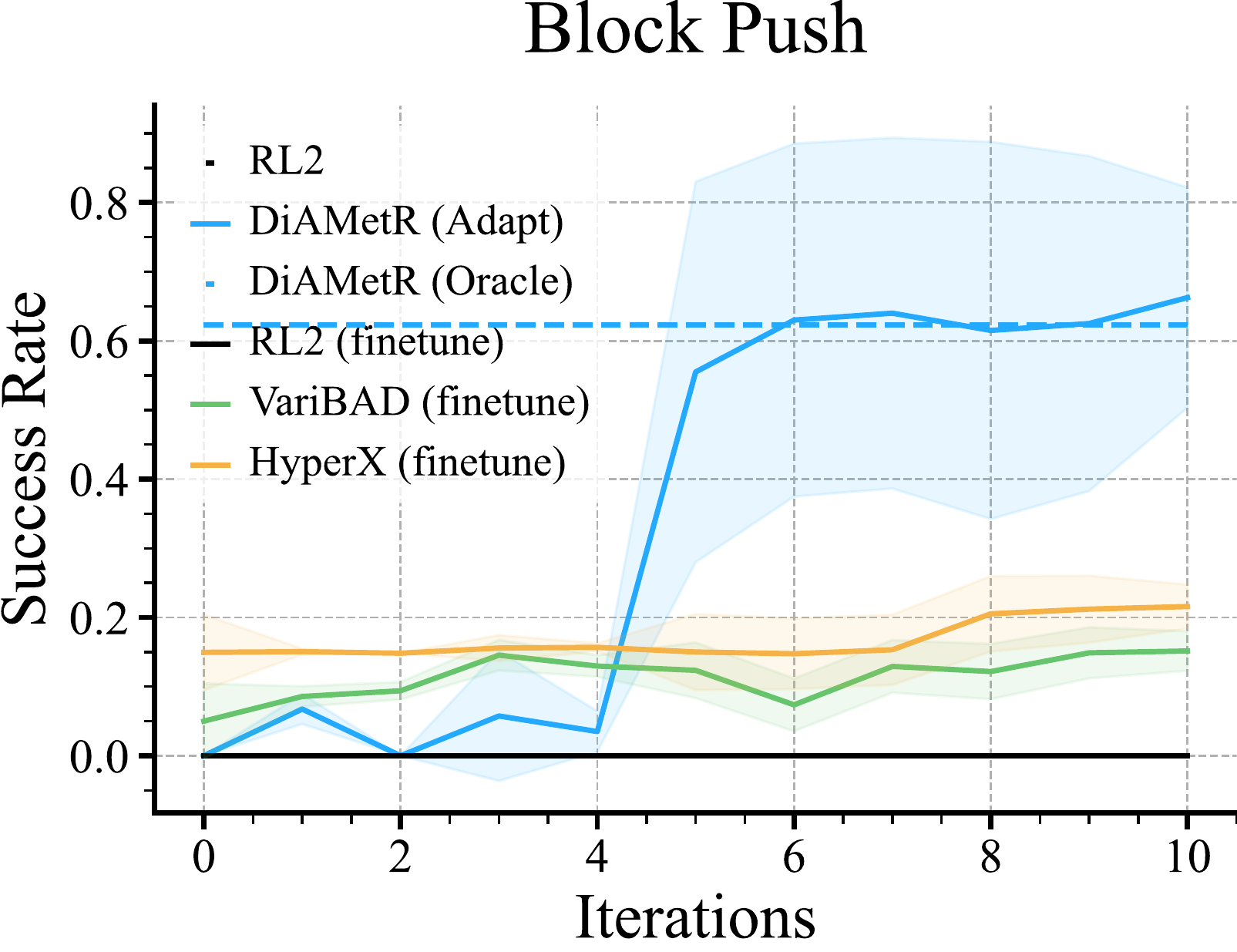}
        \caption{$\Delta \sim \mathcal{U}(0.9, 1.0)$}
    \end{subfigure}
    \caption{We compare test time adaptation of \rml~with test time finetuning of $\text{RL}^2$ on block push for various test task distributions. We run the adaptation procedure for $10$ iterations collecting $25$ rollouts per iteration.}
    \label{fig:ant_goal_test_adapt}
\end{figure}

% Might move to appendix
\section{Ablation studies}
\label{sec:ablate}
\paragraph{Can meta RL achieve robustness to task distribution shifts with improved meta-exploration?} To test if improved meta-exploration can help meta-RL algorithm achieve robustness to test-time task distribution shifts, we test HyperX~\cite{zintgraf2021exploration} on test-task distributions in different environments. HyperX leverages curiosity-driven exploration to visit novel states for improved meta-exploration during meta-training. Despite improved meta-exploration, HyperX fails to adapt to test-time task distribution shifts (see Figure~\ref{fig:adapt_diff_shifts} and Figure~\ref{fig:point_exps} for results on results on different environments). This is because HyperX aims to minimize regret on train-task distribution and doesn't leverage the visited novel states to learn new behaviors helpful in adapting to test-time task distribution shifts. Furthermore, we note that the contributions of HyperX is complementary to our contributions as improved meta-exploration would help us better learn robust meta-policies. 

\begin{figure}[H]
    \centering
    \includegraphics[width=0.24\linewidth]{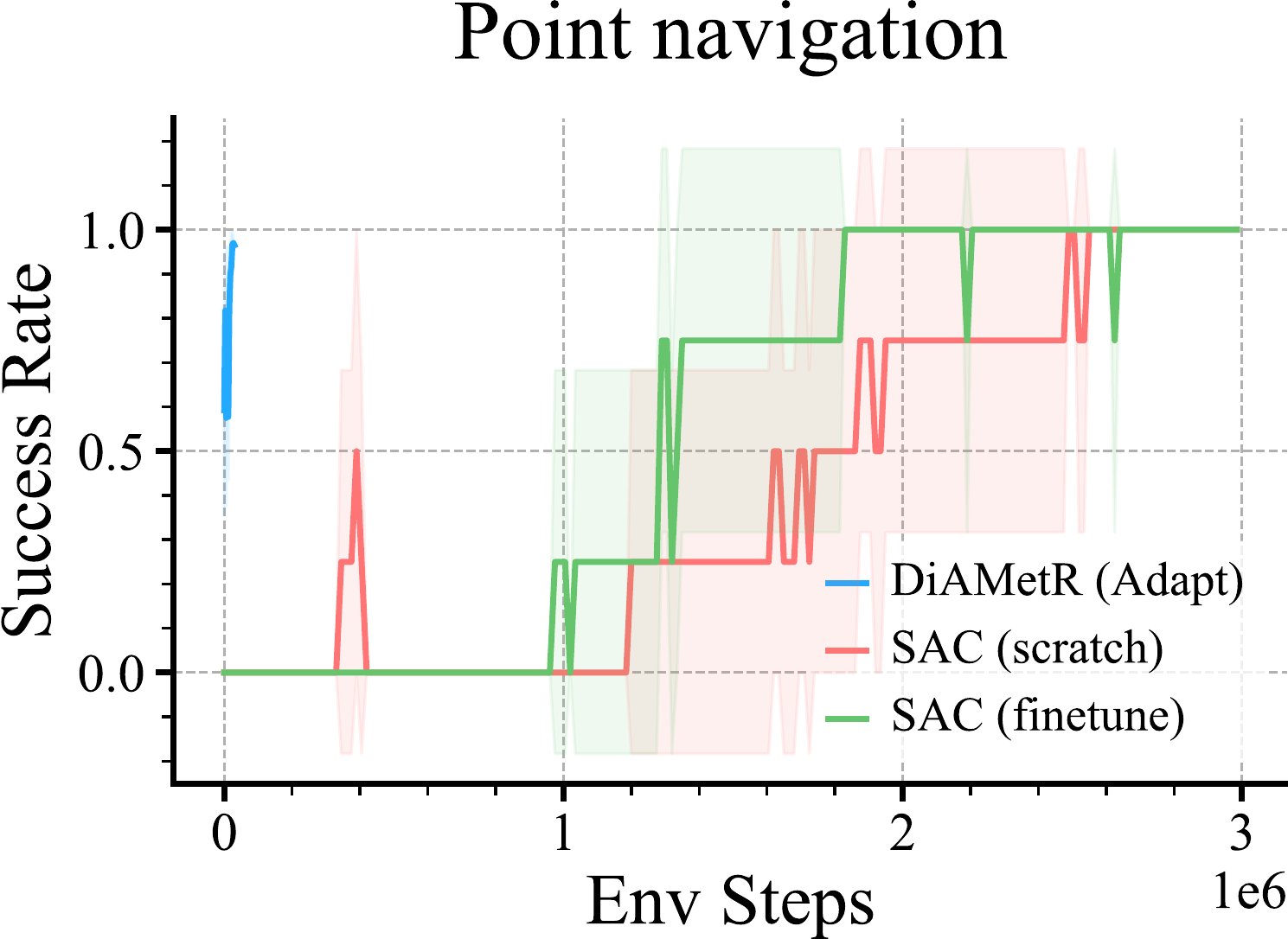}
    \caption{Both SAC trained from scratch (SAC scratch) and SAC pre-trained on training task distribution (SAC finetune) take more than a million timesteps to solve test tasks. In comparison, \rml~takes $30k$ timesteps to select the right meta-policy which then solves new tasks from test distribution in $k-1$ environment episodes (i.e. $60$ timesteps given $k=2$ and horizon $H=60$).}
    \label{fig:point_abl}
\end{figure}

\paragraph{Can RL quickly solve test time tasks?} To test if RL can quickly solve test-time tasks, we train Soft Actor Critic (SAC)~\cite{haarnoja2018soft} on $5$ tasks sampled from a particular test task distribution. To make the comparison fair, we include a baseline that pre-trains SAC on train-task distribution. Figure~\ref{fig:point_abl} shows results on \texttt{Point-navigation}. We see that both SAC trained from scratch and SAC pre-trained on training task distribution take more than a million timesteps to solve test tasks. In comparison, \rml~takes $30k$ timesteps to select the right meta-policy which then solves new tasks from test distribution in $k-1$ environment episodes (i.e. $60$ timesteps given $k=2$ and horizon $H=60$). This shows that meta-RL formulation is required for quick-adaptation to test tasks.

\section{Learning meta-policies with different support}
\begin{figure}[h]
    \centering
    \includegraphics[width=0.24\linewidth]{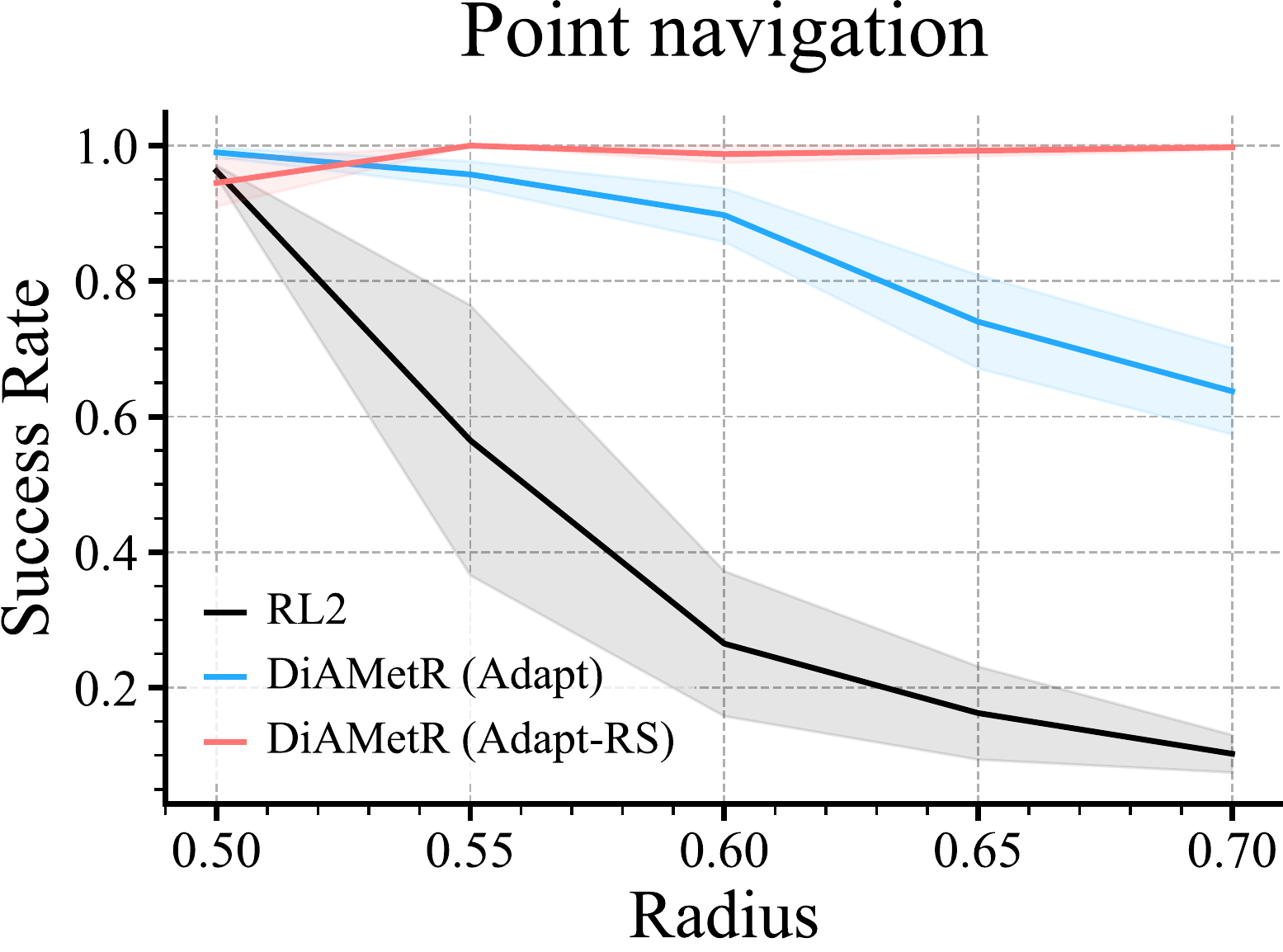}
    \includegraphics[width=0.24\linewidth]{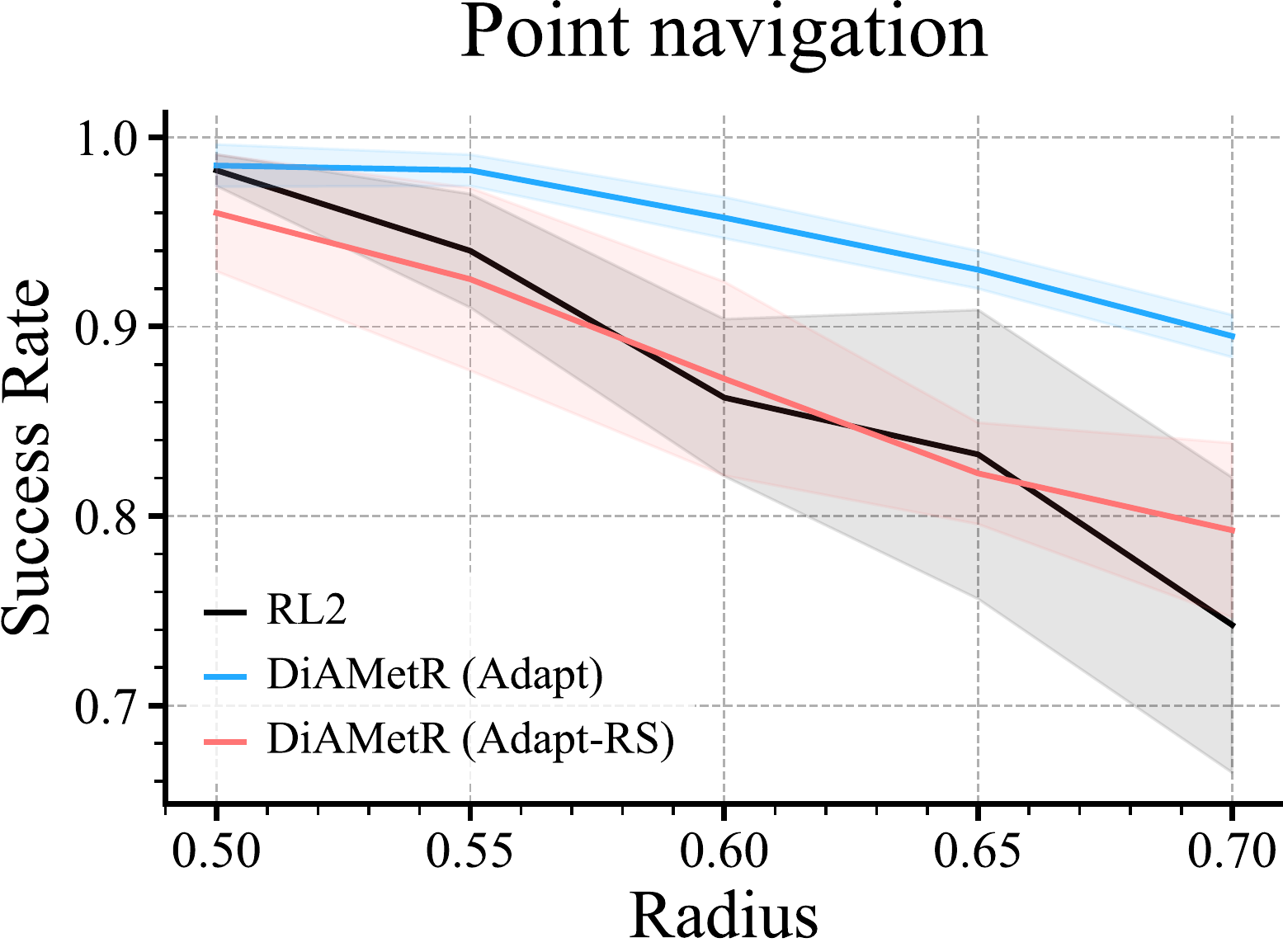}
    \caption{We investigate if learning meta-policies with different support (\rml(Adapt-RS)) is better than learning meta-policies with overlapping support (\rml(Adapt)). We evaluate it on two different families of shifted task distributions. While the left figure shows evaluations on shifted task distributions with different support (i.e. $\mathcal{U}(0, 0.5)$, $\mathcal{U}(0.5, 0.55)$, $\mathcal{U}(0.55, 0.6)$, $\mathcal{U}(0.6, 0.65)$, $\mathcal{U}(0.65, 0.7)$), the right figure shows evaluations on shifted task distributions with overlapping support (i.e. $\mathcal{U}(0, 0.55)$, $\mathcal{U}(0.0, 0.6)$, $\mathcal{U}(0.0, 0.65)$, $\mathcal{U}(0.0, 0.7)$). While \rml(Adapt-RS) performs the best on shifted task distributions with different support, \rml(Adapt) performs the best on shifted task distributions with overlapping support. Hence, whether meta-policies should have overlapping support depends on the nature of shifted test task distributions.}
    \label{fig:disk_exp}
\end{figure}

We can alternatively try to learn multiple meta-policies, each with small and different (but slightly overlapping) support. In this way, there won't be any conservativeness tradeoff. To analyze this further, we focus on point navigation domain (with train target distance distribution as $\Delta\sim\gU(0, 0.5)$) and experiment with out-of-support test task distribution.

Let's say we are learning the $i^{\text{th}}$ meta-policy (corresponding to $\epsilon_i$). Let $q_{\phi}^{i-1}(z)=\mathcal{N}(\mu^{i-1}, \sigma^{i-1})$ be the learned adversarial task distribution for $(i-1)^{\text{th}}$ meta-policy (corresponding to $\epsilon_{i-1}$). To learn meta-policies with small and different support, we make 2 changes to Algorithm~\ref{alg:meta_train_code}:
\begin{enumerate}
    \item In step $7$, we do a rejection sampling $z\sim q_\phi(z)$ with the condition that $\log q_\phi^{i-1}(z) \leq \beta$ (where $\beta$ is a hyperparameter. We found $\beta=-45$ to work well).
    \item In step $8$ and $9$, we add another constraint that $D_{\text{KL}}(p_{\text{train}}(z)||q_\phi(z)) \geq \epsilon_{i-1}$ (in addition to $D_{\text{KL}}(p_{\text{train}}(z)||q_\phi(z))\leq\epsilon_i$). This leads to learning of two weighting factors $\lambda_1$ and $\lambda_2$ (instead of just $\lambda$) that tries to ensure $D_{\text{KL}}(p_{\text{train}}(z)||q_\phi(z)) \in (\epsilon_{i-1}, \epsilon_i)$. We call this modified algorithm as \rml~(Adapt-RS) (where RS comes from rejection sampling).
\end{enumerate}

The performance of this variant depends heavily on the form of the test task distribution. We first test it on task target distance distributions $\mathcal{U}(0, 0.5), \mathcal{U}(0.5, 0.55), \mathcal{U}(0.55, 0.6), \mathcal{U}(0.6, 0.65), \mathcal{U}(0.65, 0.7)$ (essentially testing on \textit{rings} of disjoint support around the training distribution). We see that \rml(Adapt-RS) maintains a consistent success rate of $\sim1$ across various target task distributions and outperforms \rml(Adapt) and $\text{RL}^2$. This is because each meta-policy has overall smaller (hence are less conservative) and mostly different support and for this type of test distribution, this scheme can be very effective.

However, when we test it on task target distance distributions $\mathcal{U}(0, 0.5), \mathcal{U}(0, 0.55), \mathcal{U}(0.0, 0.6), \mathcal{U}(0.0, 0.65), \mathcal{U}(0.0, 0.7)$ (essentially testing on \textit{discs} which mostly include the training distribution), we see that \rml(Adapt-RS) performs same as $\text{RL}^2$ and mostly relies on the base $\text{RL}^2$ (i.e. $\epsilon=0$) for its performance.

Whether meta-policies should have overlapping support will depend on the nature of shifted test task distributions. If supports of test task distributions overlap, then it's better to have meta-policies with overlapping support. Otherwise, it's more efficient to have meta-policies with different support.
\section{Meta-policy Selection with CEM during Meta-test}

\begin{figure}[H]
    \centering
    \begin{subfigure}[b]{0.24\textwidth}
        \includegraphics[width=\linewidth]{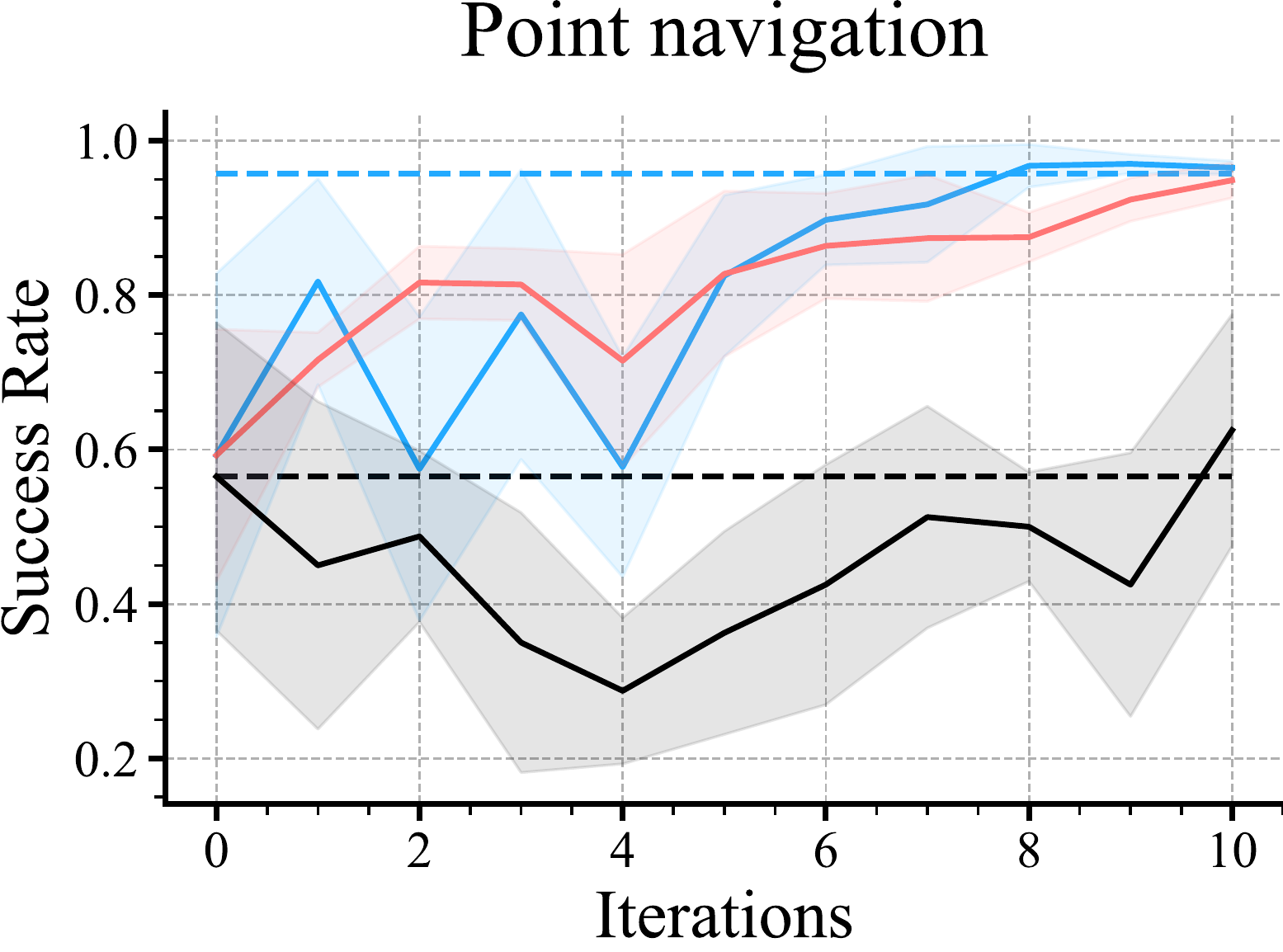}
        \caption{$\Delta \sim \mathcal{U}(0.5, 0.55)$}
    \end{subfigure}\hfill
    \begin{subfigure}[b]{0.24\textwidth}
        \includegraphics[width=\linewidth]{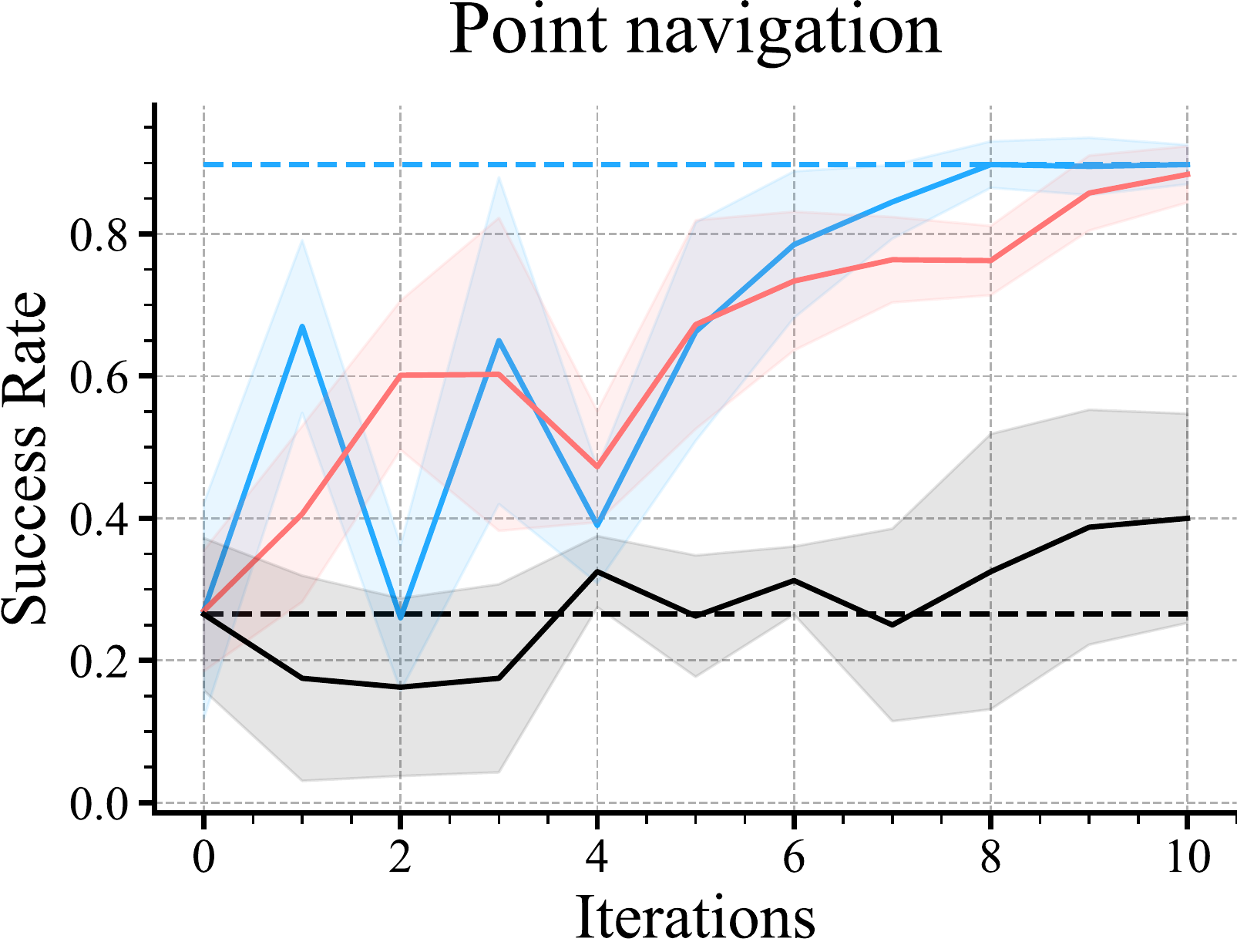}
        \caption{$\Delta \sim \mathcal{U}(0.55, 0.6)$}
    \end{subfigure}\hfill
    \begin{subfigure}[b]{0.24\textwidth}
        \includegraphics[width=\linewidth]{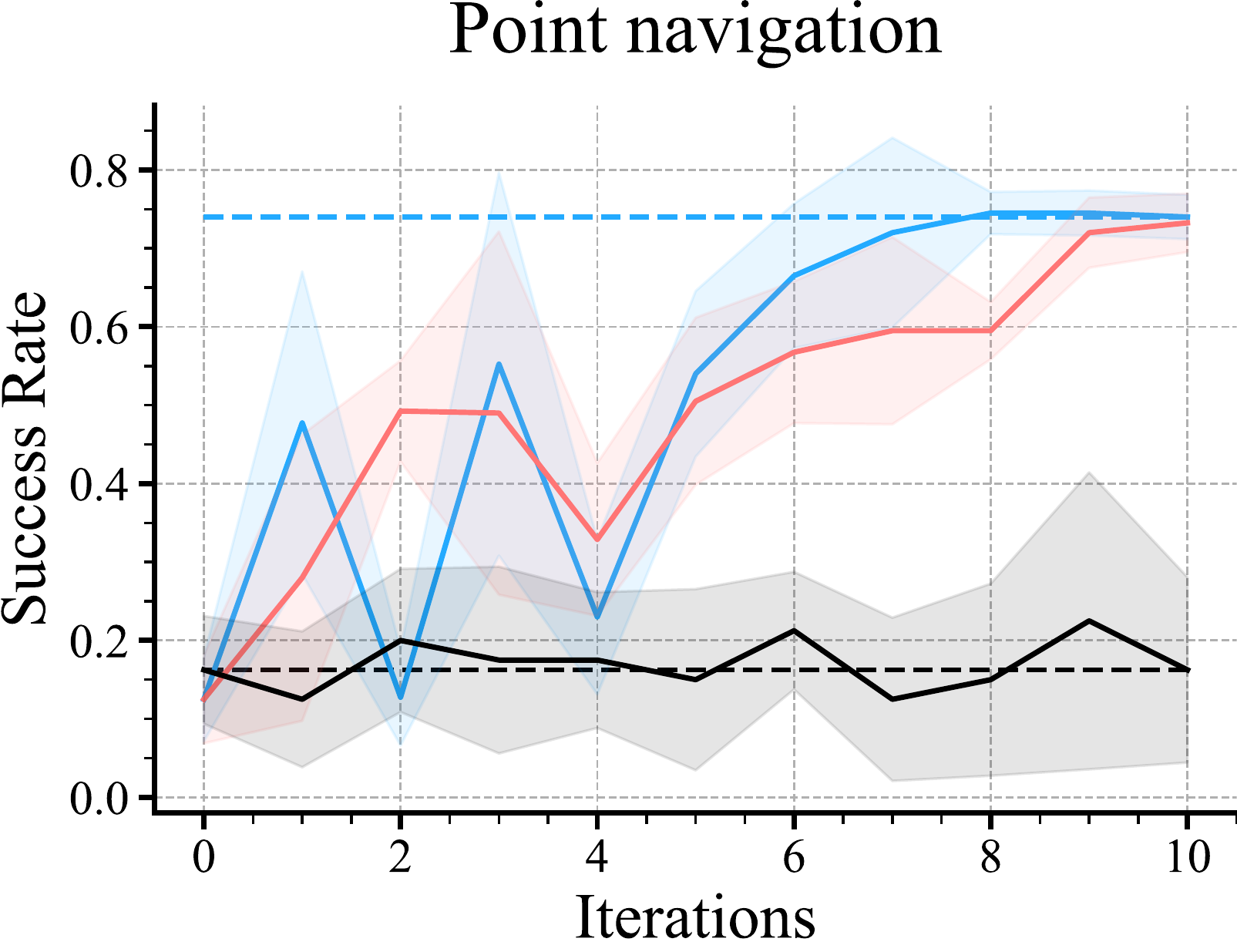}
        \caption{$\Delta \sim \mathcal{U}(0.6, 0.65)$}
    \end{subfigure}\hfill
    \begin{subfigure}[b]{0.24\textwidth}
        \includegraphics[width=\linewidth]{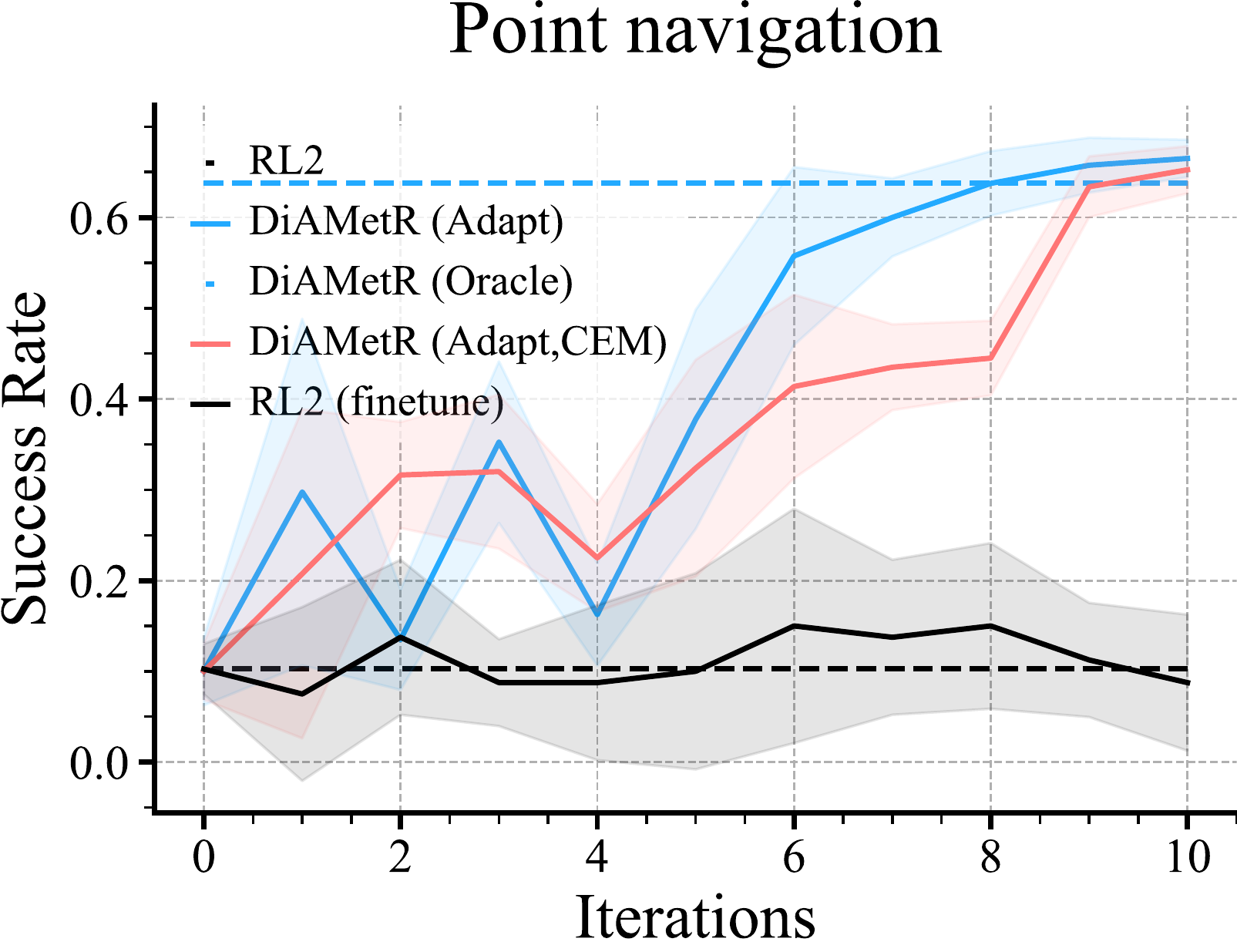}
        \caption{$\Delta \sim \mathcal{U}(0.65, 0.7)$}
    \end{subfigure}\hfill
    \caption{We compare use of Thompson's sampling (\rml(Adapt)) and Cross-entropy method (\rml(Adapt, CEM)) for test time adaptation of \rml~on point robot navigation for various test task distributions. While they have similar performance, they are both better than test-time finetuning of $\text{RL}^2$. We run the adaptation procedure for $10$ iterations collecting $25$ rollouts per iteration.
    }
    \label{fig:point_robot_test_adapt_cem}
\end{figure}

We explore using Cross-entropy method (CEM)~\cite{de2005tutorial} for meta-policy selection during meta-test phase, as an alternative to Thompson's sampling. Algorithm~\ref{alg:meta_test_code_cem} details the use of the CEM algorithm for meta-policy selection. For this evaluation, we use \texttt{point navigation} environment where tasks vary in reward functions and test task distribution is \textit{out-of-support} of training task distribution. Table~\ref{tbl:detail_task_dist} provides detailed information about these train and test task distributions. Figure~\ref{fig:point_robot_test_adapt_cem} shows that CEM has similar performance as Thompson's sampling.

% \begin{wrapfigure}{r}{0.45\textwidth}
% \vspace{-0.5cm}
\begin{algorithm}[H]
    \small
    \caption{\textbf{\rml}: Meta-test phase with CEM}
      	\label{alg:test_method}
      	\begin{algorithmic}[1]
      	\STATE Given: $p_{\text{test}}(\gT)$, $\Pi = \{\pi_{\text{meta}, \theta}^{\epsilon_i}\}_{i=1}^M$
            \STATE Sample $\pi\sim\Pi$ (with uniform probability) to collect $25$ meta-episodes
            \STATE Calculate $(\mu_\epsilon, \sigma_\epsilon)$ using $10$ (of $25$) (i.e. top $40$\%) meta-episodes with highest returns
      	\FOR{iter $t=2, 3,...,10$}
                \FOR{meta-episode $n=1,2,..,25$}
                    \STATE Sample $\epsilon\sim\gN(\mu_\epsilon, \sigma_\epsilon)$
                    \STATE Choose $\epsilon_i$ closest to $\epsilon$
                    \STATE Run $\pi_{\text{meta}, \theta}^{\epsilon_i}$ for meta-episode
                \ENDFOR
                \STATE Calculate $(\mu_\epsilon, \sigma_\epsilon)$ using $10$ (of $25$) (i.e. top $40$\%) meta-episodes with highest returns
      	\ENDFOR
      	\end{algorithmic}
\label{alg:meta_test_code_cem}
\end{algorithm}
% \vspace{-0.5cm}
% \end{wrapfigure}

\section{Proof of Proposition 4.1}

In this section, we prove the proposition in the main text about the excess regret of an $\epsilon_2$-robust policy under and $\epsilon_1$-perturbation (restated below).\\

% \textcolor{red}{Errata: the original submission has a typo, reading $\operatorname{Regret}(\pi_{\meta}^{\epsilon_1}, \gT) - \operatorname{Regret}(\pi_{\meta}^{\epsilon_2}, \gT)$ instead of the correct: $\operatorname{Regret}(\pi_{\meta}^{\epsilon_2}, \gT) - \operatorname{Regret}(\pi_{\meta}^{\epsilon_1}, \gT)$}
\tvrobustness*

\textbf{Summary of proof:} The proof proceeds in three stages: 1) deriving a form for the optimal meta-policy for a \textit{fixed} task distribution \textcolor{violet}{2)} proving that the optimal $\epsilon$-robust meta-policy takes form:
$$\pi_\meta^\epsilon(s) \propto \begin{cases} \sqrt{\frac{1 - \bar{\epsilon}}{|\gS_0|}} & s \in \gS_0 \\
\sqrt{\frac{\bar{\epsilon}}{|\gS| - |\gS_0|}} & s \notin \gS_0 
\end{cases}$$
and finally 3) showing that under the task distribution $p(\gT_g) = (1-\bar{\epsilon_1})\text{Uniform}(\gS_0) + \bar{\epsilon_1}\text{Uniform}(\gS \backslash\gS_0)$, the gap in regret takes the form in the proposition. 

\begin{proof}

For convenience, denote $\gS_1 = \gS \backslash \gS_0$ and $\regret(\pi_\meta, q(\gT)) = \E_{q}[\regret(\pi, \gT)]$. Furthermore, since the performance of a meta-policy depends only on its final-timestep visitation distribution (and any such distribution is attainable), we directly refer to $\pi(g)$ as the visited goal distribution of the meta-policy $\pi_\meta$. Recall that the regret of $\pi_\meta$ on task $\gT_g$ is given by $\frac{1}{\pi(g)}$.

We begin with the following lemma that demonstrates the optimal policy for a fixed task distribution.

\begin{lemma}
The optimal meta-policy $\pi_q^* = \argmin_{\pi} \regret(\pi, q(\gT_g))$ for a given task distribution $q(\gT_g)$ is given by 
\begin{equation}
    \pi_q^*(g) = \frac{1}{\int \sqrt{q(g')} \, dg'} \sqrt{q(g)}
\end{equation}
\end{lemma}

The proof of the lemma is similar to the argument in \citet{gupta2018unsupervised, lee2019smm}:

\begin{align}
    \pi_q^* &= \argmin_{\pi(g)} \regret(\pi,q(\gT_g)) = \argmin_{\pi(g)} \E_{\gT_g \sim q}[\frac{1}{\pi(g)}]\\
    \intertext{Letting $Z = \int_g \sqrt{q(g)}$, we can rewrite the optimization problem as minimizing an $f$-divergence (with $f(x) = \frac1x$)}
            &= \int \frac{1}{\pi(g)}q(g) \, dg\\    
            &= Z^2 \int \frac{\sqrt{q(g)} / Z}{\pi(g)} \sqrt{q(g)}/Z \, dg\\
            &= Z^2 D_f(\pi \| \sqrt{q}(g) / Z)
\end{align}

This is minimized when both are equal, i.e. when $\pi_q^*(g) = \sqrt{q} / Z$, concluding the proof.

\begin{lemma}
The optimal $\epsilon$-robust meta-policy $\pi^{*\epsilon} = \argmin \gR(\pi, p_\train, \epsilon)$ takes form $$\pi^\epsilon(g) \propto \begin{cases} \sqrt{\frac{1 - \bar{\epsilon}}{|\gS_0|}} & g \in \gS_0 \\
\sqrt{\frac{\bar{\epsilon}}{|\gS| - |\gS_0|}} & g \notin \gS_0 
\end{cases}$$
\end{lemma}

Define the distribution $q^\epsilon(\gT_g) = (1-\bar{\epsilon})\text{Uniform}(\gS_0) + \bar{\epsilon}\text{Uniform}(\gS \backslash\gS_0)$ , which is an $\epsilon$-perturbation of $p_\train$ under the TV metric. We note that there are two main cases: 1) if $\bar{\epsilon} = 1 - \frac{|\gS_0|}{|\gS|}$, then $q^\epsilon$ is uniform over the entire state space, and otherwise 2) it corresponds to uniformly taking $\epsilon$-mass from $\gS_0$ and uniformly distributing it across $\gS_1$. Using the lemma, we can derive the optimal policy for $q^\epsilon$, which we denote $\pi^\epsilon$: 
\begin{align}
    \pi^\epsilon &= \argmin_{\pi(g)} \regret(\pi, q^\epsilon(\gT)) = \frac{1}{\int \sqrt{q^\epsilon(g')} \, dg'} \sqrt{q(g)},
    \intertext{Writing $Z_\epsilon = \int \sqrt{q(g')} \, dg'$, we can write this explicitly as }
    &= \frac{1}{Z_\epsilon}\begin{cases} \sqrt{\frac{1 - \bar{\epsilon}}{|\gS_0|}} & g \in \gS_0 \\
\sqrt{\frac{\bar{\epsilon}}{|\gS| - |\gS_0|}} & g \notin \gS_0 
\end{cases}
    % \intertext{for which, the meta-policy incurs regret:}
    % \regret(\pi^\epsilon, q(\gT)) &= \E_{\gT_g \sim q}[\frac{1}{\frac{1}{\E_{g \sim q}[\sqrt{q(g)}]} \sqrt{q(g)}}]\\
    % &= (\E_{g \sim q}[\sqrt{q(g)}])^2
\end{align}

We now show that there exists no other distribution $q'(\gT)$ with $TV(p_\train, q') \leq \epsilon$ for which $ \regret(\pi^\epsilon, q') \geq \regret(\pi^\epsilon, q^\epsilon)$. We break this into the two cases for $q^\epsilon$: if $q^\epsilon$ is uniform over all goals, then $\pi^\epsilon$ visits all goals equally often, and so incurs the same regret on every task distribution. The more interesting case is the second: consider any other task distribution $q'(g)$, and let $q_0', q_1'$ be the probabilities of sampling goals in $\gS_0$ and $\gS_1$ respectively under $q'$:  $q_0' = \E_{g \sim q'}[1(g \in \gS_0)]$ and $q_1' = 1- q_0'$. The regret of $\pi^\epsilon$ on $q'$ is given by 
\begin{align}
    \regret(\pi^\epsilon, q'(\gT)) &= \E_{g \sim q'}[\frac{1}{\frac{1}{Z_\epsilon} \sqrt{q(g)}}]\\
    &= Z_\epsilon(q_0' \sqrt{\frac{|\gS_0|}{1 - \bar{\epsilon}}}  + q_1' \sqrt{\frac{|\gS| - |\gS_0|}{\bar{\epsilon}}}) 
    \intertext{By construction of $\bar{\epsilon}$, we have that $\sqrt{\frac{|\gS| - |\gS_0|}{\bar{\epsilon}}} \geq \sqrt{\frac{|\gS_0|}{1 - \bar{\epsilon}}}$, and so this expression is maximized for the largest value of $q_1'$. Under a $\epsilon$-perturbation in the TV metric, the maximal value of $q_1$ is given by $\beta + \epsilon = \bar{\epsilon}:$}
    &\leq Z_\epsilon((1 - \bar{\epsilon})\sqrt{\frac{|\gS_0|}{1 - \bar{\epsilon}}}  + \bar{\epsilon} \sqrt{\frac{|\gS| - |\gS_0|}{\bar{\epsilon}}}) \\
    \intertext{This is exactly the regret under our chosen task proposal distribution $q^\epsilon(\gT)$ (which has $q_1 = \bar{\epsilon}$)}
    &=\regret(\pi^\epsilon, q^\epsilon(\gT)).
\end{align}

These two steps can be combined to demonstrate that $\pi^\epsilon$ is a solution to the robust objective. Specifically, we have that 
\begin{align}
    \gR(\pi_\epsilon, p_\train, \epsilon) = \max_{q': TV(p_\train, q') \leq \epsilon} \regret(\pi_\epsilon, q') = \regret(\pi_\epsilon, q^\epsilon)\\
\intertext{so, for any other meta-policy $\pi_\meta$, we have}
\gR(\pi, p_\train, \epsilon) = \max_{q': TV(p_\train, q') \leq \epsilon} \regret(\pi, q') \geq \regret(\pi, q^\epsilon) \geq \regret(\pi^\epsilon, q^\epsilon) = \gR(\pi^\epsilon, p_\train, \epsilon)
\end{align}

This concludes the proof of the lemma. \qedhere

Finally, to complete the proof of the original proposition, we write down (and simplify) the gap in regret between $\pi^{\epsilon_1}$ and $\pi^{\epsilon_2}$  for the task distribution $q^{\epsilon_1}$ (as described above). We begin by writing down the regret of $\pi^{\epsilon_1}$:

\begin{align}
    \regret(\pi^{\epsilon_1}, q^{\epsilon_1}(\gT)) &= Z_{\epsilon_1}((1 - \bar{\epsilon_1})\sqrt{\frac{|\gS_0|}{1 - \bar{\epsilon_1}}}  + \bar{\epsilon_1} \sqrt{\frac{|\gS_1|}{\bar{\epsilon_1}}})\\
    &=  (\sqrt{|\gS_0|(1 - \bar{\epsilon_1})}  + \sqrt{|\gS_1|(\bar{\epsilon_1})})((1 - \bar{\epsilon_1})\sqrt{\frac{|\gS_0|}{1 - \bar{\epsilon_1}}}  + \bar{\epsilon_1} \sqrt{\frac{|\gS_1|}{\bar{\epsilon_1}}})\\
    &= (1 - \bar{\epsilon_1})|\gS_0| + \bar{\epsilon_1}|\gS_1| + 2 \sqrt{\bar{\epsilon_1}(1-\bar{\epsilon_1})|\gS_0||\gS_1|}\\
    \intertext{Next, we write the regret of $\pi^{\epsilon_2}$}
    \regret(\pi^{\epsilon_2}, q^{\epsilon_1}(\gT)) &= Z_{\epsilon_2}((1 - \bar{\epsilon_1})\sqrt{\frac{|\gS_0|}{1 - \bar{\epsilon_2}}}  + \bar{\epsilon_1} \sqrt{\frac{|\gS_1|}{\bar{\epsilon_2}}})\\
    &= (\sqrt{|\gS_0|(1 - \bar{\epsilon_2})}  + \sqrt{|\gS_1|(\bar{\epsilon_2})})((1 - \bar{\epsilon_1})\sqrt{\frac{|\gS_0|}{1 - \bar{\epsilon_2}}}  + \bar{\epsilon_1} \sqrt{\frac{|\gS_1|}{\bar{\epsilon_2}}})\\
    &= (1-\bar{\epsilon_1})|\gS_0| + \bar{\epsilon_1} |\gS_1| + \bar{\epsilon_1}\sqrt{|\gS_0||\gS_1|\frac{(1-\bar{\epsilon_2})}{\bar{\epsilon_2}}} + (1-\bar{\epsilon_1})\sqrt{|\gS_0||\gS_1|\frac{(\bar{\epsilon_2})}{1- \bar{\epsilon_2}}}\\
    &= (1-\bar{\epsilon_1})|\gS_0| + \bar{\epsilon_1} |\gS_1| + \bar{\epsilon_1}\sqrt{|\gS_0||\gS_1|\frac{(1-\bar{\epsilon_2})}{\bar{\epsilon_2}}} + (1-\bar{\epsilon_1})\sqrt{|\gS_0||\gS_1|\frac{(\bar{\epsilon_2})}{1- \bar{\epsilon_2}}}\\
    &= (1-\bar{\epsilon_1})|\gS_0| + \bar{\epsilon_1} |\gS_1| + \sqrt{|\gS_0|\gS_1|\bar{\epsilon_1}(1-\bar{\epsilon_1})} \left(\sqrt{\frac{\bar{\epsilon_1}}{(1-\bar{\epsilon_1})}\frac{(1-\bar{\epsilon_2})}{\bar{\epsilon_2}}} + \sqrt{\frac{(1-\bar{\epsilon_1})}{\bar{\epsilon_1}}\frac{\bar{\epsilon_2}}{(1 - \bar{\epsilon_2})}} \right)\\
    \intertext{Now writing $c(\epsilon_1, \epsilon_2) = \sqrt{\frac{\bar{\epsilon_2}^{-1} -1 }{\bar{\epsilon_1}^{-1} - 1}} = \sqrt{\frac{1-\bar\epsilon_2}{\bar\epsilon_2}\frac{\bar\epsilon_1}{1-\bar{\epsilon_1}}}$}
    &= (1-\bar{\epsilon_1})|\gS_0| + \bar{\epsilon_1} |\gS_1| + \sqrt{|\gS_0|\gS_1|\bar{\epsilon_1}(1-\bar{\epsilon_1})} \left(c(\epsilon_1, \epsilon_2) + \frac{1}{c(\epsilon_1, \epsilon_2)}\right)\\
    &= \regret(\pi^{\epsilon_1}, q^{\epsilon_1}(\gT)) + \sqrt{|\gS_0|\gS_1|\bar{\epsilon_1}(1-\bar{\epsilon_1})} \left(c(\epsilon_1, \epsilon_2) + \frac{1}{c(\epsilon_1, \epsilon_2)} - 2\right)
    % &= \regret(\pi^{\epsilon_1}, q^{\epsilon_1}(\gT)) + \sqrt{|\gS_0|\gS_1|\bar{\epsilon_1}(1-\bar{\epsilon_1})} \left(\sqrt{\frac{\bar{\epsilon_1}}{(1-\bar{\epsilon_1})}\frac{(1-\bar{\epsilon_2})}{\bar{\epsilon_2}}} + \sqrt{\frac{(1-\bar{\epsilon_1})}{\bar{\epsilon_1}}\frac{\bar{\epsilon_2}}{(1 - \bar{\epsilon_2})}} -2 \right)\\
    % &= \regret(\pi^{\epsilon_1}, q^{\epsilon_1}(\gT)) + \sqrt{|\gS_0|\gS_1|\bar{\epsilon_1}(1-\bar{\epsilon_1})} \left(\sqrt{\frac{\bar{\epsilon_2}^{-1} - 1}{\bar{\epsilon_1}^{-1} - 1}} + \sqrt{\frac{(1-\bar{\epsilon_1})}{\bar{\epsilon_1}}\frac{\bar{\epsilon_2}}{(1 - \bar{\epsilon_2})}} -2 \right)\\
    % % + \sqrt{\frac{1-\bar{\epsilon_1}}{\bar{\epsilon_1}}\frac{(\bar{\epsilon_2})}{1- \bar{\epsilon_2}}}}\right)\\
\end{align}
This concludes the proof of the proposition. \qedhere
\end{proof}

\section{Hyperparameters Used}
\label{sec:hyperparams}
Table~\ref{tbl:vae_hyperparams} describes the hyperparameters used for the structured VAE for learning reward function distribution

\begin{table}[H]
  \caption{Hyperparameters for structured VAE}
  \label{tbl:vae_hyperparams}
  \small
  \centering
    \begin{tabular}{l|l}
    \toprule
    latent $z$ dimension (for reward and dynamics distribution) & $16$\\
    $p(z)$ (for reward and dynamics distribution) & $\gN(0,I)$\\
    $q_\psi(z|\bar{h})$ (for reward distribution) & \texttt{MLP}(hidden-layers=[256, 256, 256])\\
    $r_\omega^h(z)$ & \texttt{MLP}(hidden-layers=[256, 256, 256])\\
    $q_\psi(z|(s_t, a_t)_{t=1}^T)$ (for dynamics distribution) & \texttt{GRU}(hidden-layers=[256, 256, 256])\\
    $p_\omega^{\text{dyn}}(s,a)$ & \texttt{MLP}(hidden-layers=[256, 256, 256])\\
    Train trajectories (from train task replay buffer) & $1e6/(\text{Episodic Horizon }H$)\\
    Train Epochs & $100$\\
    initial $\log \sigma$ & $-5$\\
    $\{\epsilon_i\}_{i=1}^M$ (\textit{out-of-support} test task distributions) & $\{0.0, 0.1, 0.2, 0.3, 0.4, 0.5, 0.6, 0.7, 0.8\}$\\
    $\{\epsilon_i\}_{i=1}^M$ (\textit{in-support} test task distributions) & $\{0.0, 0.05, 0.1, 0.15, 0.2, 0.25, 0.3, 0.35, 0.4\}$\\
    $n_\text{tr}$ (num tasks in empirical $p_{\train}(\gT)$) & $200$\\
    \bottomrule
  \end{tabular}
\end{table}

We use off-policy $\text{RL}^2$~\cite{ni2022recurrent} as our base meta-learning algorithm. We borrow the implementation from \hyperlink{https://github.com/twni2016/pomdp-baselines}{https://github.com/twni2016/pomdp-baselines}. We use the hyperparameters from the config file \hyperlink{https://github.com/twni2016/pomdp-baselines/blob/main/configs/meta/ant_dir/rnn.yml}{https://github.com/twni2016/pomdp-baselines/blob/main/configs/meta/ant\_dir/rnn.yml} but found $1500$ \texttt{num-iters} was sufficient for convergence of the meta-RL algorithm. Furthermore, we use $200$ \texttt{num-updates-per-iter}. Our codebase can be found at \hyperlink{https://drive.google.com/drive/folders/1KTjst_n0PlR0O7Ez3-WVj0jbgnl1ELD3?usp=sharing}{https://drive.google.com/drive/folders/1KTjst\_n0PlR0O7Ez3-WVj0jbgnl1ELD3?usp=sharing}.

We parameterize $q_\phi(z)$ as a normal distribution $\gN(\mu, \sigma)$ with $\phi = (\mu, \sigma)$ as parameters. We use REINFORCE with trust region constraints (i.e. Proximal Policy Optimization~\cite{schulman2017ppo}) for optimizing $q_\phi(z)$. We borrow our PPO implementation from the package \hyperlink{https://github.com/ikostrikov/pytorch-a2c-ppo-acktr-gail}{https://github.com/ikostrikov/pytorch-a2c-ppo-acktr-gail} and default hyperparameters from \hyperlink{https://github.com/ikostrikov/pytorch-a2c-ppo-acktr-gail/blob/master/a2c\_ppo\_acktr/arguments.py}{https://github.com/ikostrikov/pytorch-a2c-ppo-acktr-gail/blob/master/a2c\_ppo\_acktr/arguments.py}. Table~\ref{tbl:ppo_hyperparams} describes the hyperparameters for PPO that we changed.

\begin{table}[H]
  \caption{Hyperparameters for PPO for training $q_\phi$ per every meta-RL iteration}
  \label{tbl:ppo_hyperparams}
  \small
  \centering
    \begin{tabular}{l|l}
    \toprule
    num-processes & $1$\\
    ppo-epoch & $10$\\
    num-iters & $3$\\
    num-env-trajectories-per-iter & $100$\\
    \bottomrule
  \end{tabular}
\end{table}

We use off-policy VariBAD~\cite{dorfman2020offline} implementation from the package \hyperlink{https://github.com/twni2016/pomdp-baselines/tree/main/BOReL}{https://github.com/twni2016/pomdp-baselines/tree/main/BOReL} with their default hyperparameters. We use HyperX~\cite{zintgraf2021exploration} implementation from the package \hyperlink{https://github.com/lmzintgraf/hyperx}{https://github.com/lmzintgraf/hyperx} with their default hyperparameters. To make the comparisons fair, we ensure that the policy and the Q-function in VariBAD and HyperX have same architecture as that in off-policy $\text{RL}^2$~\cite{ni2022recurrent}.

\section{Visualizing meta-policies chosen by Thompson's sampling}
In this section, we visualize the behavior of different meta-policies (robust to varying levels of distribution shift) towards end of their training. We additionally plot the meta-policies chosen by Thompson's sampling during meta-test phase for different task distribution shifts. We choose \texttt{Ant navigation} task for this evaluation with training task target distance distribution as $\gU(0, 0.5)$ and test task distributions being \textit{out-of-support} of training task distribution (see Table~\ref{tbl:detail_task_dist} for detailed description of these distributions).

\begin{figure}[H]
    \centering
    \includegraphics[width=0.4\linewidth]{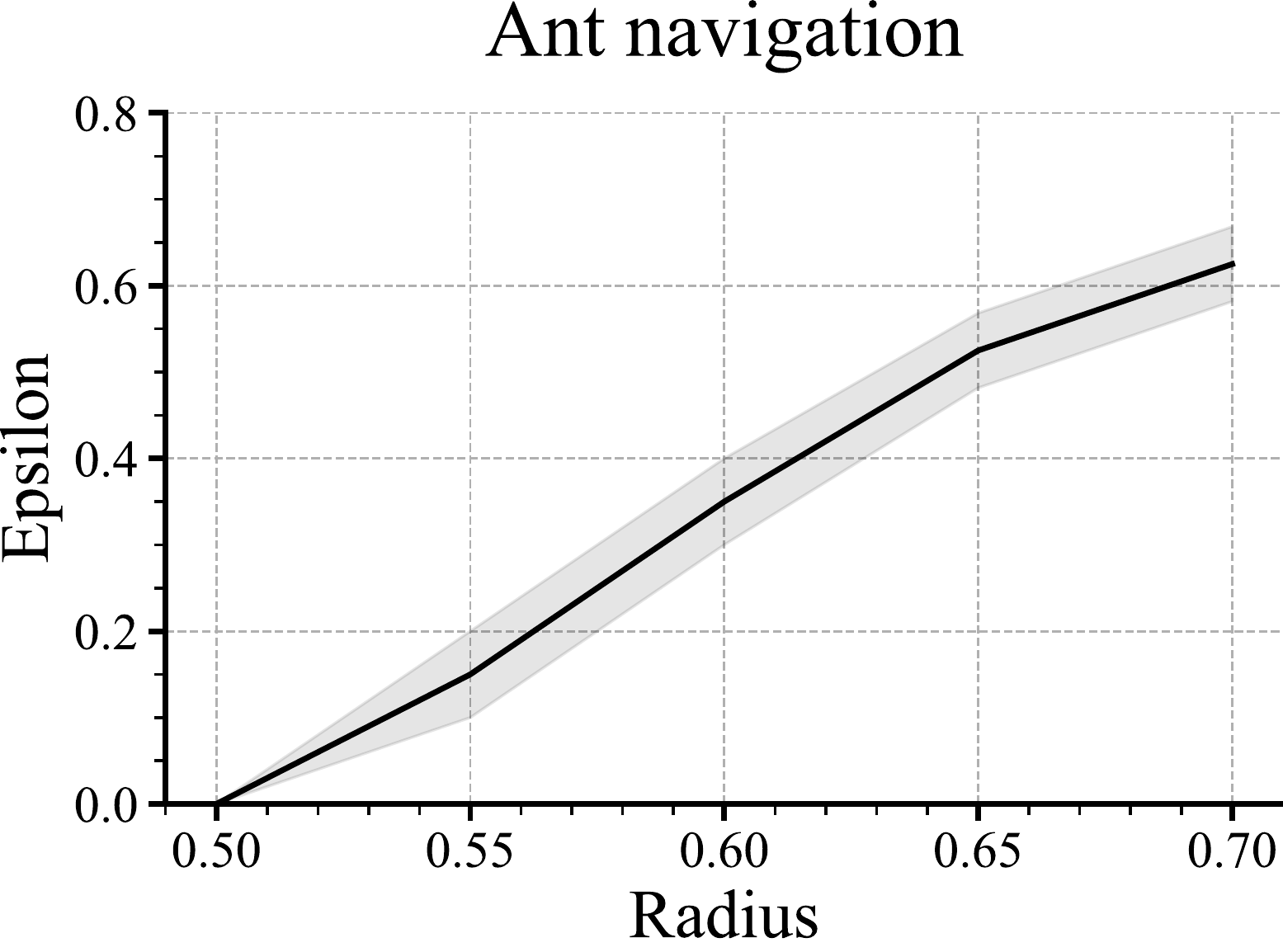}
    \caption{$\epsilon$ values corresponding to meta-policies chosen by Thompson's sampling during meta-test for different test task distributions of \texttt{Ant navigation}. The first point $r_{\text{train}}$ on the horizontal axis indicates the training target distance $\Delta$ distribution $\mathcal{U}(0, r_{\text{train}})$ and the subsequent points $r^i_{\text{test}}$ indicate the shifted test target distance $\Delta$ distribution $\mathcal{U}(r^{i-1}_\text{test}, r^i_\text{test})$.}
    \label{fig:viz_eps}
\end{figure}

\begin{figure}
    \centering
    \begin{subfigure}[b]{0.2\textwidth}
        \includegraphics[width=\linewidth]{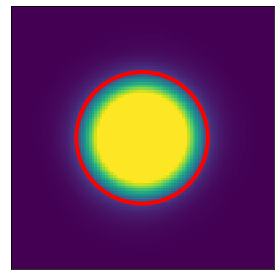}
        \caption{$\epsilon = 0.0$}
    \end{subfigure}
    \begin{subfigure}[b]{0.2\textwidth}
        \includegraphics[width=\linewidth]{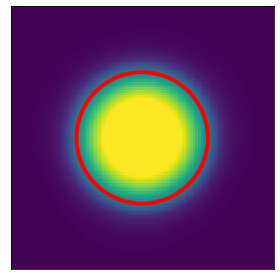}
        \caption{$\epsilon = 0.1$}
    \end{subfigure}
    \begin{subfigure}[b]{0.2\textwidth}
        \includegraphics[width=\linewidth]{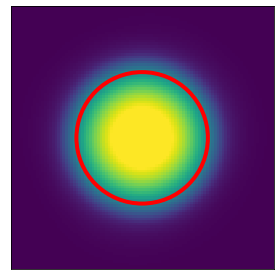}
        \caption{$\epsilon = 0.2$}
    \end{subfigure}
    \begin{subfigure}[b]{0.2\textwidth}
        \includegraphics[width=\linewidth]{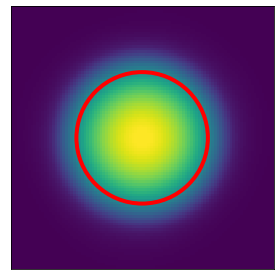}
        \caption{$\epsilon = 0.3$}
    \end{subfigure}
    \begin{subfigure}[b]{0.2\textwidth}
        \includegraphics[width=\linewidth]{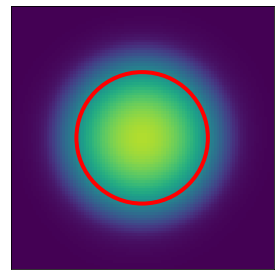}
        \caption{$\epsilon = 0.4$}
    \end{subfigure}
    \begin{subfigure}[b]{0.2\textwidth}
        \includegraphics[width=\linewidth]{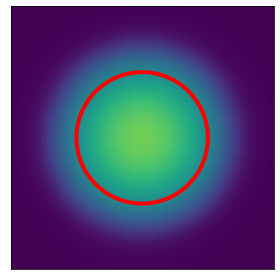}
        \caption{$\epsilon = 0.5$}
    \end{subfigure}
    \begin{subfigure}[b]{0.2\textwidth}
        \includegraphics[width=\linewidth]{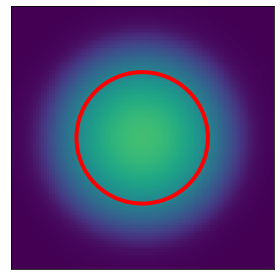}
        \caption{$\epsilon = 0.6$}
    \end{subfigure}
    \begin{subfigure}[b]{0.2\textwidth}
        \includegraphics[width=\linewidth]{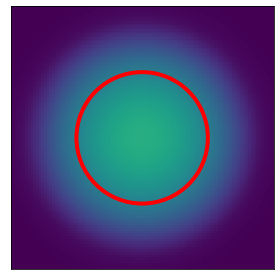}
        \caption{$\epsilon = 0.7$}
    \end{subfigure}
    \begin{subfigure}[b]{0.2\textwidth}
        \includegraphics[width=\linewidth]{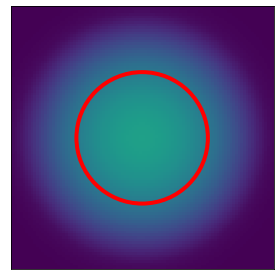}
        \caption{$\epsilon = 0.8$}
    \end{subfigure}
    \caption{We visualize the Ant's $x-y$ position visitation heatmaps for different meta-policies towards the end of their training (i.e., in their last $1$ million environment steps, out of $15.2$ million environment steps of training). Here, $\epsilon$ indicates the level of robustness of the meta-policy $\pi_\theta^\epsilon$. The red circle visualizes the training task distribution for \texttt{Ant navigation}.}
    \label{fig:diff_eps_meta_policy}
\end{figure}

To visualize the behavior of different meta-policies, we extract the $x-y$ position of Ant from the states visited by these meta-policies in their last million environment steps (out of their total $15.2$ million environment steps). We pass these $x-y$ positions through a gaussian kernel and generate visitation heatmaps of the Ant's $x-y$ position, as shown in Figure~\ref{fig:diff_eps_meta_policy}. Figure~\ref{fig:viz_eps} plots the $\epsilon$ values corresponding to meta-policies chosen by Thompson's sampling during meta-test for different test task distributions.

\section{Cost of learning multiple meta-policies during meta-train phase}
Training a population of meta-policies mainly requires more memory (both RAM and GPU) as the meta-policies are trained in parallel. It is true that granularity of epsilon affects the performance of DiAMeTR (as argued in section 4.3). If we increase the number of trained meta-policies, even though DiAMetR's final performance on various shifted test task distribution would improve, it would take more time/samples for test-time adaptation. While choosing the number of trained meta-policies, we need to balance between final asymptotic performance and time/samples taken for test-time adaptation. The main benefit of learning a distribution of such meta-policies is that it amortizes over many different shifted test time distributions and the meta-policies do not need to be relearned for each of these.
% \input{texs/visualizing_policies_chosen.tex}

% Optionally include extra information (complete proofs, additional experiments and plots) in the appendix.
% This section will often be part of the supplemental material.

\end{document}